\documentclass[DIV15, a4paper]{scrartcl}

\usepackage[ruled, linesnumbered, vlined, commentsnumbered]{algorithm2e}
\usepackage{authblk}
\usepackage{prettyref}
\usepackage{graphicx}
\usepackage{subfig}
\usepackage{booktabs}
\usepackage{colortbl}
\usepackage{amssymb}
\usepackage{amsfonts}
\usepackage{amsmath}
\usepackage{amsthm}
\usepackage{multirow}
\usepackage{tabularx}
\usepackage{footnote}
\usepackage{threeparttable}
\usepackage{hyperref}
\usepackage{amsopn}
\usepackage{hhline}
\usepackage{cite}
\usepackage{subfig}
\usepackage{setspace}
\usepackage{algorithmic}
\usepackage{rotating}

\usepackage{xcolor}  
\definecolor{myblue}{RGB}{34,31,217}

\definecolor{mycyan}{gray}{.7}

\DeclareMathOperator*{\argmax}{argmax}

\newrefformat{fig}{Fig.~\ref{#1}}
\newrefformat{tab}{Table~\ref{#1}}
\newrefformat{sec}{Section~\ref{#1}}
\newrefformat{app}{Appendix~\ref{#1}}
\newrefformat{alg}{Algorithm~\ref{#1}}
\newrefformat{property}{Property~\ref{#1}}
\newrefformat{theorem}{Theorem~\ref{#1}}
\newrefformat{corollary}{Corollary~\ref{#1}}
\newrefformat{proposition}{Proposition~\ref{#1}}
\newrefformat{def}{Definition~\ref{#1}}
\newrefformat{lemma}{Lemma~\ref{#1}}
\newrefformat{eq}{equation~(\ref{#1})}

\usepackage{graphicx}
\definecolor{Gray}{gray}{0.9}
\usepackage{scrpage2}

\usepackage{lscape}

\begin{document}
\title{\textbf\LARGE\fontfamily{cmss}\selectfont What Weights Work for You? Adapting Weights for Any Pareto Front Shape in Decomposition-based Evolutionary Multi-Objective Optimisation}


\author[1]{\normalsize\fontfamily{lmss}\selectfont Miqing Li, and Xin Yao}
\affil[1]{\normalsize\fontfamily{lmss}\selectfont CERCIA, School of Computer Science, University of Birmingham, Birmingham B15 2TT, U.~K.}
\affil[$\ast$]{\normalsize\fontfamily{lmss}\selectfont Email: limitsing@gmail.com, x.yao@cs.bham.ac.uk}

\renewcommand\Authands{Miqing Li and Xin Yao}

\date{}
\maketitle

{\normalsize\fontfamily{lmss}\selectfont\textbf{Abstract:} 	The quality of solution sets generated by decomposition-based evolutionary 
	multiobjective optimisation (EMO) algorithms 
	depends heavily on the consistency between a given problem's Pareto front shape 
	and the specified weights' distribution. 
	A set of weights distributed uniformly in a simplex often lead to 
	a set of well-distributed solutions on a Pareto front with a simplex-like shape, 
	but may fail on other Pareto front shapes.
	It is an open problem on how to specify a set of appropriate weights 
	without the information of the problem's Pareto front beforehand.
	In this paper, 
	we propose an approach to adapt the weights during the evolutionary process (called AdaW).
	AdaW progressively seeks a suitable distribution of weights for the given problem 
	by elaborating five parts in the weight adaptation --- 
	weight generation,
	weight addition, 
	weight deletion,
	archive maintenance, 
	and weight update frequency.
	Experimental results have shown the effectiveness of the proposed approach.
	AdaW works well for Pareto fronts with very different shapes: 
	1) the simplex-like, 
	2) the inverted simplex-like,
	3) the highly nonlinear,
	4) the disconnect,
	5) the degenerated,
	6) the badly-scaled, and 
	7) the high-dimensional.}

{\normalsize\fontfamily{lmss}\selectfont\textbf{Keywords:} Multi-objective optimisation, many-objective optimisation, evolutionary algorithms, decomposition-based EMO, weight adaptation}

\section{Introduction}

Decomposition-based evolutionary multiobjective optimisation (EMO) is a general-purpose algorithm framework. 
It decomposes a multi-objective optimisation problem (MOP) into a number of single-objective optimisation sub-problems 
on the basis of a set of weights (or called weight vectors) and 
then uses a search heuristic to optimise these sub-problems simultaneously and cooperatively.
Compared with conventional Pareto-based EMO, 
decomposition-based EMO has clear strengths, 
e.g., providing high selection pressure toward the Pareto front \cite{Zhang2007}, 
being easy to work with local search operators \cite{Ishibuchi2003, Martinez2012, Derbel2016},
owning high search ability for combinatorial MOPs \cite{Ishibuchi1998, Mei2011, Shim2012, Almeida2012},
and being capable of dealing with MOPs with many objectives \cite{Hughes2005, Asafuddoula2015b, LiK2015, Yuan2016} 
and MOPs with a complicated Pareto set \cite{Li2009a,Liu2014,LiK2014,Zhou2016}.

\begin{figure}[tbp]
	\begin{center}
		\footnotesize
		\begin{tabular}{cc}
			\includegraphics[scale=0.29]{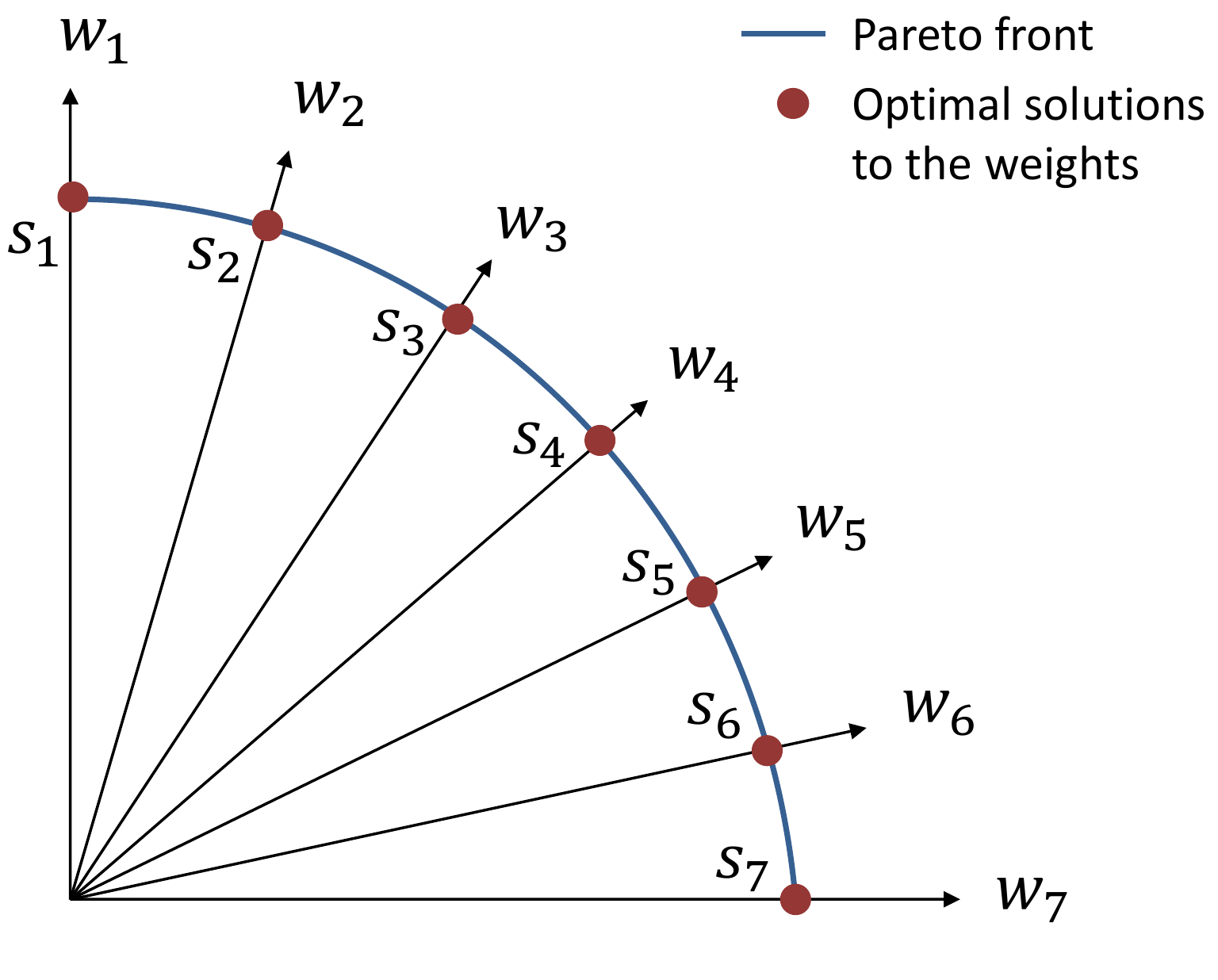}&
			\includegraphics[scale=0.29]{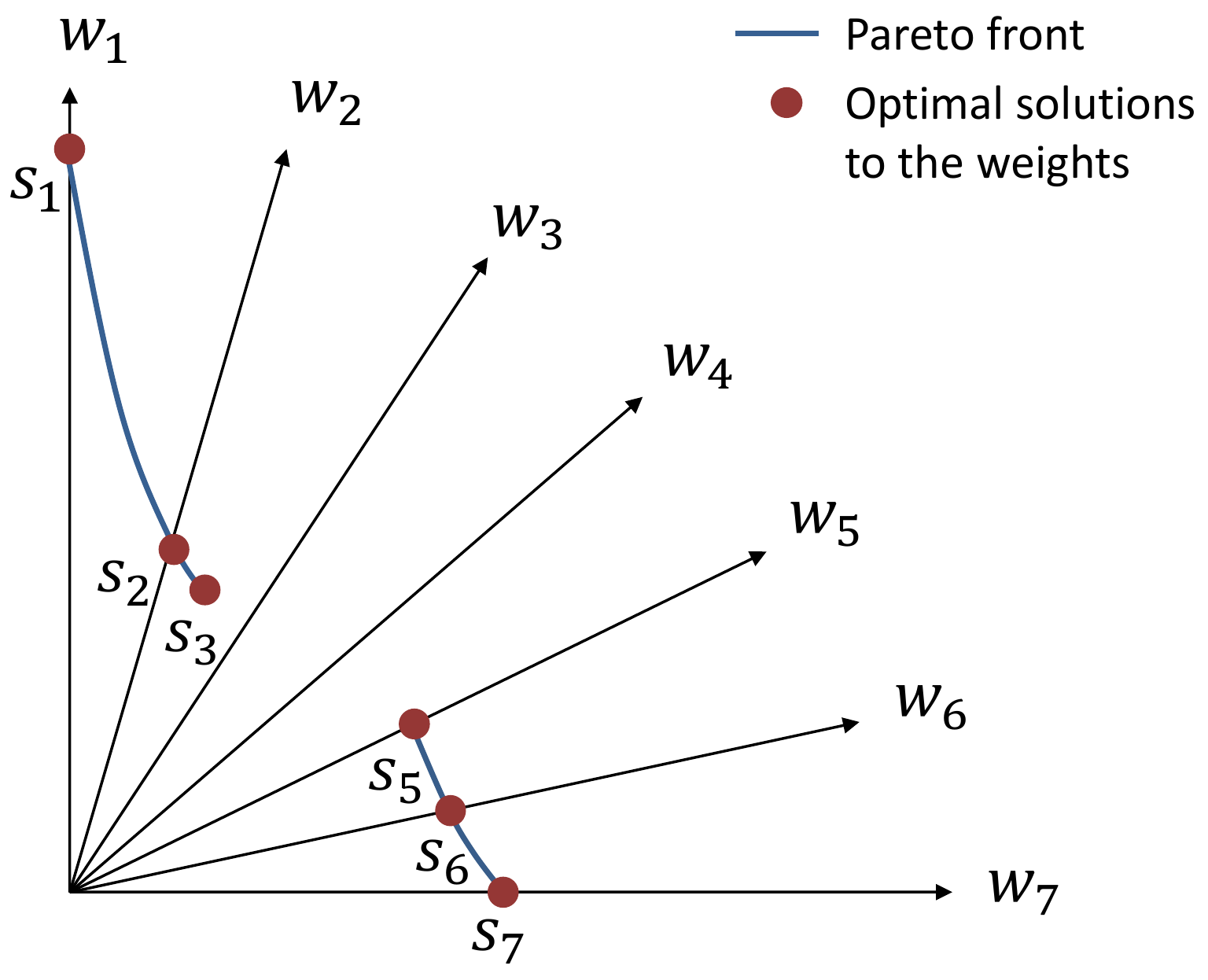}\\
			(a) On a concave Pareto front & (b) On a disconnected, convex Pareto front \\
		\end{tabular}
	\end{center}
	\caption{An example that uniformly distributed weights may lead to different distributions of optimal solutions. 
		(a) Solutions $s_1$ to $s_7$ are the optimal solutions of weights $w_1$ to $w_7$, respectively. 
		(b) Solutions $s_1, s_2, s_3, s_6$ and $s_7$ are the optimal solutions of $w_1, w_2, w_3, w_6$ and $w_7$, respectively,
		while solution $s_5$ is the optimal solution of $w_4$ and $w_5$.}
	\label{Fig:example1}
\end{figure}

A key feature in decomposition-based EMO is that the diversity of the evolutionary population
is controlled explicitly by a set of weight vectors 
(or a set of reference directions/points determined by this weight vector set).
Each weight vector corresponds to one subproblem, 
ideally associated with one solution in the population;
thus, 
diverse weight vectors may lead to different Pareto optimal solutions.
In general, 
a well-distributed solution set can be obtained 
if the set of weight vectors and the Pareto front of a given problem share the same/similar distribution shape.  
In many existing studies,
the weight vectors are predefined and distributed uniformly in a unit simplex. 
This specification can make decomposition-based algorithms well-suited 
to MOPs with a ``regular'' (i.e., simplex-like) Pareto front, 
e.g., a triangle plane or a sphere.
Figure~\ref{Fig:example1}(a) shows such an example, 
where a set of uniformly-distributed weight vectors correspond to a set of uniformly-distributed Pareto optimal solutions.

However, 
when the shape of an MOP's Pareto front is far from the standard simplex,
a set of uniformly distributed weight vectors may not result in a uniform distribution of
Pareto optimal solutions.
On MOPs with an ``irregular'' Pareto front (e.g., disconnected,  degenerate, inverted simplex-like or badly-scaled),
decomposition-based algorithms appear to struggle~\cite{Qi2014,Li2016,Ishibuchi2017,Li2017}. 
In such MOPs,
some weight vectors may have no intersection with the Pareto front. 
This could lead to several weight vectors corresponding to one Pareto optimal solution.
In addition, 
there may exist a big difference of distance between adjacent Pareto optimal solutions 
(obtained by adjacent weight vectors) in different parts of the Pareto front. 
This is due to the inconsistency between the shape of the Pareto front and the shape of the weight vector distribution.
Overall, 
no intersection between some weight vectors and the Pareto front may cause 
the number of obtained Pareto optimal solutions to be smaller than that of weight vectors, 
while the distance difference regarding adjacent solutions in different parts of the Pareto front 
can result in a non-uniform distribution of solutions.

Figure~\ref{Fig:example1}(b) gives an example that 
a set of Pareto optimal solutions are obtained by a set of uniformly-distributed weight vectors on an ``irregular'' Pareto front.
As can be seen,
weight vectors $w_3$ and $w_4$ have no intersection with the Pareto front, 
and weight vectors $w_4$ and $w_5$ correspond to only one Pareto optimal solution ($s_5$).
In addition, 
the obtained Pareto optimal solutions are far from being uniformly distributed, 
e.g., the distance between $s_1$ and $s_2$ being considerably farther than that between other adjacent solutions.

The above example illustrates the difficulties of predefining weight vectors in decomposition-based EMO.
It could be very challenging (or even impossible) to find a set of optimal weight vectors beforehand for any MOP, 
especially in real-world scenarios where the information of a problem's Pareto front is often 
unknown.

A potential solution to this problem is to seek adaptation approaches 
that can progressively modify the weight vectors according to the evolutionary population 
during the optimisation process.
Several interesting attempts have been made along this line \cite{Trivedi2016}. 
A detail review of these adaptation approaches will be presented in next section.

Despite the potential advantages of these adaptation approaches for ``irregular'' Pareto fronts,
the problem is far from being completely resolved. 
On one hand, 
varying the weight vectors which are pre-set and ideal for problems with ``regular'' Pareto fronts 
may compromise the performance of an algorithm on these problems themselves.
On the other hand,
varying the weight vectors materially changes the subproblems over the course of the optimisation,
which could significantly deteriorate the convergence of the algorithm \cite{Giagkiozis2013b}.
Overall, 
as pointed out in~\cite{LiB2015,Ishibuchi2017}, 
how to set the weight vectors is still an open question;
the need for effective methods is pressing.

In this paper, 
we present an adaptation method (called AdaW) to progressively adjust 
the weight vectors during the evolutionary process. 
AdaW updates the weight vectors periodically based on the information produced by the evolving population itself,
and then in turn guides the population by these weight vectors 
which are of a suitable distribution for the given problem.
Specifically, 
an archive is used to find out potential undeveloped, but promising weight vectors.
Then a weight vector deletion operation is used to remove 
existing unpromising weight vectors or/and weight vectors 
associated with the crowded solutions in the population.

The rest of the paper is organised as follows.
Section~II reviews related work. 
Section~III is devoted to the proposed adaptation method, 
including the basic idea and the five key issues of this adaptation.
Experimental results are presented in Section~IV. 
Finally, 
Section~V concludes the paper.

\section{Related Work}

A basic assumption in decomposition-based EMO is that 
the diversity of the weight vectors will result in the diversity of the Pareto optimal solutions. 
This motivates several studies on how to generate a set of uniformly distributed weight vectors \cite{Trivedi2016}, 
such as the simplex-lattice design \cite{Das1998}, 
two-layer simplex lattice design \cite{Deb2014}, 
multi-layer simplex lattice design \cite{Jiang2017},  
uniform design \cite{Tan2013},
and a combination of the simplex-lattice design and uniform design~\cite{Ma2014}.
A weakness of such systematic weight generators is that 
the number of generated weight vectors is not flexible,
especially in a high-dimensional space. 
This contrasts with the uniform random sampling method \cite{Jaszkiewicz2002} 
which can generate an arbitrary number of weight vectors for any dimension.  
In addition, 
some work has shown that if the geometry of the problem is known
\textit{in priori} then the optimal distribution of the weight vectors 
for a specific scalarizing function can be readily identified \cite{Giagkiozis2013,Wang2016b}. 

The variety of weight generators gives us ample alternatives 
in initialising the weight vectors, 
each of which provides an explicit way of specifying 
a set of particular search directions in decomposition-based algorithms. 
However, 
the premise that these weight generators work well is
the Pareto front of the problem sharing the simplex-like regular shape. 
An ``irregular'' Pareto front (e.g., disconnected, degenerate, inverted simplex-like or
badly-scaled) may make these weight generators struggle,
in which multiple weight vectors may correspond to one single point.
This leads to a waste of computational resources, 
and further renders the algorithm's performance inferior.

An intuitive solution to this problem is to adaptively update
the weight vectors during the optimisation process.
Several interesting attempts have been made
along this line. 
In early studies \cite{Ishibuchi1998,Jin2001,Jaszkiewicz2002}, 
researchers considered randomly-generated weight vectors (search directions) at each generation.
This can make more computational cost allocated on the area around the nondominated solutions found so far~\cite{LiH2015}.
Recently, 
Li et al.~\cite{LiH2015} presented that this strategy could be helpful in dealing with MOPs having an irregular Pareto front. 
They introduced an external population, 
which is used to store promising solutions by the $\epsilon$ dominance relation \cite{Laumanns2002}, 
to help generating the random weights.

The main trend of the weight adaptation has been on attempts to 
add/delete the weight vectors in sparse/dense regions in order to diversify solutions. 
For example, 
in \cite{Li2011}
the authors periodically adjusted each weight to increase the distance of its corresponding solutions to others.
In \cite{Jiang2011}, 
Jiang et al. presented an adaptive weight adjustment method which samples the regression curve of the weight vectors 
on the basis of an external archive. 
Gu et al. \cite{Gu2012} used the equidistant interpolation to periodically update the weight vectors on the estimated Pareto front.
Later, 
they proposed a weight adaptation method by training an self-organising map according to the current solutions \cite{Gu2017}.
Jain and Deb~\cite{Jain2014} introduced an adaptive version of NSGA-III~\cite{Deb2014} (A-NSGA-III) for irregular Pareto fronts. 
In A-NSGA-III, 
an ($m-1$)-simplex of reference points (weight vectors) centred around a crowded reference point are added in the update process, 
and the reference points which are not associated with any of the solutions are deleted.
Qi et al.~\cite{Qi2014} proposed an adaptive weight adjustment (MOEA/D-AWA) strategy for irregular Pareto fronts and
integrated the strategy within MOEA/D-DRA \cite{Zhang2009b}.
MOEA/D-AWA has two phases: 
first a set of pre-set weight vectors are used until the algorithm is considered to approach the Pareto front, 
and then the weight vectors are adjusted periodically by removing the most crowded weight vectors and 
adding new weight vectors in sparse regions.  
Liu et al. \cite{Liu2016,Liu2017} proposed an adaptive weight update in MOEA/D-M2M~\cite{Liu2014} for degenerate Pareto fronts 
according to the angle among solutions (as a similar measure) in the evolutionary population.   
Cheng et al. \cite{Cheng2015a,Cheng2016} adapted the reference vectors (weight vectors) to guide the evolutionary search.
In the presented reference vector guided evolutionary algorithm (RVEA), 
two reference vector adaptations are conducted to deal with badly-scaled Pareto fronts and irregular fronts respectively.
Very recently, 
Zhang et al. \cite{Zhang2017} designed a weight vector adaptation via a linear interpolation 
for bi-objective optimisation problems with a discontinuous Pareto front.
Wang et al. \cite{Wang2017b} considered both the ideal and nadir points to update the weight vectors in MOEA/D for a better distribution.
Cai et al. \cite{Cai2017} proposed two types of weight (direction) vector adjustments for many-objective problems, 
with one aiming at the number of the direction vectors and the other aiming at 
the positions of the direction vectors.
Asafuddoula et al. \cite{Asafuddoula2017} adapted the weight vectors 
on the basis of information collected over a ``learning period'', 
and stored the original weights which had been removed for the future use. 
In addition, 
some researchers introduced a set of weight vectors into non-decomposition-based EMO, 
and conducted the weight vector adaptation for Pareto-based search~\cite{Wang2013a,Wang2015c} and 
indicator-based search~\cite{Tian2017a}.
And some other researchers adaptively adjusted the search directions 
according to the distribution of the evolutionary population, 
which can also be seen as an adaptation of weight vectors~\cite{Xiang2017,Xiang2017b}.

In spite of the above progresses, 
challenges facing the weight adaptation remain:

\begin{itemize}

\item Adapting the weight vectors affects the convergence of the algorithm. 
Varying the weight vectors essentially changes the subproblems. 
After a change of the subproblems, 
their associated solutions need to readjust the search directions.
This could lead to the solutions to \textit{wander} in the objective space 
during the optimisation process \cite{Giagkiozis2013b}.

\item Adapting the weight vectors may compromise the performance of the algorithm on regular Pareto fronts.
Even for a regular Pareto front, 
there may exist multiple weight vectors corresponding to one single solution during the evolutionary process. 
As such, 
a change of the weight vectors which were already ideal for the considered regular Pareto front 
may lead solutions towards wrong search directions.

\item It is difficult to adapt the weight vectors for different Pareto fronts.
Many weight adaptation methods are designed or suitable for only certain types of Pareto fronts.
The variety of Pareto fronts 
(disconnected, degenerate, inverted simplex-like, badly-scaled, highly-nonlinear, or/and high-dimensional)
is a challenge to any adaptation method.
\end{itemize}

\section{The Proposed Algorithm}

\begin{figure*}[tbp]
	\begin{center}
		\footnotesize
		\begin{tabular}{ccc}
			\includegraphics[scale=0.35]{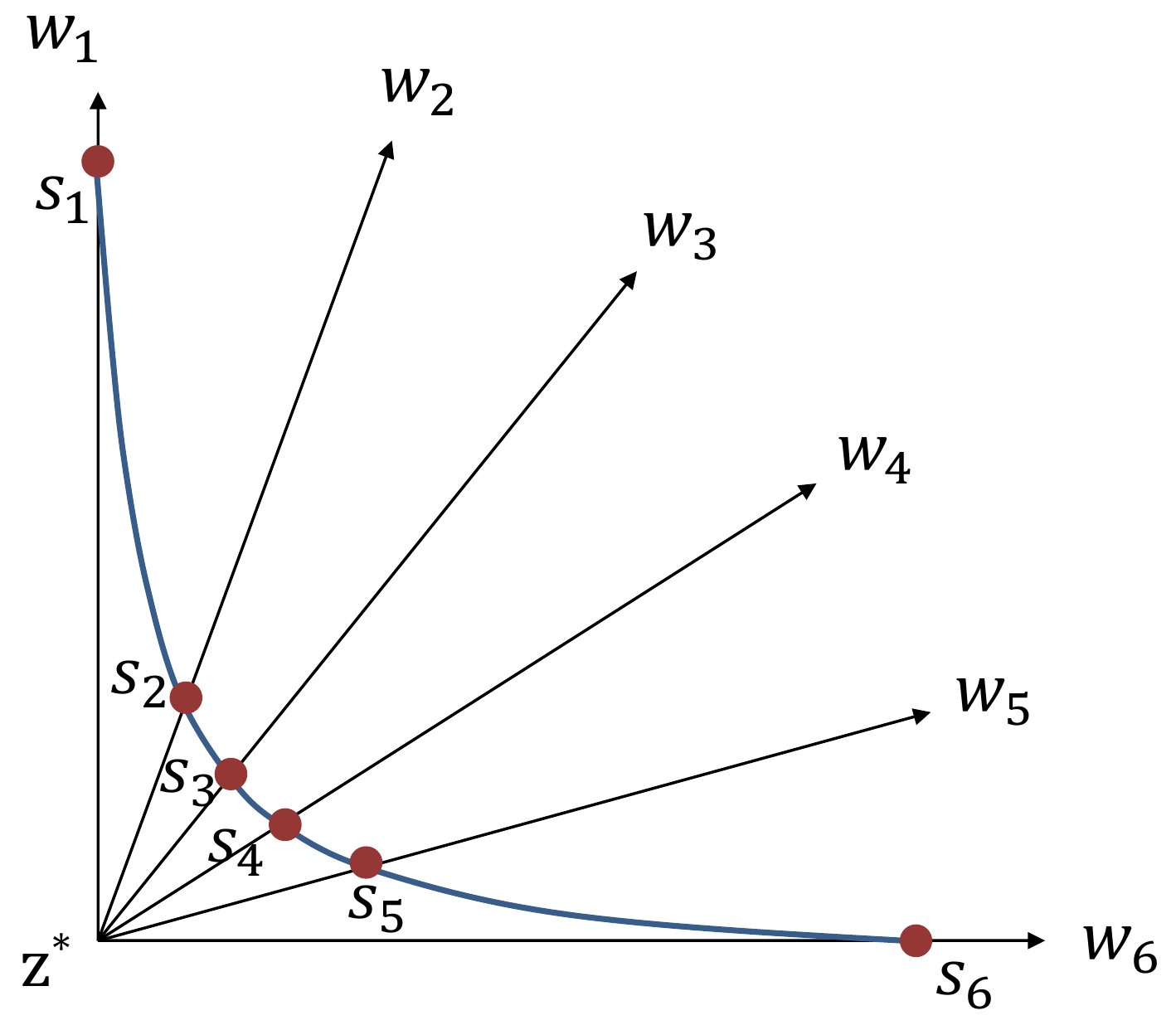}~&~
			\includegraphics[scale=0.35]{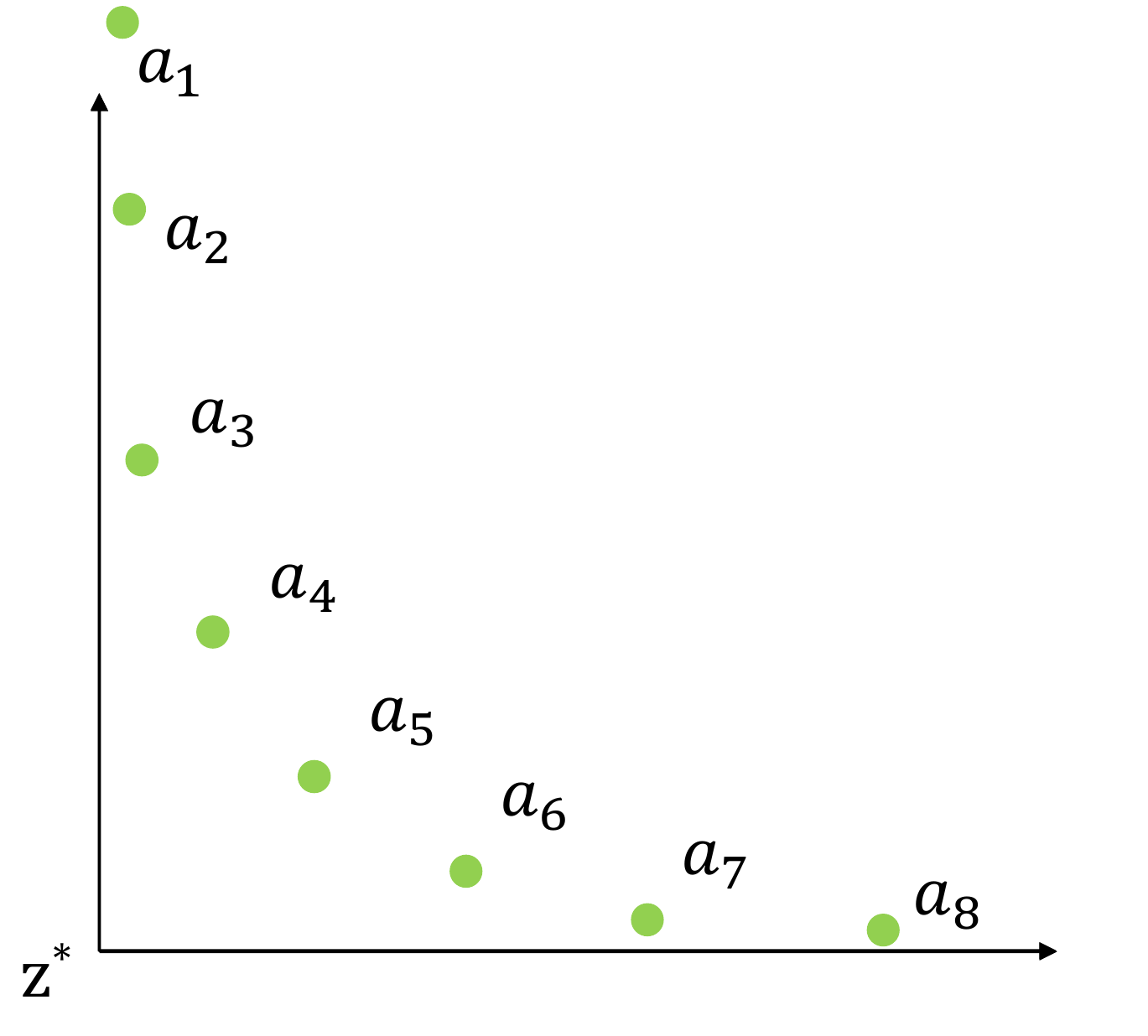}&
			\includegraphics[scale=0.35]{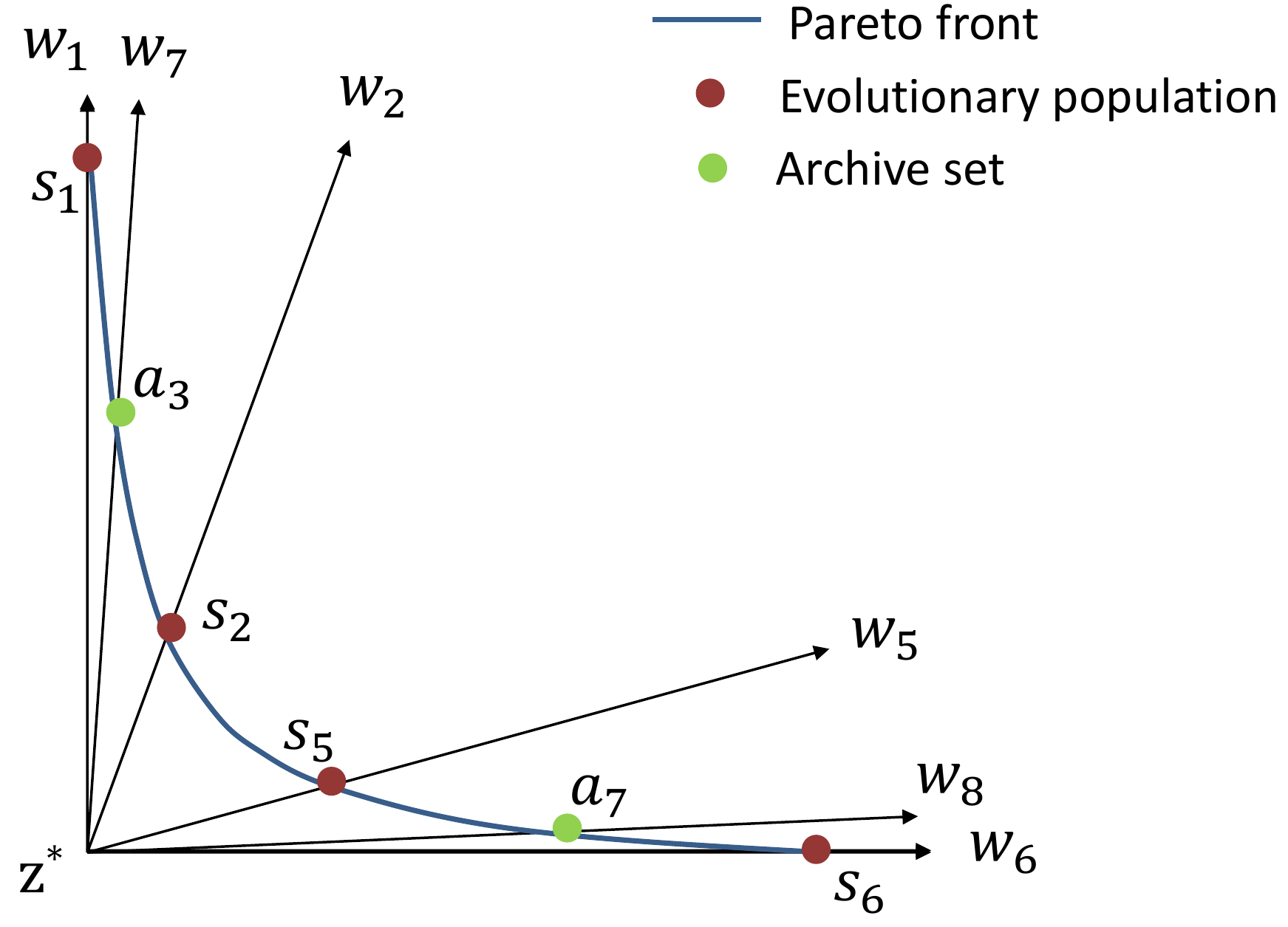}\\
			(a) Before update & (b) A well-maintained archive & (c) After update~~~~~~~~~ \\
		\end{tabular}
	\end{center}
	\caption{An illustration of updating the weight vectors of the population by the aid of a well-maintained archive set of nondominated solutions.}
	\label{Fig:example2}
\end{figure*}

\subsection{Basic Idea}

When optimising an MOP,
the current nondominated solutions (i.e., the best solutions found so far) 
during the evolutionary process can indicate the evolutionary status~\cite{Li2014e,Liu2016}.
The nondominated solution set, 
with the progress of the evolution, 
gradually approximates the Pareto front,
thus being likely to reflect the shape of the Pareto front provided that it is well maintained.
Despite that such a set probably evolves slowly in comparison with the evolutionary population 
which is driven by the scalarizing function in decomposition-based algorithms, 
the set may be able to provide new search directions that are unexplored 
by the scalarizing function-driven population.

\mbox{Figure~\ref{Fig:example2}} gives an illustration of updating the search directions (weight vectors) 
of the population by the aid of a well-maintained archive set of nondominated solutions. 
As can be seen,
before the update a set of uniformly distributed weight vectors correspond to 
a poorly distributed population along the Pareto front.
In the update, 
the two solutions from the archive ($a_3$ and $a_7$) whose areas are not explored well in the population 
are added (\mbox{Figure~\ref{Fig:example2}}(c)) and 
their corresponding weights are considered as new search directions to guide the evolution ($w_7$ and $w_8$).  
In addition, 
the weight vectors that are associated with crowded solutions ($s_3$ and $s_4$) in the population are deleted. 
Then, 
a new population is formed with poorly-distributed weight vectors but well-distributed solutions.

The above is the basic idea of the weight vector adaptation in our proposed work. 
However, 
materialising it requires a proper handling of several important issues.
They are
\begin{itemize}
	\item How to maintain the archive?
	
	\item Which solutions from the archive should enter the evolutionary population to generate new weight vectors?
	
	\item How to generate weight vectors on the basis of these newly-entered solutions?
	
	\item Which old weight vectors in the population should be deleted?
	
	\item What is the frequency of updating the weight vectors? 
	i.e., how long should we allow the population to evolve by the current weight vectors?  
\end{itemize}
In next five subsections, 
we will describe in sequence how we handle these issues,
followed by the main framework of the algorithm.

\subsection{Archive Maintenance}

In AdaW, 
an archive with a pre-set capacity is to store the nondominated solutions produced during the evolutionary process.
When the number of solutions in the archive exceeds the capacity, 
a maintenance mechanism is used to remove some solutions with poor distribution.
Here, 
we consider the population maintenance method in \cite{Li2016}.
This method which iteratively deletes the solution having the biggest crowding degree in the set
can preserve a set of representative nondominated solutions \cite{Li2016}.
The crowding degree of a solution is estimated by considering both the number and location of its neighbors in a niche.
Formally,
the crowding degree of a solution $p$ in the set $A$ is defined as
\begin{equation}
D(p)= 1 - \prod_{q\in A, q\neq p} R(p,q)
\label{eq:density}
\end{equation}
\begin{equation}
R(p,q) = \left \{ \begin{array}{@{}ll}
\left. d(p,q) \middle/ r \right., & \textrm{if} ~d(p,q) \leq r \\
~1, & \textrm{otherwise}
\end{array}\right.
\end{equation}
where $d(p,q)$ denotes the Euclidean distance between solutions $p$ and $q$,
and $r$ is the radius of the niche, 
set to be the median of the distances from all the solutions to their $k$th nearest solution in the set. 
Note that similar to \cite{Li2016} all the objectives are normalised 
with respect to their minimum and maximum in the considered set in AdaW.

It is worth mentioning that there are two slight differences of the settings from that in \cite{Li2016}.
First, 
parameter $k$ for the $k$th nearest neighbor was set to $3$ in \cite{Li2016}, 
while here $k$ is set to the number of objectives of the problem. 
This leads to less parameters required by the algorithm.
Second, 
the median, 
instead of the average in \cite{Li2016},
of the distances from all the solutions to their $k$th nearest neighbor is considered. 
This could alleviate the effect of the dominance resistant solutions (DRS), 
i.e., the solutions with a quite poor value in some objectives but with (near) optimal 
values in some other objectives \cite{Ikeda2001}.

\subsection{Weight Addition}

In AdaW, 
we aim to add weight vectors (into the evolutionary population) 
whose search directions/areas are \textit{undeveloped} and \textit{promising}. 
Both criteria can be measured by contrasting the evolutionary population with the archive set.
For the former, 
if the niche in which a solution of the archive is located in has no solution in the evolutionary population,    
it is likely that niche is undeveloped. 
For the latter,
if a solution of the archive performs better on its search direction (weight vector) 
than any solution of the evolutionary population,
it is likely that the niche of that solution is promising.

Specifically, 
to find out the solutions whose niche is undeveloped by the evolutionary population, 
we consider the niche size which is determined by the archive itself.
That is,
the radius of the niche is set to the median of the distances 
from all the solutions to their closest solution in the archive.
After finding out these candidate solutions whose niche are not developed by the evolutionary population,
we then consider whether they are promising or not.
First, 
we obtain their corresponding weight vectors (which will be detailed in next section).
Then for each of these weight vectors, 
we find its neighbouring weight vectors\footnote{The definition of neighbouring weight vectors is based on that in decomposition-based EMO} 
in the evolutionary population, 
and further determine the solutions associated with the neighboring weight vectors. 
Finally we compare these solutions with the candidate solution
on the basis of the candidate solution's weight vector. 
Formally, 
let $q$ be one of the candidate solutions in the archive and 
$w_q$ be its corresponding weight vector.
Let $w_p$ be one of the neighbouring weight vectors of $w_q$ in the evolutionary population, 
and $p$ be the solution associated with $w_p$ in the evolutionary population.
We define that $q$ outperforms $p$ on the basis of $w_q$, 
if
\begin{equation}
g(q, w_q) < g(p, w_q)
\end{equation}       
where $g()$ is the considered scalarizing function, 
or 
\begin{equation}
g(q, w_q) = g(p, w_q)~~\textrm{and}~~\sum_{i=1}^m f_i(q) < \sum_{i=1}^m f_i(p)
\end{equation} 
where $f_i(q)$ denotes the objective value of $q$ in the $i$th objective 
and $m$ is the number of objectives.
If $q$ outperforms all of the neighboring solutions on the basis of its weight vector, 
then $q$ will enter the evolutionary population, 
along with its search direction (weight vector). 
After that, 
the neighbouring information of $q$'s weight vector in the evolutionary population
(i.e., the solutions that the neighbouring weight vectors corresponding to) 
is updated by $q$.

\subsection{Weight Generation}

Given a reference point,
the optimal weight vector to a solution (e.g., $w_7$ to $a_3$ in \mbox{Figure~\ref{Fig:example2}(c)}) 
with respect to the Tchebycheff scalarizing function  
can be easily generated.
This is already a frequently used approach in the weight vector adaptation \cite{Gu2012,Qi2014}.  

Formally, 
let $z^* = (z_1^*, z_2^*,..., z_m^*)$ be the reference point\footnote{The reference point in decomposition-based algorithms is
	 often set to be equal to or slightly smaller than the best value found so far\cite{Wang2016a,Qi2014}; 
	 here we set it to $10^{-4}$ smaller than the best value, 
	 following the suggestions in~\cite{Wang2017}.} and 
$w = (\lambda_1, \lambda_2,..., \lambda_m)$ be the optimal weight vector to a solution $q$.
Then it holds that
\begin{equation}
\frac{f_1(q) - z_1^*}{\lambda_1} = \frac{f_2(q) - z_2^*}{\lambda_2} = \cdots = \frac{f_m(q) - z_m^*}{\lambda_m}
\end{equation}  
Since $\lambda_1 + \lambda_2 + ... + \lambda_m = 1$, 
we have 
\begin{equation}
w = (\lambda_1,..., \lambda_m) = (\frac{f_1(q) - z_1^*}{\sum_{i=1}^m f_i(q) - z_i^*},...,\frac{f_m(q) - z_m^*}{\sum_{i=1}^m f_i(q) - z_i^*})
\end{equation}

\subsection{Weight Deletion}

After the weight addition operation, 
AdaW needs to delete some weight vectors in the evolutionary population 
to keep the number of the weight vectors unchanged (i.e., back to the predefined population size $N$).
In view of that 
ideally each weight vector is associated with one distinct solution in decomposition-based EMO, 
we consider the situation that multiple weight vectors share one solution 
(e.g., weight vectors $w_4$ and $w_5$ sharing solution $s_5$ in \mbox{Figure~\ref{Fig:example1}(b)}).
Specifically, 
we find out the solution who is shared by the most weight vectors in the population.   
In these weight vectors, 
we delete the one whose scalarizing function value is the worst. 
Formally, 
let $p$ be the solution shared by the most weight vectors $w_1,w_2,...,w_n$ out of the population.
Then the weight vector to be deleted is 
\begin{equation}
\argmax_{1\leq i \leq n}~g(p,w_i)  
\end{equation} 
In addition, 
there may exist several solutions in the population having the same largest number of weight vectors.
For this case, 
we compare their worst weight vectors --- 
the weight vector having the highest (worst) scalarizing function value will be deleted.

The above deletion operation is repeated until the number of the weight vectors restores (i.e., back to $N$).
However, 
there may exist one situation that even when every solution in the population corresponds to 
only one weight vector, 
the number of the weight vectors still exceeds $N$.
In this situation,
we use the same diversity maintenance method of Section~III-B to iteratively delete the most crowded solution 
(along with its weight vector) in the population until the number of the weight vectors reduces to $N$.

\subsection{Weight Update Frequency}

The timing and frequency of updating the weight vectors of the evolutionary population 
play an important role in weight vector adaptation methods.
Since varying the weight vectors essentially changes the subproblems to be optimised,
a frequent change can significantly affect the convergence of the algorithm~\cite{Giagkiozis2013b}.
In AdaW, 
the weight update operation is conducted every $5\%$ of the total generations/evaluations. 
In addition, 
when the algorithm approaches the end of the optimisation process, 
a change of the weight vectors may lead to the solutions evolving 
insufficiently along those specified search directions (weight vectors).
Therefore, 
AdaW does not change the weight vectors during the last $10\%$ generations/evaluations.

\renewcommand{\algorithmiccomment}[1]{$/^{*}$ #1 $^{*}/$ }
{\renewcommand\baselinestretch{1.0}\selectfont
	\begin{algorithm}[tp]
		\caption{The Algorithm AdaW}
		\label{alg:algorithm}
		\begin{algorithmic}[1]
			\begin{small}
				\REQUIRE{$N$ (size of the evolutionary population $P$, i.e., size of the weight vector set $W$), 
					$N_A$ (size of the archive set $A$),
				$T$ (neighbourhood size), $Gen_{max}$ (maximum generations in the evolution).}
				\STATE Initialise the population $P$ and
				a set of weight vectors $W$.
			    \STATE Calculate the reference point according to $P$.
				\STATE Determine the neighbours of each weight vector of $W$.
				\STATE Associate the weight vectors with the solutions in the population randomly.
				\STATE Put the nondominated solutions of $P$ into the archive $A$.
				\STATE $Gen \leftarrow 1$.
				\WHILE {$Gen < Gen_{max}$}
				\FOR {each subproblem (weight vector) $w\in W$}
				\STATE Determine the mating pool by selecting the solutions associated with the neighbours of $w$ or from the whole population in a probability.
				\STATE Generate a new solution $p$ by using the variation operators on the solutions in the mating pool.
				\STATE Update the reference point by $p$.
				\STATE Update by $p$ the solutions associated with the neighbours of $w$ or from the whole population in a probability.
				\IF {$\nexists q\in A, q\prec p$}
				\STATE $A \leftarrow A \cup p$
				\STATE $A \leftarrow A/\{q\in A~|~p\prec q\}$
				\ENDIF
				\ENDFOR
				\IF {$|A| > N_A$}
				\STATE Maintain the archive $A$ (Section~III-B).
				\ENDIF
				\IF {$Gen = Gen_{max} \times 5\% \wedge Gen < Gen_{max} \times 90\%$}
				\STATE Generate and find the promising, undeveloped weight vectors, and add them into $W$ (Section~III-C and Section~III-D).
				\STATE Delete the poorly-performed weight vectors and/or the weight vectors associated with the crowded solutions until the size of $W$ is reduced to $N$ (Section~III-E).
				\STATE Update the neighbours of each weight vector of $W$.
				\ENDIF
				\STATE $Gen \leftarrow Gen + 1$.
				\ENDWHILE
				\RETURN {$P$}
			\end{small}
		\end{algorithmic}
	\end{algorithm}
\par}

\subsection{Algorithm Framework}

Algorithm~\ref{alg:algorithm} gives the main procedure of AdaW. 
As can be seen, 
apart from the weight vector update (Steps 21--25) and archive operations (Steps 5, 13--16 and 18--20), 
the remaining steps are the common steps in a generic decomposition-based algorithm. 
Here, 
we implemented them by a widely-used MOEA/D version in \cite{Li2009a}. 
Next, 
we analyse the time complexity of the proposed algorithm.

Additional computational costs of AdaW (in comparison with the basic MOEA/D) 
are from the archive operations and weight vector update.
In one generation of AdaW, 
updating the archive (Steps 13--16) requires $O(mNN_A)$ comparisons, 
where $m$ is the number of the problem's objectives, 
$N$ is the population size, 
and $N_A$ is the archive size.
Maintaining the archive (Steps 18--20) requires $O(mN_A^2)$ comparisons~\cite{Li2016}.
The computational cost of the weight vector update is governed by three operations, 
weight vector addition (Step 22), 
weight vector deletion (Step 23),
and neighbouring weight vector update (Step 24). 
In the weight vector addition,
undeveloped solutions are first determined. 
This includes calculating the niche radius and finding out undeveloped solutions, 
which require $O(mN_A^2)$ and $O(mNN_A)$ comparisons, 
respectively.
After $L$ undeveloped solutions are found, 
we check if they are promising by comparing them with the solutions 
that their neighbouring weight vectors corresponding to.
The computational complexity of finding the neighbours of the $L$ weight vectors is 
bounded by $O(mLN)$ or $O(TLN)$ ($T$ denotes the neighbourhood size), 
whichever is greater.
Then,
checking if these $L$ solutions are promising requires $O(mTL)$ comparisons.
In the weight vector deletion, 
considering the situation that one solution shared by multiple weight vectors 
requires $O(LN)$ comparisons and   
removing the weight vectors which are associated with crowded solutions requires 
$O(m(L+N)^2)$ comparisons~\cite{Li2016}.
Finally, 
after the weight vector deletion completes, 
updating the neighbours of each weight vector in the population requires 
$O(mN^2)$ or $O(TN^2)$ comparisons, 
whichever is greater. 

To sum up, 
since $O(N) = O(N_A)$ and $0 \leqslant L \leqslant N_A$, 
the additional computational cost of AdaW is bounded by $O(mN^2)$ or $O(TN^2)$ whichever is greater, 
where $m$ is the number of objectives and $T$ is the neighbourhood size.
This governs the proposed algorithm given a lower time complexity ($O(mTN)$) 
required in the basic MOEA/D \cite{Zhang2007}.

\section{Results}

Three state-of-the-art weight vector adaptation approaches, 
A-NSGA-III~\cite{Jain2014}, 
RVEA~\cite{Cheng2016}
and MOEA/D-AWA~\cite{Qi2014}, 
along with the baseline MOEA/D~\cite{Li2009a},
were considered as peer algorithms\footnote{The codes of all the peer algorithms were from \url{http://bimk.ahu.edu.cn/index.php?s=/Index/Software/index.html}~\cite{Tian2017}.} 
to evaluate the proposed AdaW.
These adaptations had been demonstrated to be competitive on MOPs with various Pareto fronts.
In MOEA/D, 
the Tchebycheff scalarizing function was used in which ``multiplying the weight vector''
was replaced with ``dividing the weight vector'' in order to obtain more uniform solutions~\cite{Qi2014,Deb2014}. 

In view of the goal of the proposed work, 
we selected 17 test problems with a variety of representative Pareto fronts 
from the existing problem suites~\cite{Veldhuizen1999,Zitzler2000,Deb2005a,Deb2005b,Deb2014,Jain2014,Cheng2017}.  
According to the properties of their Pareto fronts, 
we categorised the problems into seven groups to challenge the 
algorithms in balancing the convergence and diversity of solutions.
They are 
\begin{enumerate}
	\item problems with a simple-like Pareto front: DTLZ1, DTLZ2 and convex DTLZ2 (CDTLZ2).
	
	\item problems with an inverted simple-like Pareto front: inverted DTLZ1 (IDTLZ1) and inverted DTLZ2 (IDTLZ2).
	
	\item problems with a highly nonlinear Pareto front: SCH1 and FON.
	
	\item problems with a disconnect Pareto front: ZDT3 and DTLZ7.
	
	\item problems with a degenerate Pareto front: DTLZ5 and VNT2.
	
	\item problems with a badly-scaled Pareto front: scaled DTLZ1 (SDTLZ1), scaled DTLZ2 (SDTLZ2) and SCH2.
	
	\item problems with a high-dimensional Pareto front: 10-objective DTLZ2 (DTLZ2-10), 10-objective inverted DTLZ1 (IDTLZ1-10) and DTLZ5(2,10). 
\end{enumerate}
All the problems were configured as described in their original papers~\cite{Veldhuizen1999,Zitzler2000,Deb2005a,Deb2005b,Deb2014,Jain2014,Cheng2017}.

To compare the performance of the algorithms, 
the inverted generational distance (IGD)~\cite{Coello2004a,Zhang2007} was used. 
IGD, 
which measures the average distance from uniformly distributed points along the Pareto front to 
their closest solution in a set,
can provide a comprehensive assessment of the convergence and diversity of the set. 
In addition, 
for a visual understanding of the search behaviour of the five algorithms,
we also plotted their final solution set in a single run on all the test problems.
This particular run was associated with the solution set 
which obtained the median of the IGD values out of all the runs.  

All the algorithms were given real-valued variables. 
Simulated binary crossover (SBX) \cite{Agrawal1995} and polynomial mutation (PM) \cite{Deb2001}
(with the distribution indexes 20) were used to perform the variation.
The crossover probability was set to $p_c = 1.0$ and mutation probability to $p_m = 1/d$,
where $d$ is the number of variables in the decision space.

In decomposition-based EMO, 
the population size which correlates with the number of the weight vectors cannot be set arbitrarily.
For a set of uniformly-distributed weight vectors in a simplex, 
we set 100, 105 and 220 for the 2-, 3- and 10-objective problems, 
respectively.
Like many existing studies, 
the number of function evaluations was set to 25,000, 30,000 and 100,000 for 2-, 3- and 10-objective problems,
respectively.
Each algorithm was executed 30 independent runs on each problem.

Parameters of the peer algorithms were set as specified/recommended in their original papers. 
In MOEA/D, 
the neighbourhood size, 
the probability of parent solutions selected from the neighbours, 
and the maximum number of replaced solutions were set to $10\%$ of the population size,
0.9, 
and $1\%$ of the population size, 
respectively.
In RVEA, 
the rate of changing the penalty function and the frequency to conduct the reference vector
adaptation were set to $2$ and $0.1$, 
respectively.
In MOEA/D-AWA, 
the maximal number of adjusting subproblems and the computational resources for the weight vector adaptation were
set to $0.05N$ and $20\%$, 
respectively.
In addition, 
the size of the external population in MOEA/D-AWA was set to $1.5N$.

Several specific parameters are required in the proposed AdaW. 
As stated in Section~III-F, 
the time of updating the weight vectors and the time of not allowing the update were 
every $5\%$ of the total generations and the last $10\%$ generations, 
respectively. 
Finally,
the maximum capacity of the archive was set to $2N$.

\mbox{Tables~\ref{Table:IGDresults}} gives the IGD results (mean and standard deviation) 
of the five algorithms on all the 17 problems. 
The better mean for each problem was highlighted in boldface. 
To have statistically sound conclusions, 
the Wilcoxon's rank sum test
\cite{Zitzler2008} at a 0.05 significance level was used to test the significance
of the differences between the results obtained by AdaW and the four peer algorithms.

\begin{sidewaystable}[tbp]
\begin{center}
	\caption{IGD results (mean and SD) of the five algorithms. 
		The better mean for each case is highlighted in boldface.}
	\label{Table:IGDresults}
	\scriptsize
	\begin{tabular}{@{}c@{}|@{}c@{}|l|l|l|l|l@{}}
		\hline
		\multicolumn{1}{c|}{Property} & \multicolumn{1}{c|}{Problem} & \multicolumn{1}{c|}{MOEA/D} & \multicolumn{1}{c|}{A-NSGA-III} & \multicolumn{1}{c|}{RVEA} & \multicolumn{1}{c|}{MOEA/D-AWA} & \multicolumn{1}{c}{AdaW} \\ \hline
		\multirow{3}[0]{*}{Simplex-like} & DTLZ1 & \textbf{1.909E--02(3.1E--04)}$^\dagger$ & 2.463E--02(8.0E--03)$^\dagger$ & 1.974E--02(2.2E--03) & 1.941E--02(6.1E--04) & 1.944E--02(3.1E--04) \\
		& DTLZ2 & 5.124E--02(4.6E--04) & 5.222E--02(1.4E--03)$^\dagger$ & \textbf{5.020E--02(7.3E--05)}$^\dagger$ & 5.070E--02(3.8E--04)$^\dagger$ & 5.126E--02(6.0E--04)\\
		& CDTLZ2 & 4.388E--02(1.0E--04)$^\dagger$ & 8.766E--02(2.8E--02)$^\dagger$ & 4.198E--02(1.4E--03)$^\dagger$ & 3.879E--02(3.2E--03)$^\dagger$ & \textbf{2.852E--02(5.9E--04)} \\ \hline
		
		\multirow{2}[0]{*}{Inverted simplex-like} & IDTLZ1 & 3.175E--02(7.9E--04)$^\dagger$ & 2.091E--02(1.5E--03)$^\dagger$ & 6.404E--02(4.6E--02)$^\dagger$ & 2.698E--02(6.2E--04)$^\dagger$ & \textbf{1.961E--02(4.8E--04)} \\
		& IDTLZ2 & 9.010E--02(1.5E--04)$^\dagger$ & 7.200E--02(6.7E--03)$^\dagger$ & 7.736E--02(1.7E--03)$^\dagger$ & 7.166E--02(5.2E--03)$^\dagger$ & \textbf{5.037E--02(6.2E--04)} \\ \hline
		
		\multirow{2}[0]{*}{Highly nonlinear} & SCH1 & 4.835E--02(1.7E--03)$^\dagger$ & 5.411E--02(9.7E--03)$^\dagger$ & 4.643E--02(4.1E--03)$^\dagger$ & 2.604E--02(3.6E--03)$^\dagger$ & \textbf{1.703E--02(1.5E--04)}\\
		& FON & \textbf{4.596E--03(1.6E--05)}$^\dagger$ & 5.333E--03(4.5E--04)$^\dagger$ & 5.161E--03(1.8E--04)$^\dagger$ & 4.739E--03(5.2E--05)$^\dagger$ & 4.632E--03(8.3E--05)\\ \hline
		
		\multirow{2}[0]{*}{Disconnect} 
		& ZDT3 & 1.107E--02(5.1E--04)$^\dagger$ & 3.735E--02(4.1E--02)$^\dagger$ & 9.128E--02(4.2E--02)$^\dagger$ & 3.125E--02(5.1E--02)$^\dagger$ & \textbf{4.840E--03(5.6E--04)} \\
		& DTLZ7 & 1.297E--01(1.1E--03)$^\dagger$ & 7.079E--02(2.3E--03)$^\dagger$ & 1.012E--01(4.6E--03)$^\dagger$ & 1.318E--01(9.0E--02)$^\dagger$ & \textbf{5.275E--02(6.0E--04)} \\ \hline
		
		\multirow{2}[0]{*}{Degenerate} & DTLZ5 & 1.811E--02(1.0E--05)$^\dagger$ & 9.759E--03(1.2E--03)$^\dagger$ & 6.816E--02(5.3E--03)$^\dagger$ & 9.584E--03(2.9E--04)$^\dagger$ & \textbf{3.976E--03(2.4E--04)}\\
		& VNT2 & 4.651E--02(2.7E--04)$^\dagger$ & 2.143E--02(3.2E--03)$^\dagger$ & 3.492E--02(4.6E--03)$^\dagger$ & 1.961E--02(7.4E--04)$^\dagger$ & \textbf{1.155E--02(2.3E--04)} \\ \hline
		
		\multirow{3}[0]{*}{Badly scaled} & SDTLZ1 & 5.584E+00(2.0E+00)$^\dagger$ & 7.426E--01(4.2E--02)$^\dagger$ & 1.522E+00(2.0E+00)$^\dagger$ & 2.988E+00(5.0E--01)$^\dagger$ & \textbf{6.571E--01(6.0E--02)}\\
		& SDTLZ2 & 6.071E+00(2.0E+00)$^\dagger$ & 1.357E+00(4.7E--02)$^\dagger$ & 1.295E+00(1.7E--02)$^\dagger$ & 4.176E+00(5.7E--01)$^\dagger$ &  \textbf{1.244E+00(5.2E--02)} \\ 
		& SCH2 & 1.049E--01(2.6E--04)$^\dagger$ & 5.109E--02(4.4E--02)$^\dagger$ & 4.488E--02(3.6E--04)$^\dagger$ & 5.538E--02(3.0E--03)$^\dagger$ & \textbf{2.097E--02(3.1E--04)}\\ \hline
		
		\multirow{3}[0]{*}{Many objectives} 
		& DTLZ2-10 & 5.172E--01(1.4E--02) & 5.314E--01(6.2E--02)$^\dagger$ & \textbf{4.924E--01(2.6E--05)}$^\dagger$ & 5.234E--01(3.1E--02) & 5.202E--01(1.4E--02)\\
		& IDTLZ1-10 & 2.721E--01(7.7E--03)$^\dagger$ & 1.507E--01(6.5E--03)$^\dagger$ & 2.461E--01(9.0E--03)$^\dagger$ & 2.421E--01(9.0E--03)$^\dagger$ & \textbf{1.071E--01(3.3E--03)}\\
		& DTLZ5(2,10) & 1.708E--01(1.6E--03)$^\dagger$ & 4.431E--01(1.0E--01)$^\dagger$ & 1.520E--01(2.3E--02)$^\dagger$ & 3.830E--02(1.3E--02)$^\dagger$ & \textbf{2.150E--03(1.8E--05)} \\ \hline	
	\end{tabular}{}
\end{center}
\scriptsize ``$\dagger$'' indicates that the result of the peer algorithm is significantly
different from that of AdaW at a 0.05 level by the Wilcoxon's rank sum test.
\end{sidewaystable}%

\subsection{On Simplex-like Pareto Fronts}

\begin{figure*}[tbp]
	\begin{center}
		\footnotesize
		\begin{tabular}{@{}c@{}c@{}c@{}c@{}c@{}}
			\includegraphics[scale=0.14]{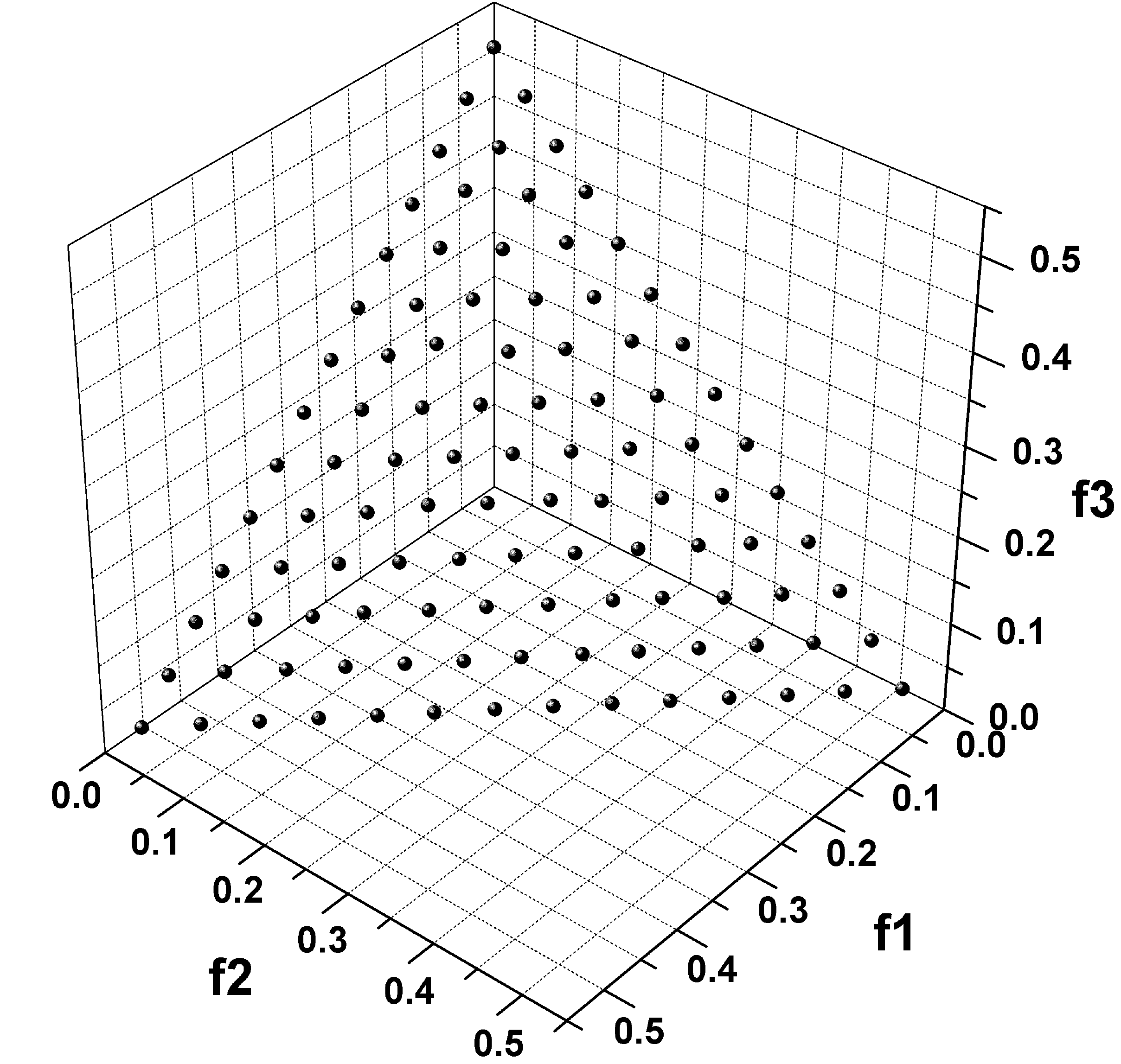}&
			\includegraphics[scale=0.14]{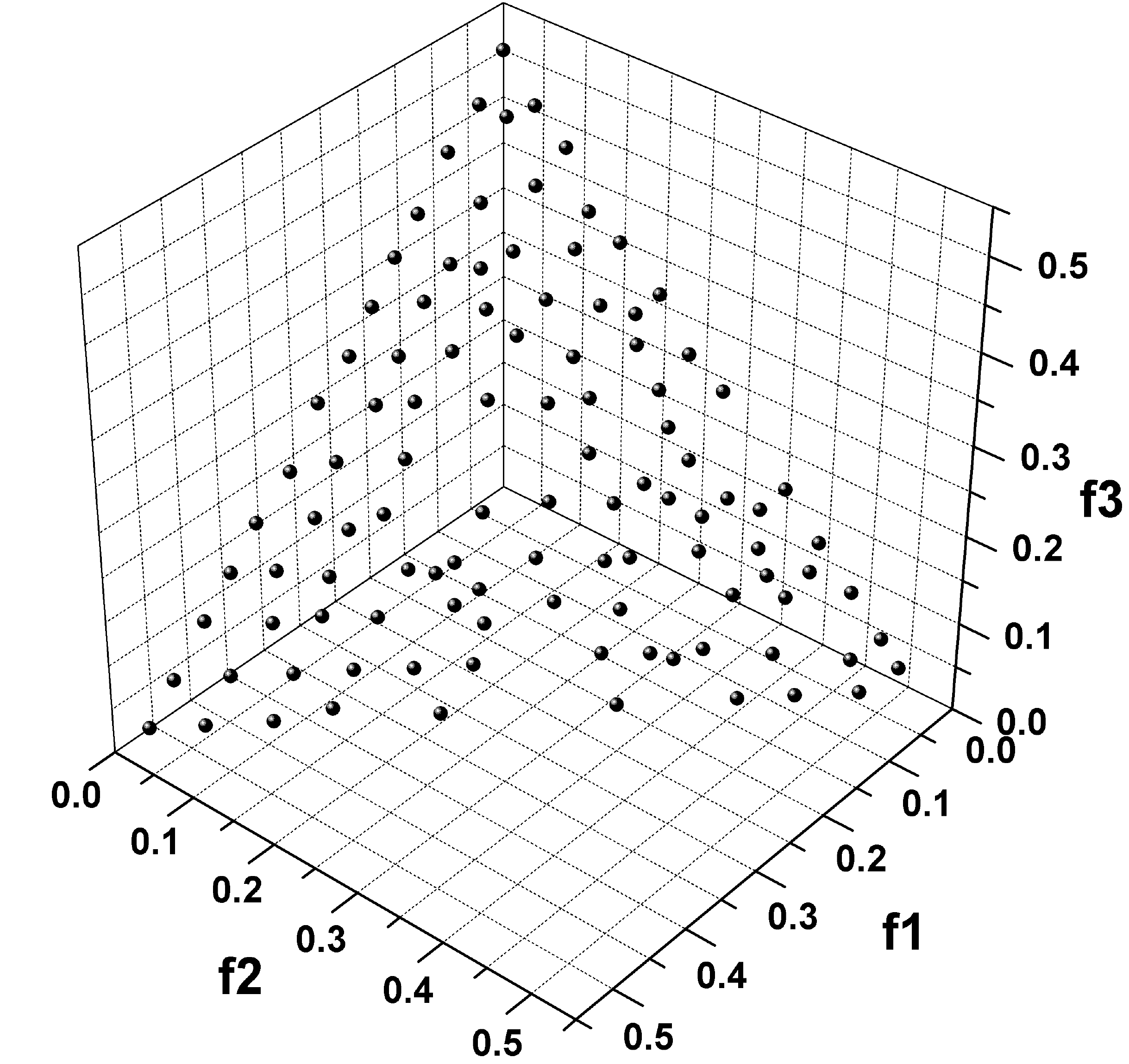}&
			\includegraphics[scale=0.14]{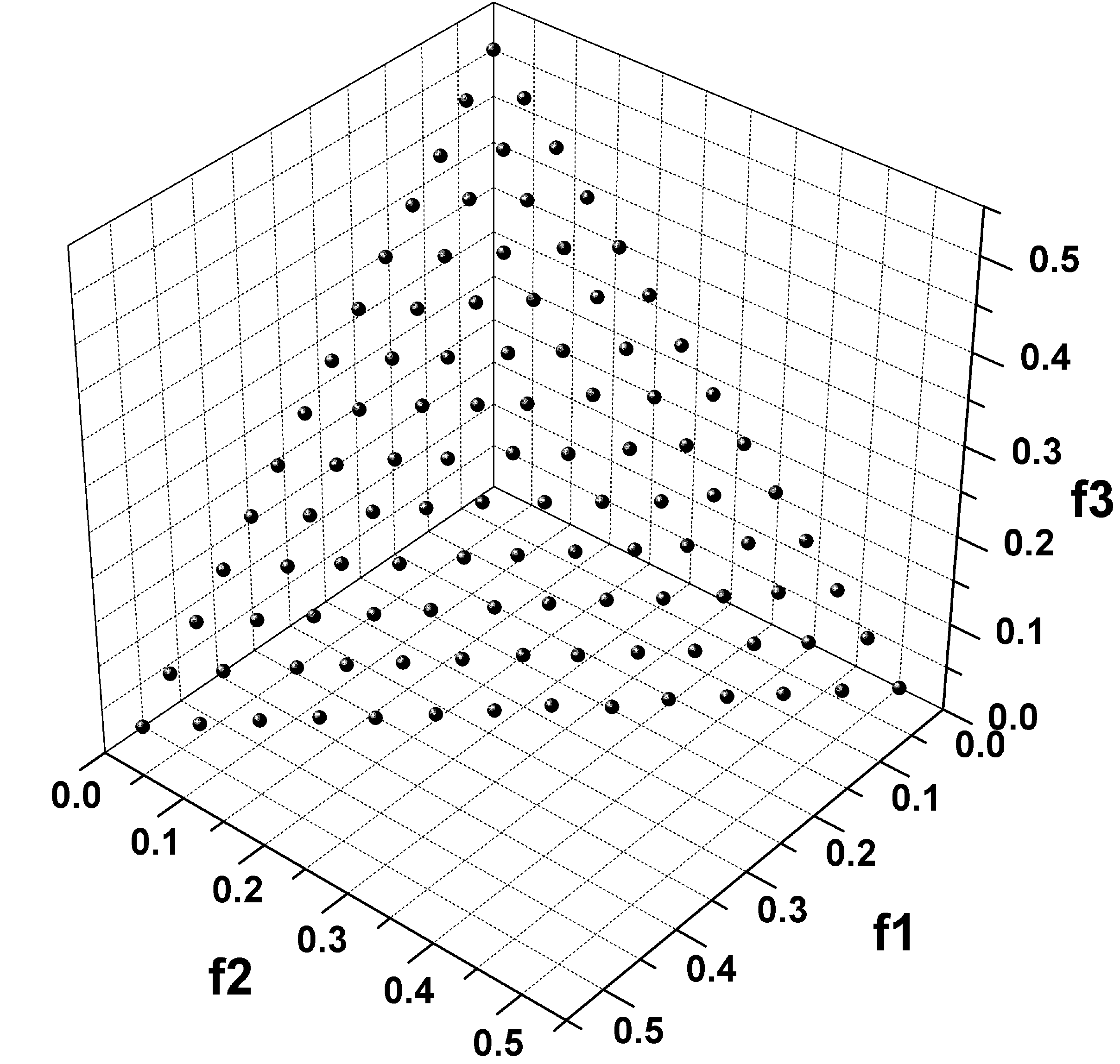}&
			\includegraphics[scale=0.14]{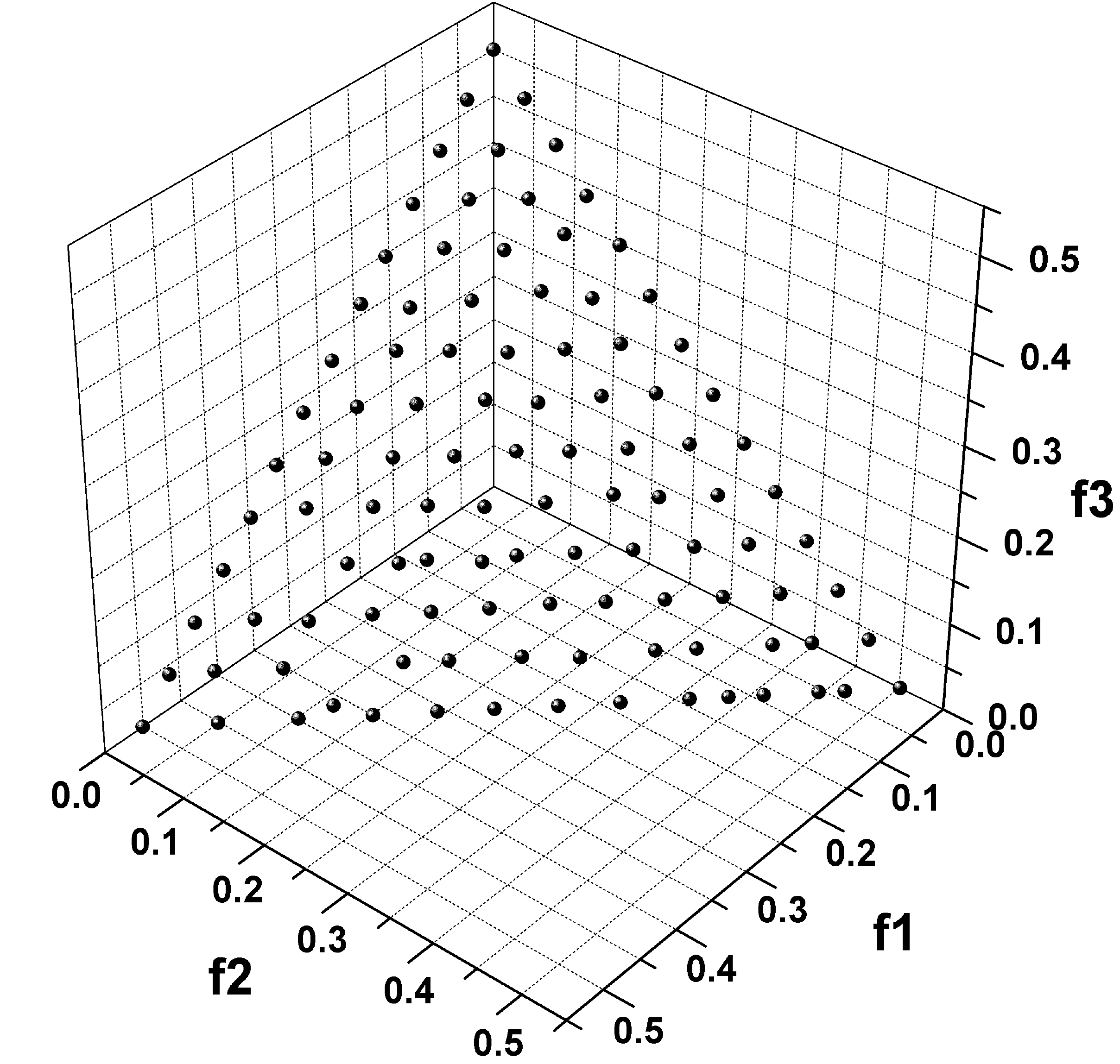}&
			\includegraphics[scale=0.14]{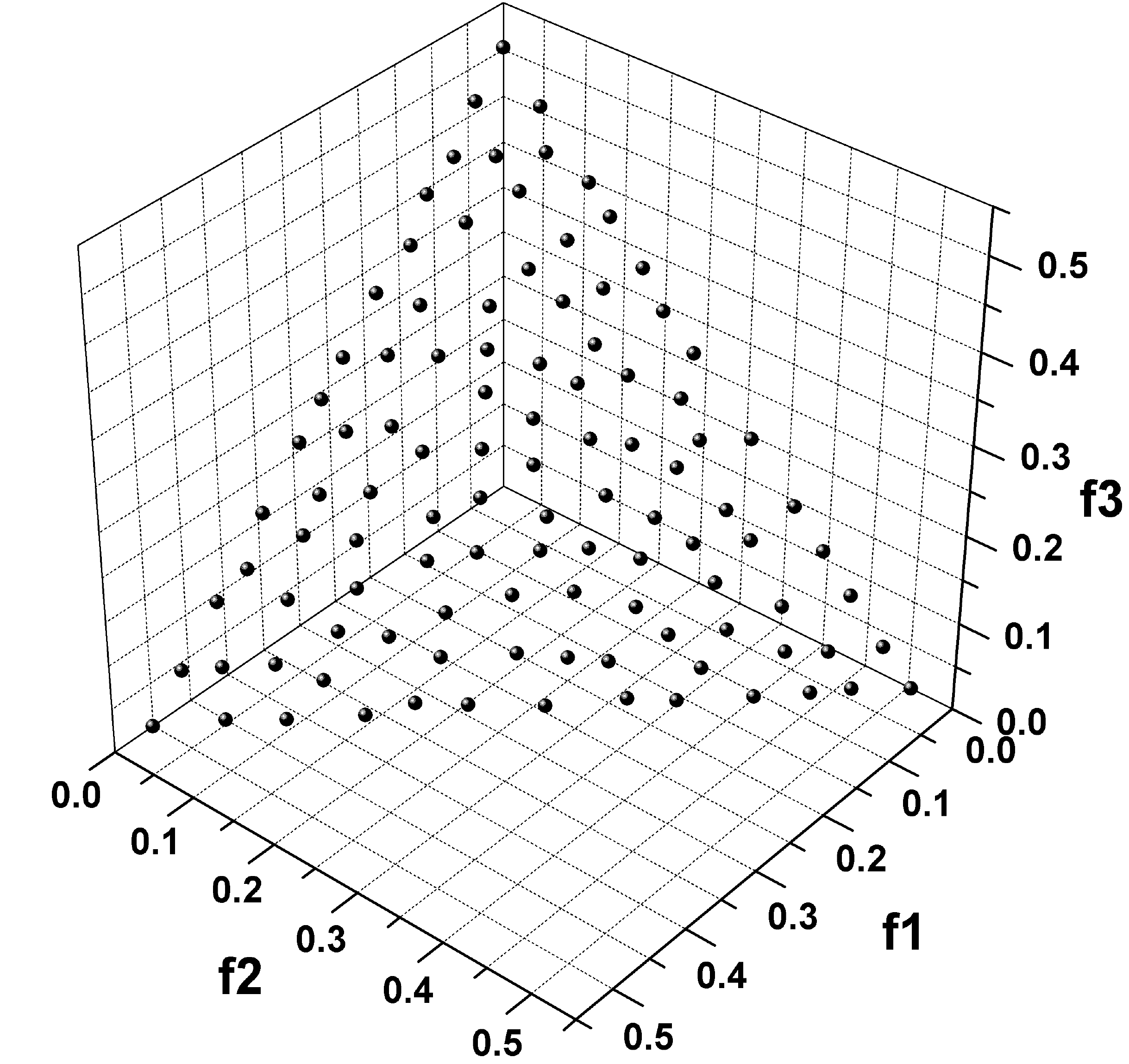}\\
			(a) MOEA/D & (b) A-NSGA-III & (c) RVEA & (d) MOEA/D-AWA & (e) AdaW \\
		\end{tabular}
	\end{center}
	\caption{The final solution set of the five algorithms on DTLZ1.}
	\label{Fig:DTLZ1-3}
\end{figure*}

\begin{figure*}[tbp]
	\begin{center}
		\footnotesize
		\begin{tabular}{@{}c@{}c@{}c@{}c@{}c@{}}
			\includegraphics[scale=0.14]{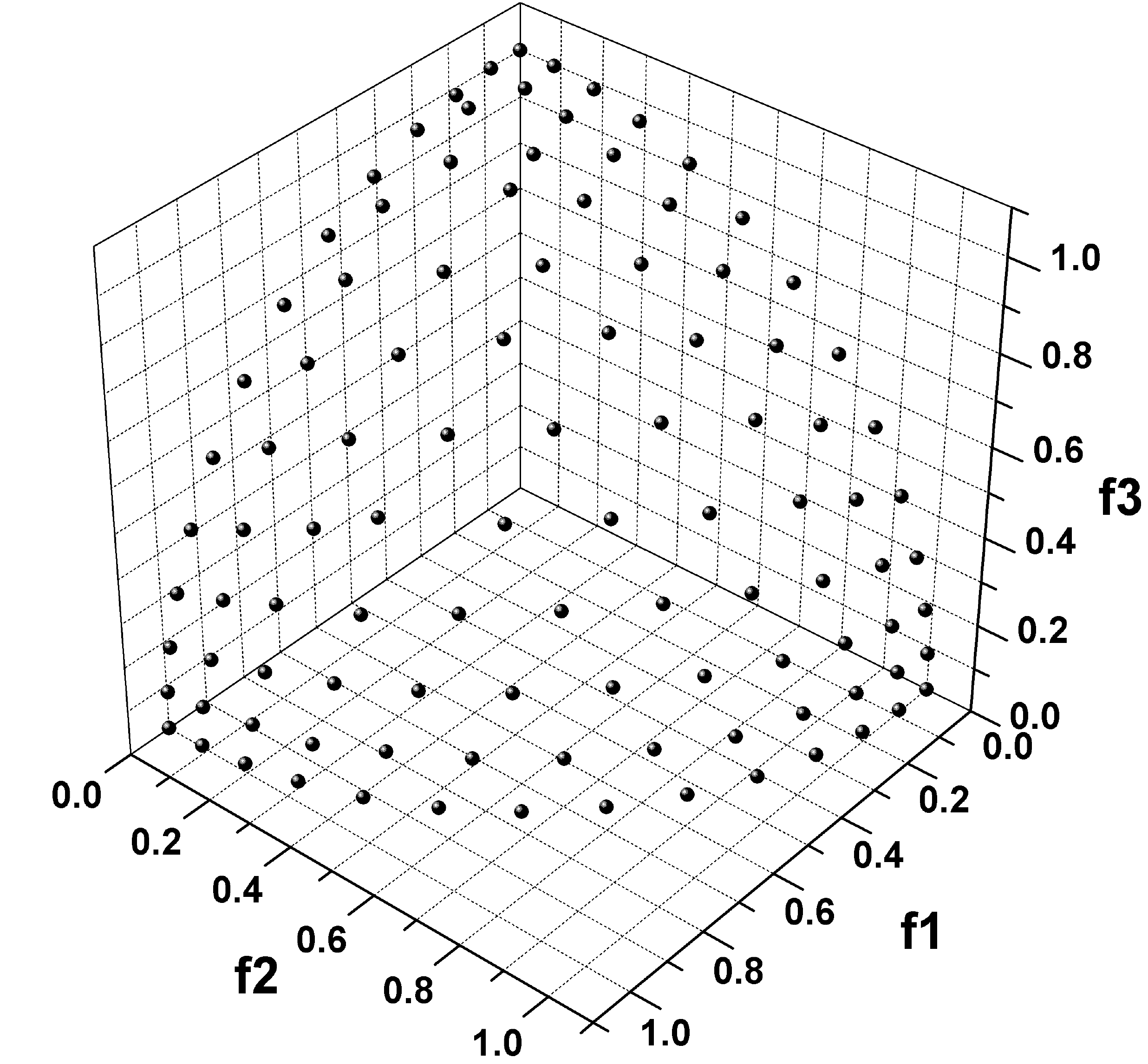}&
			\includegraphics[scale=0.14]{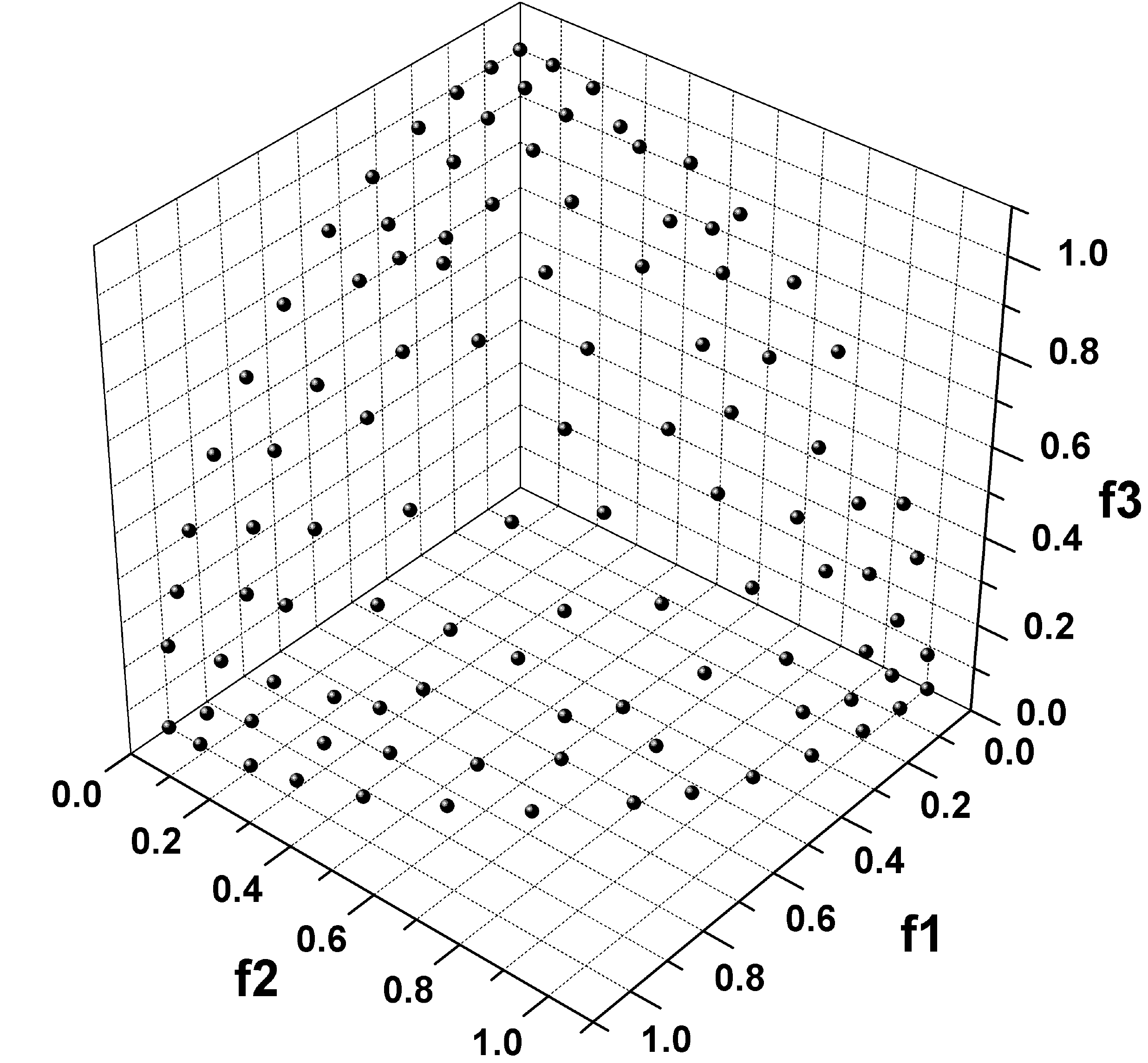}&
			\includegraphics[scale=0.14]{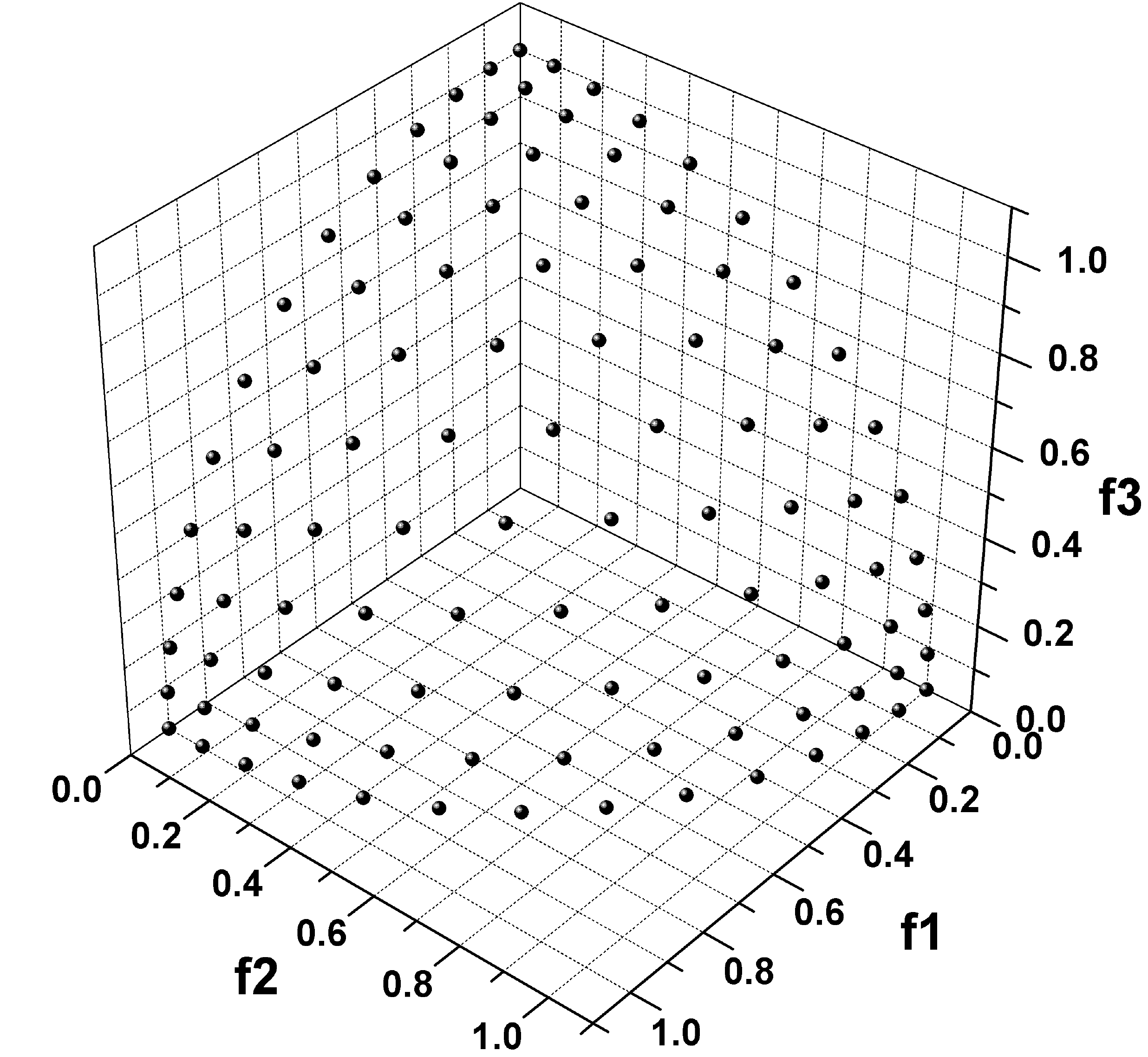}&
			\includegraphics[scale=0.14]{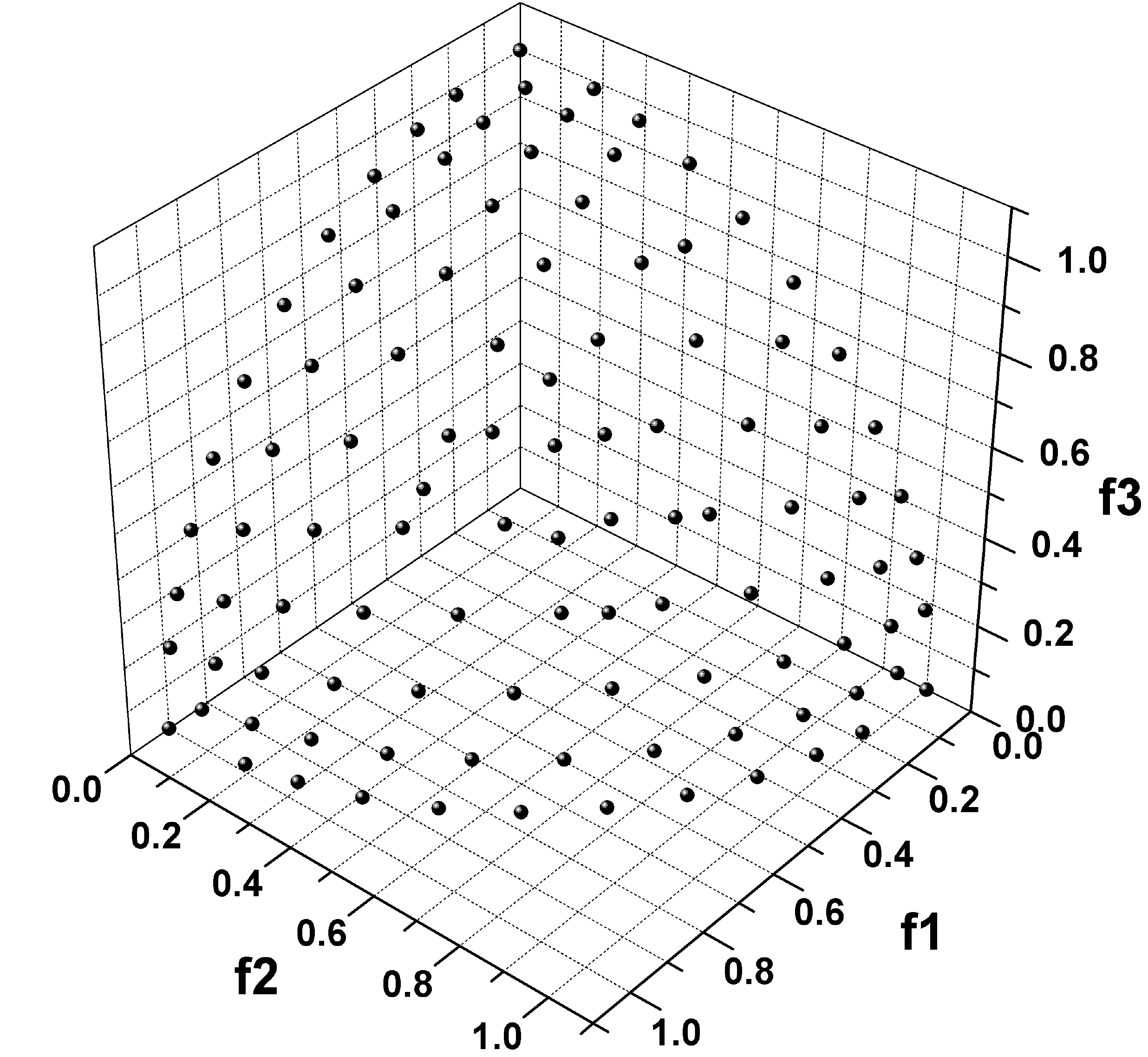}&
			\includegraphics[scale=0.14]{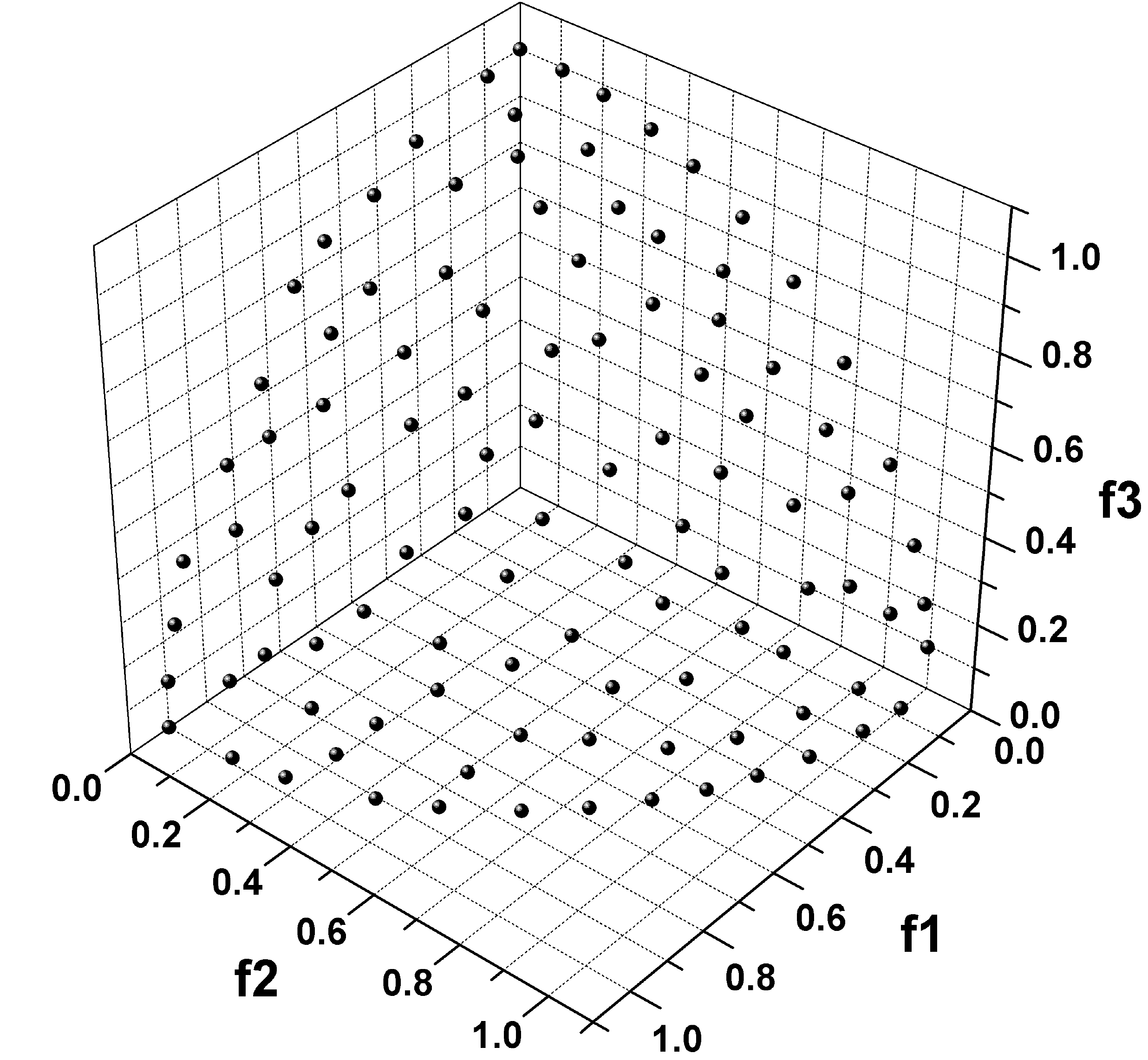}\\
			(a) MOEA/D & (b) A-NSGA-III & (c) RVEA & (d) MOEA/D-AWA & (e) AdaW \\
		\end{tabular}
	\end{center}
	\caption{The final solution set of the five algorithms on DTLZ2.}
	\label{Fig:DTLZ2-3}
\end{figure*}

\begin{figure*}[!]
	\begin{center}
		\footnotesize
		\begin{tabular}{@{}c@{}c@{}c@{}c@{}c@{}}
			\includegraphics[scale=0.14]{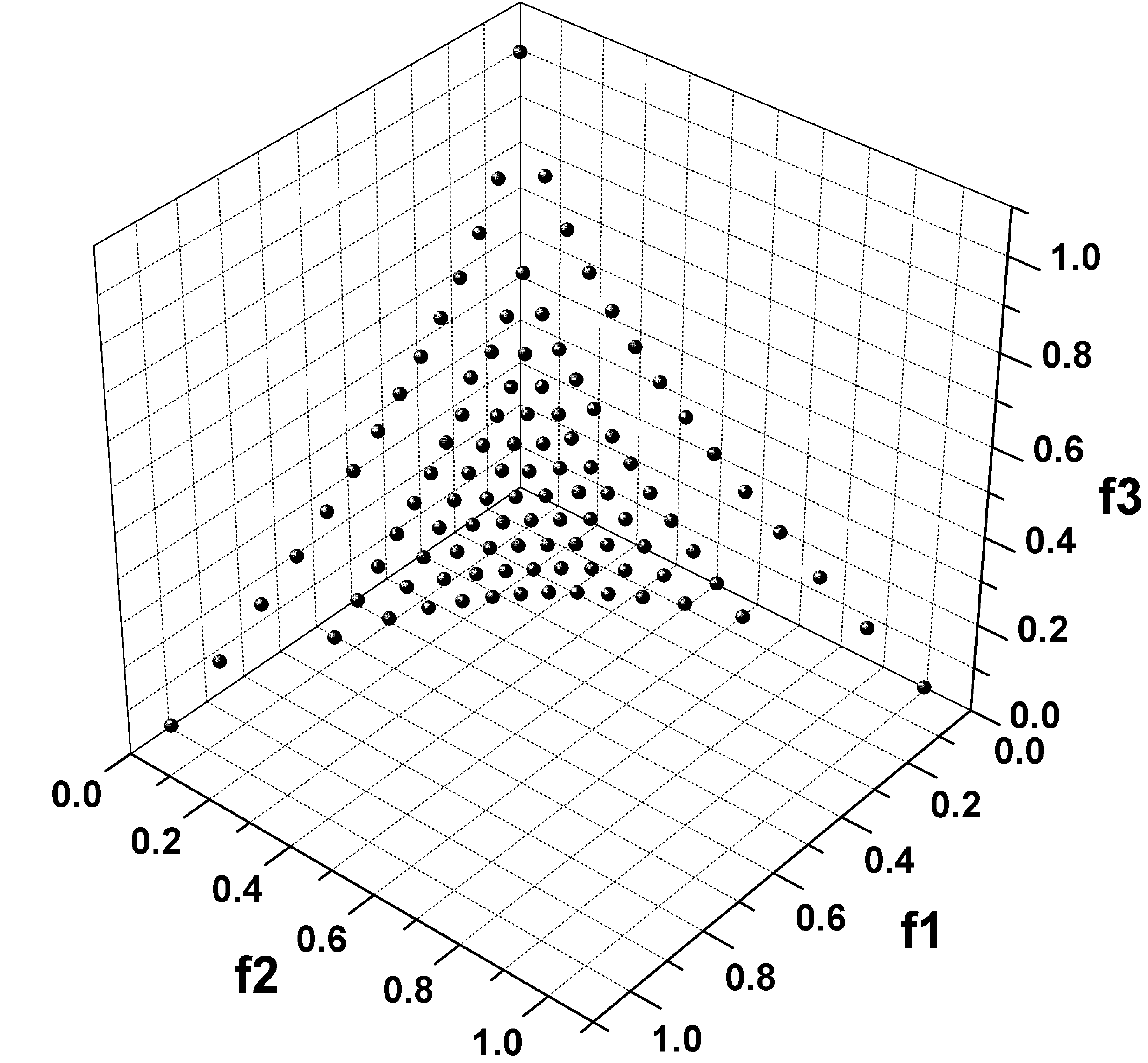}&
			\includegraphics[scale=0.14]{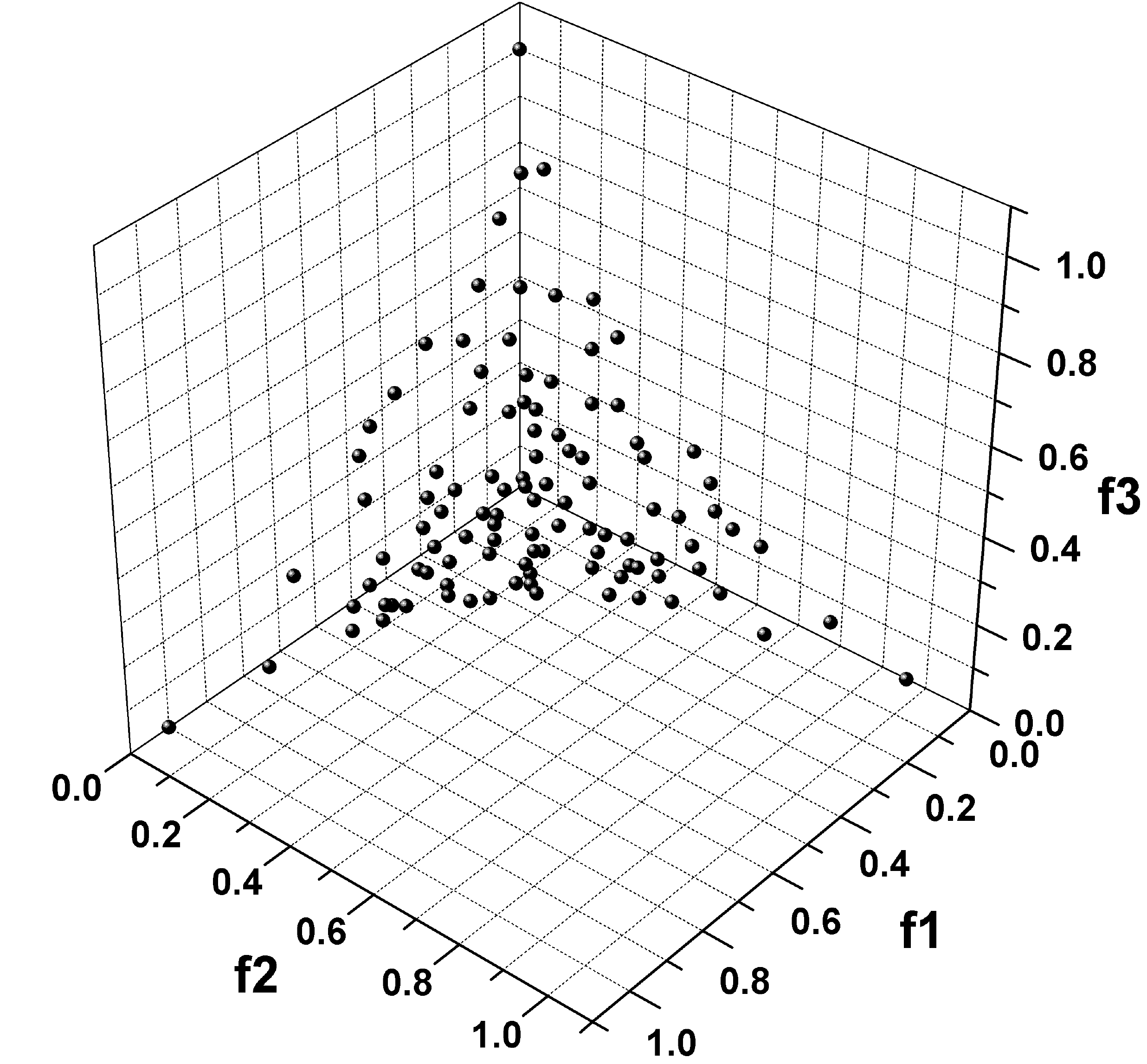}&
			\includegraphics[scale=0.14]{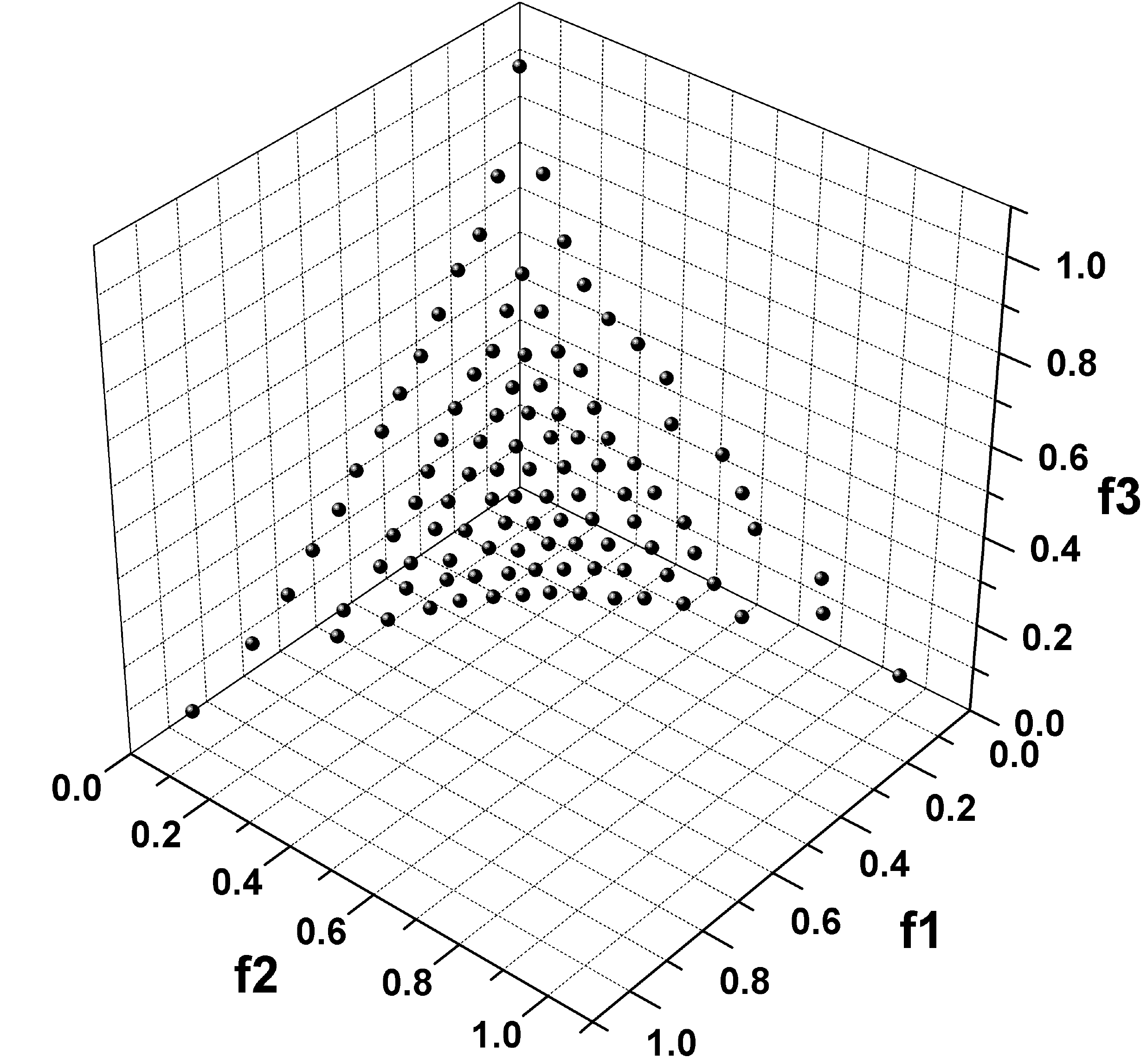}&
			\includegraphics[scale=0.14]{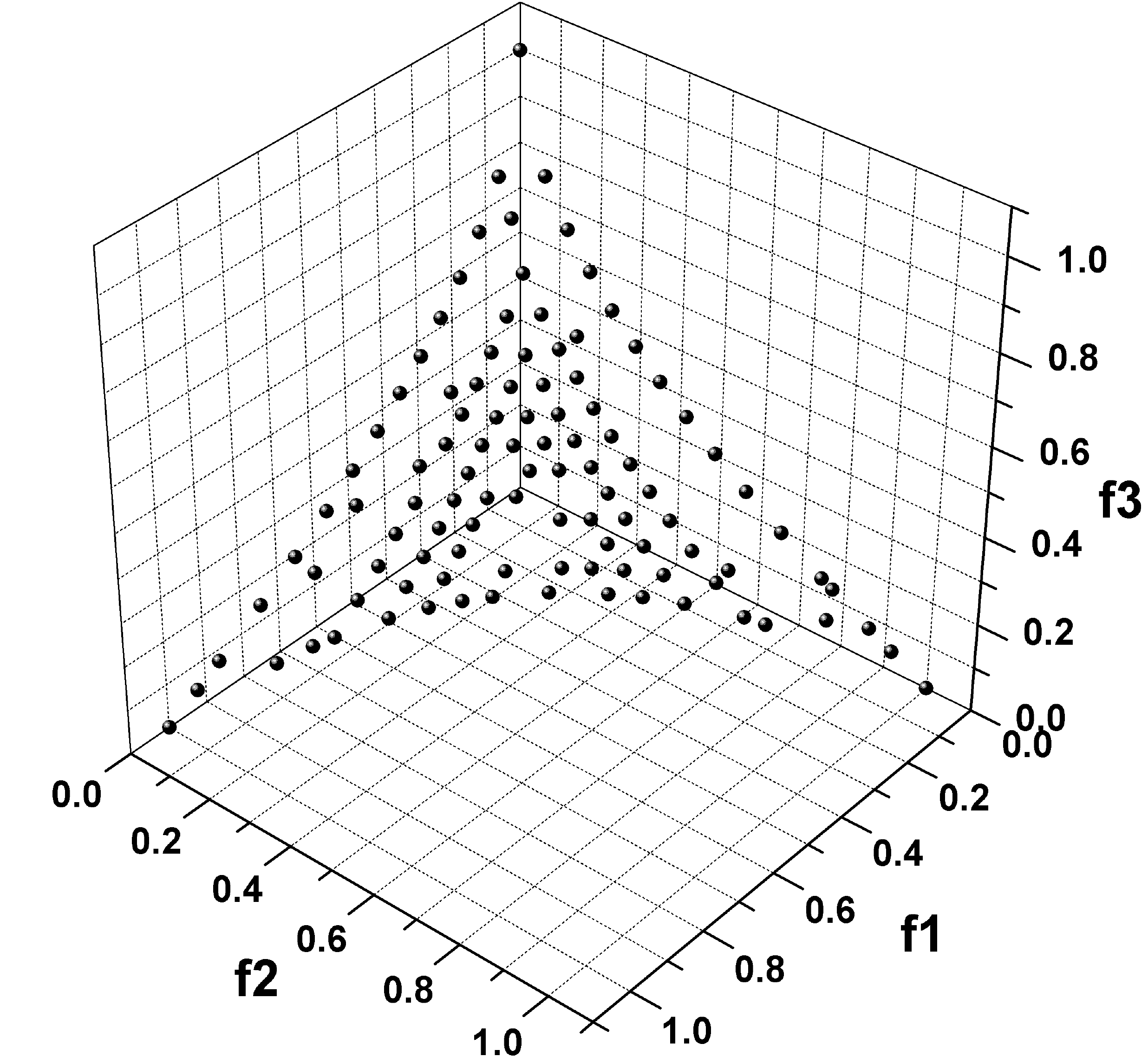}&
			\includegraphics[scale=0.14]{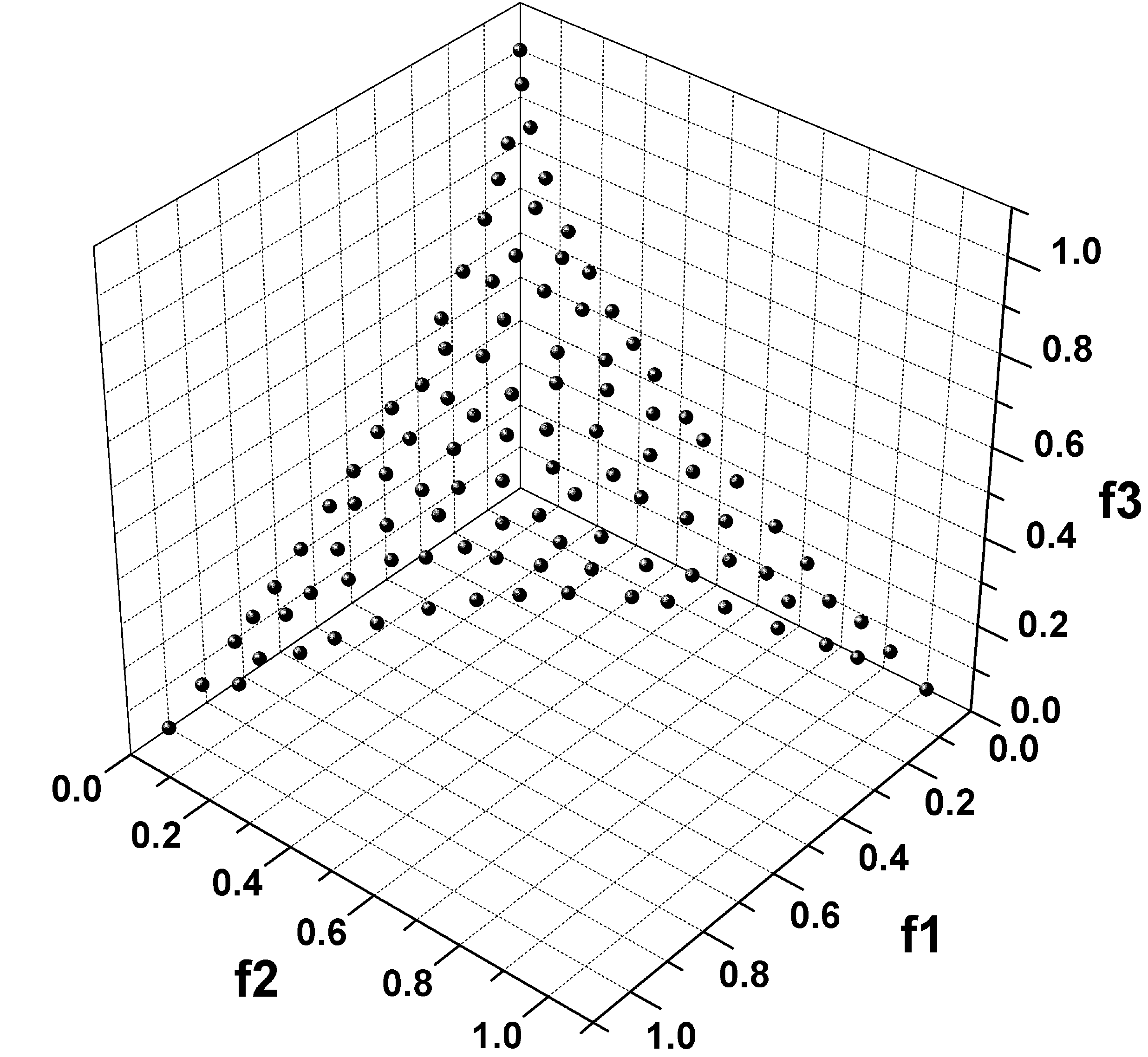}\\
			(a) MOEA/D & (b) A-NSGA-III & (c) RVEA & (d) MOEA/D-AWA & (e) AdaW \\
		\end{tabular}
	\end{center}
	\caption{The final solution set of the five algorithms on the convex DTLZ1.}
	\label{Fig:CDTLZ2-3}
\end{figure*}

On MOPs with a simplex-like Pareto front, 
decomposition-based algorithms are expected to perform well. 
\mbox{Figures~\ref{Fig:DTLZ1-3}--\ref{Fig:CDTLZ2-3}} plot the final solution set 
of the five algorithms on DTLZ1, 
DTLZ2 and CDTLZ2,
respectively.
As can be seen,
MOEA/D, RVEA, MOEA/D-AWA and AdaW can all obtain a well-distributed solution set, 
despite the set of AdaW not being so ``regular'' as that of the other three algorithms.
An interesting observation is that A-NSGA-III 
(adapting the weight vectors in NSGA-III) appears to struggle 
in maintaining the uniformity of the solutions, 
especially for DTLZ1 and CDTLZ2. 
This indicates that adapting the weight vectors may compromise the performance of 
decomposition-based approach on simplex-like Pareto fronts, 
as NSGA-III had been demonstrated to work very well on these three MOPs \cite{Deb2014}. 
In addition,
it is worth mentioning that on the convex DTLZ2 
there is an interval between the outer and inner solutions in the solution sets of MOEA/D, RVEA and MOEA/D-AWA.   
In contrast, 
the proposed AdaW has no such interval, 
thereby returning a better IGD result as shown in \mbox{Table~\ref{Table:IGDresults}}. 


\begin{figure*}[tbp]
	\begin{center}
		\footnotesize
		\begin{tabular}{@{}c@{}c@{}c@{}c@{}c@{}}
			\includegraphics[scale=0.14]{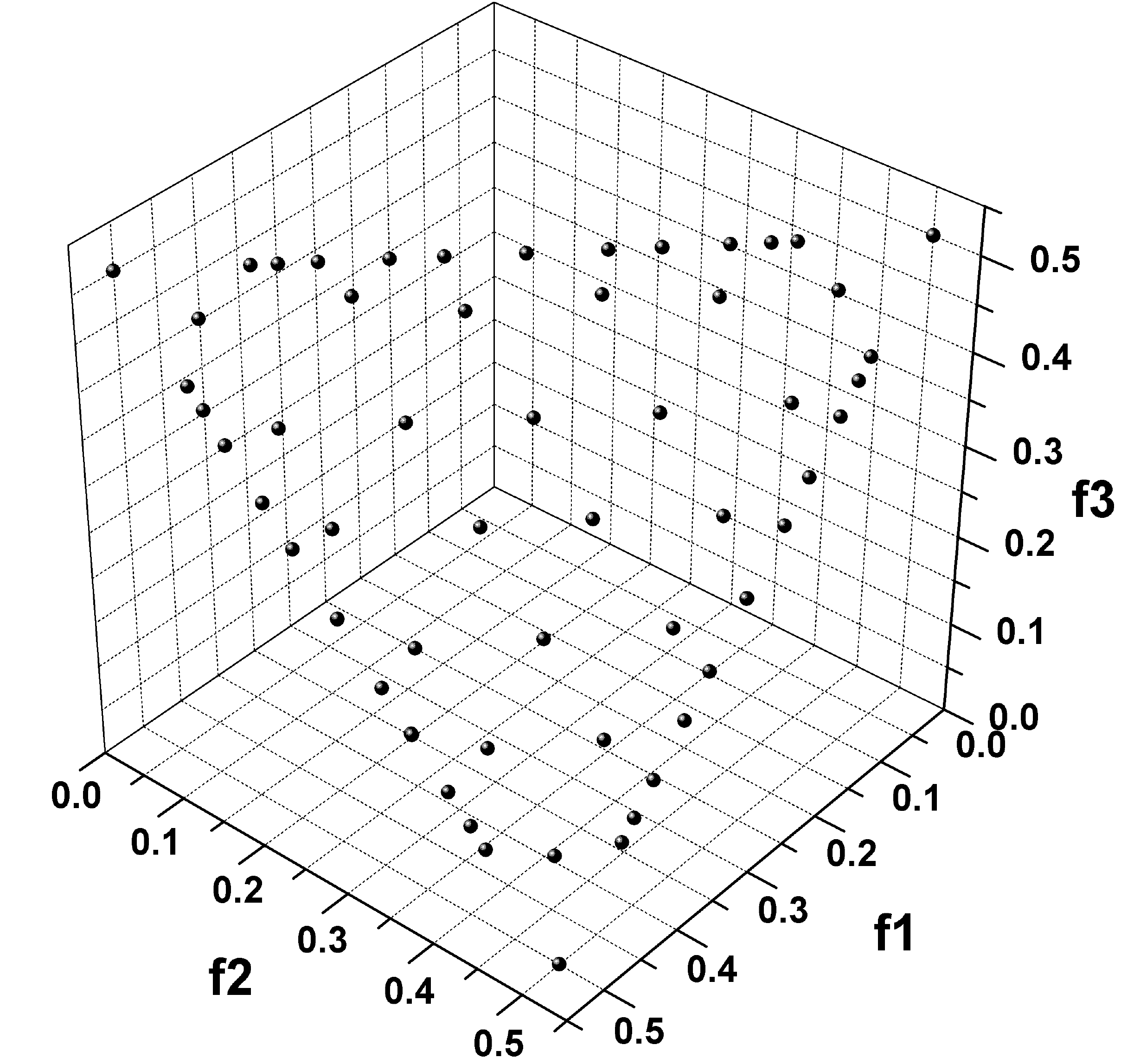}&
			\includegraphics[scale=0.14]{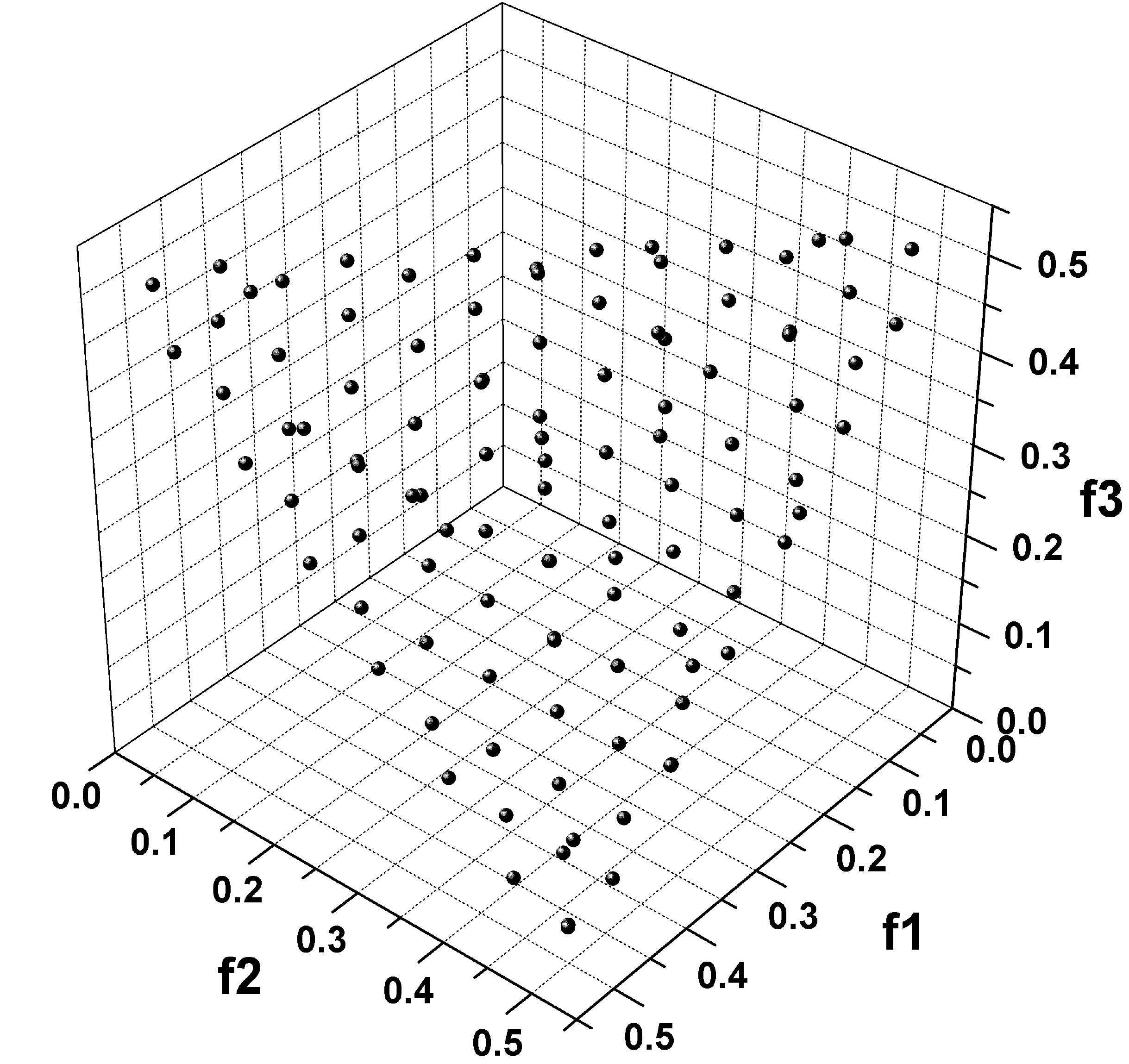}&
			\includegraphics[scale=0.14]{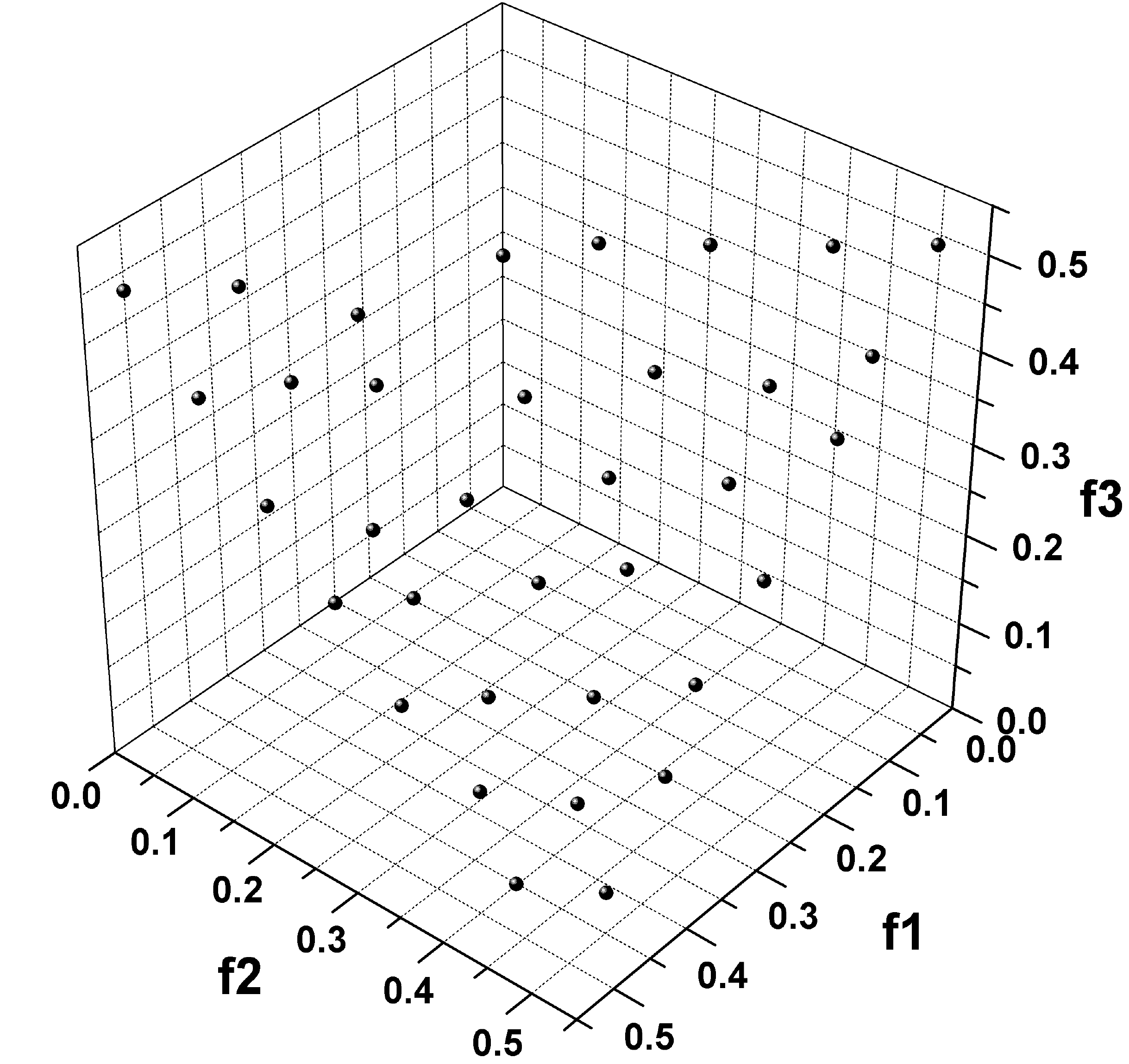}&
			\includegraphics[scale=0.14]{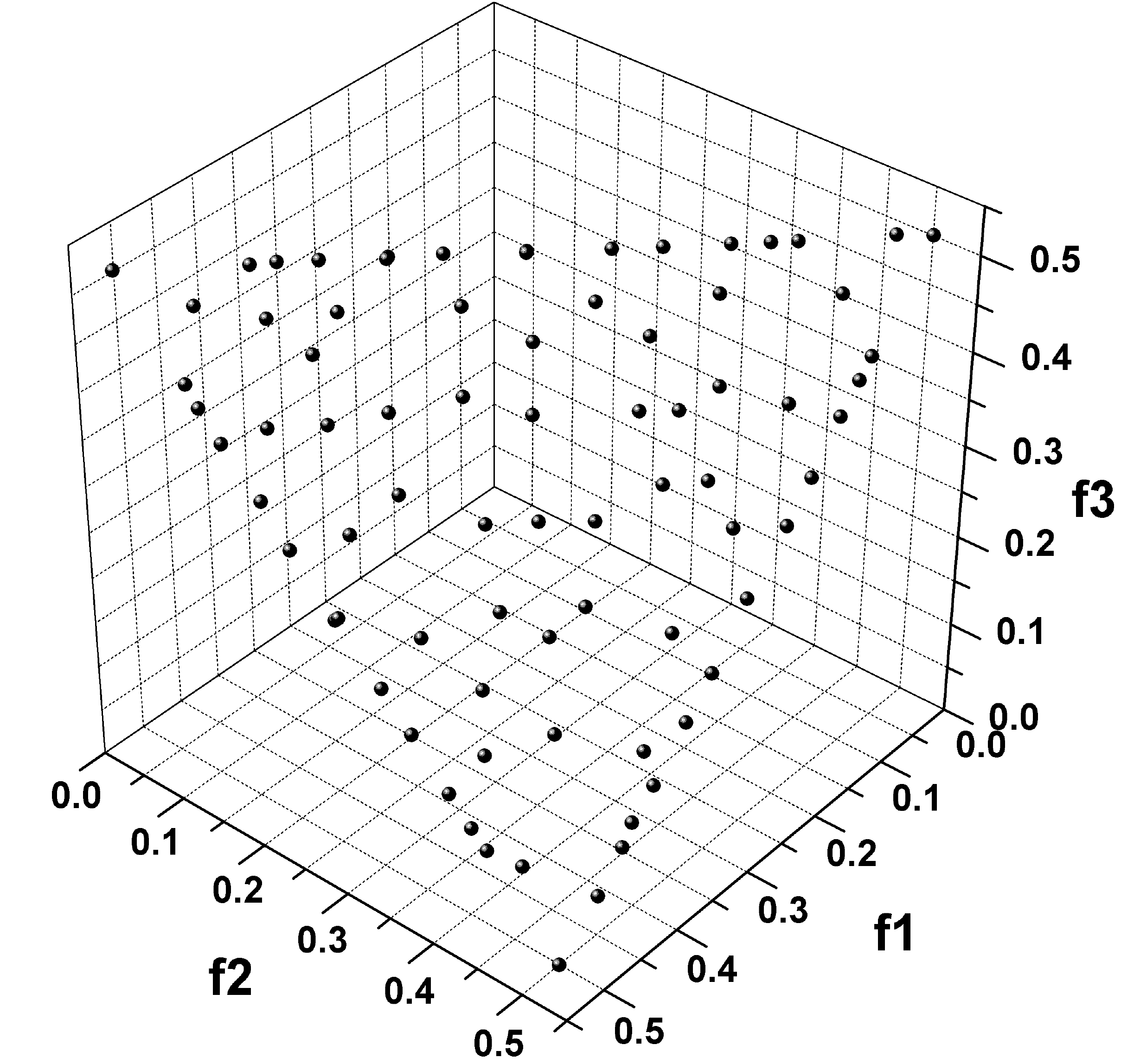}&
			\includegraphics[scale=0.14]{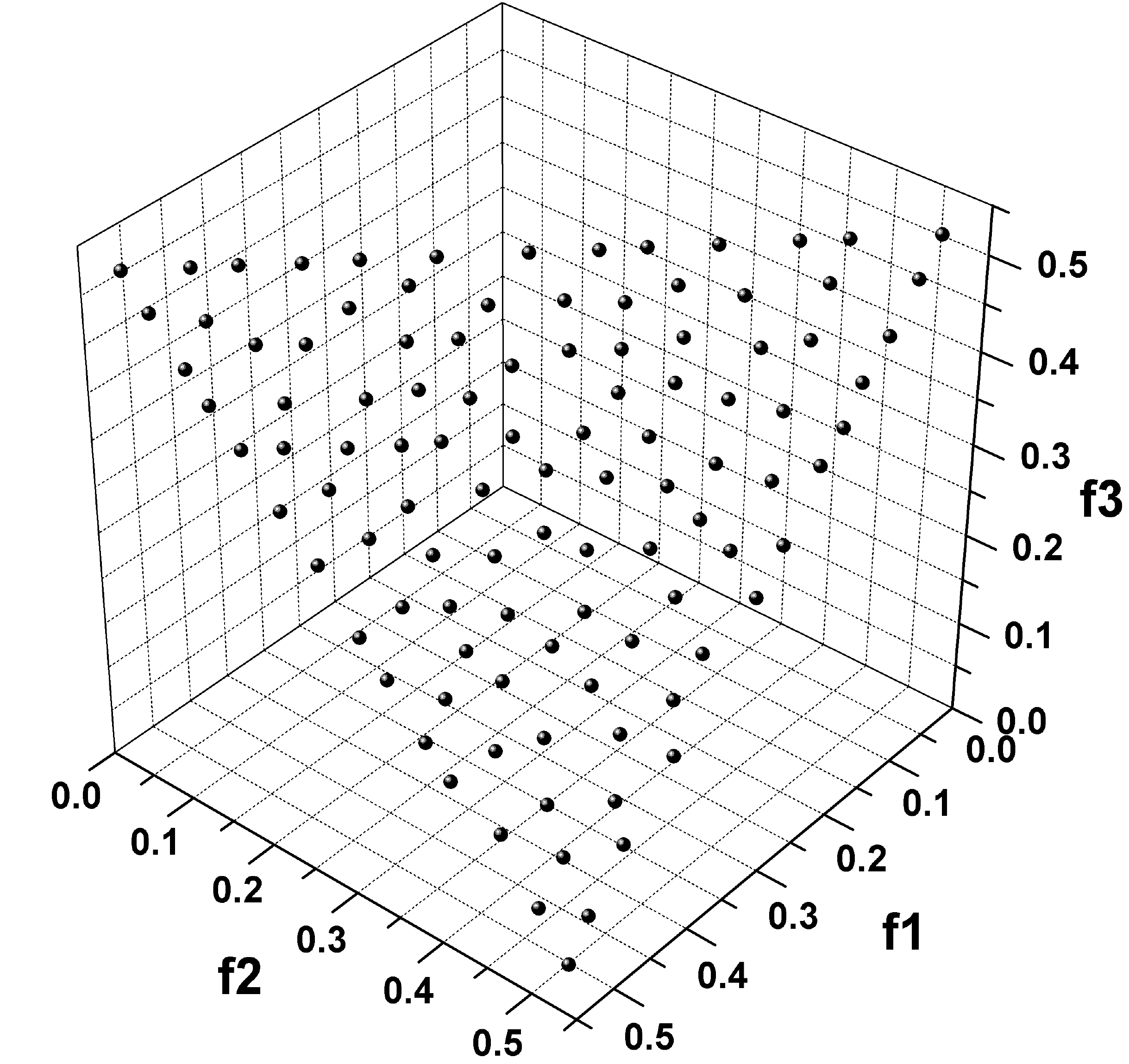}\\
			(a) MOEA/D & (b) A-NSGA-III & (c) RVEA & (d) MOEA/D-AWA & (e) AdaW \\
		\end{tabular}
	\end{center}
	\caption{The final solution set of the five algorithms on the inverted DTLZ1.}
	\label{Fig:IDTLZ1-3}
\end{figure*}

\subsection{On Inverted Simplex-like Pareto Fronts}

\begin{figure*}[tbp]
	\begin{center}
		\footnotesize
		\begin{tabular}{@{}c@{}c@{}c@{}c@{}c@{}}
			\includegraphics[scale=0.14]{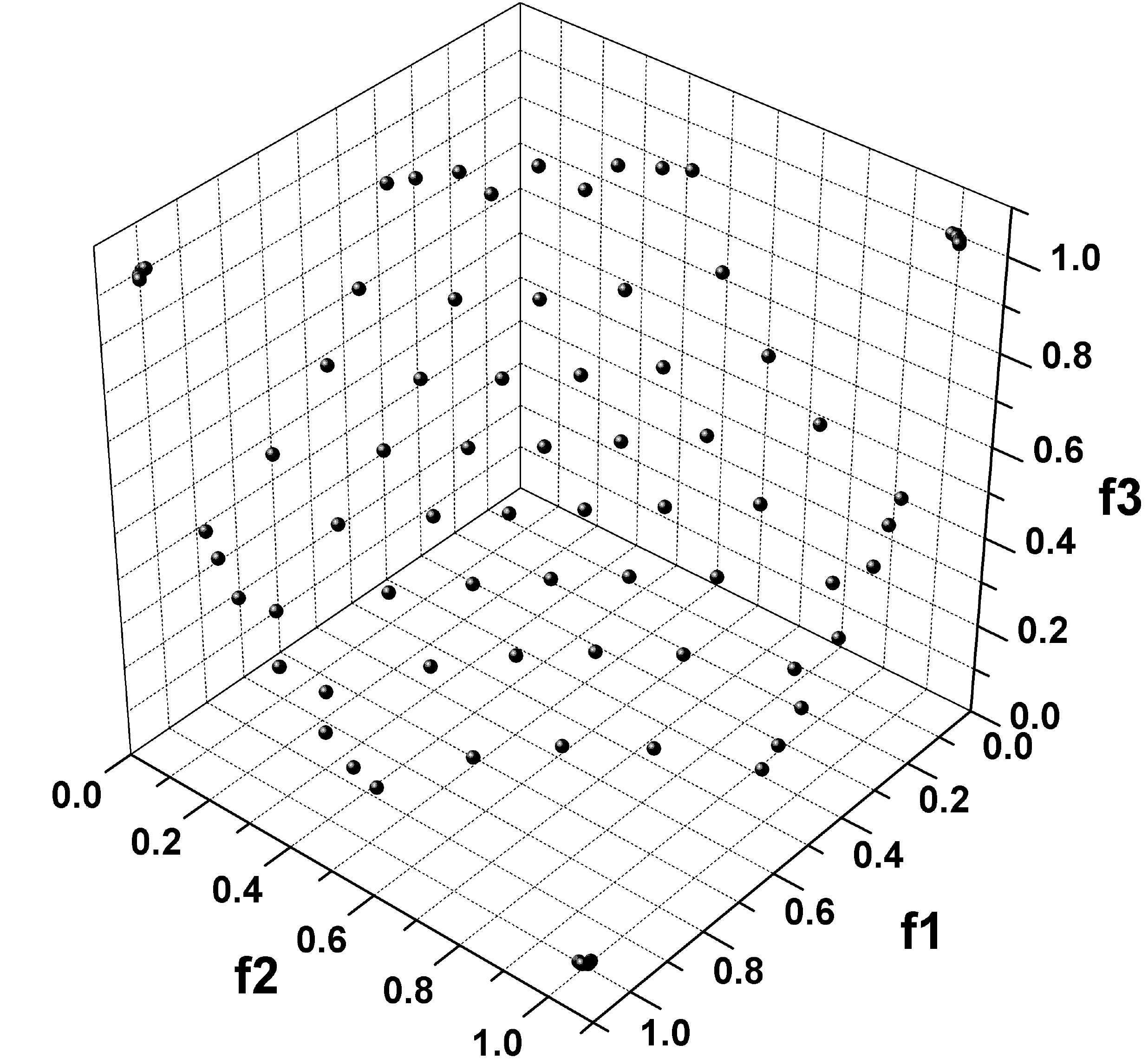}&
			\includegraphics[scale=0.14]{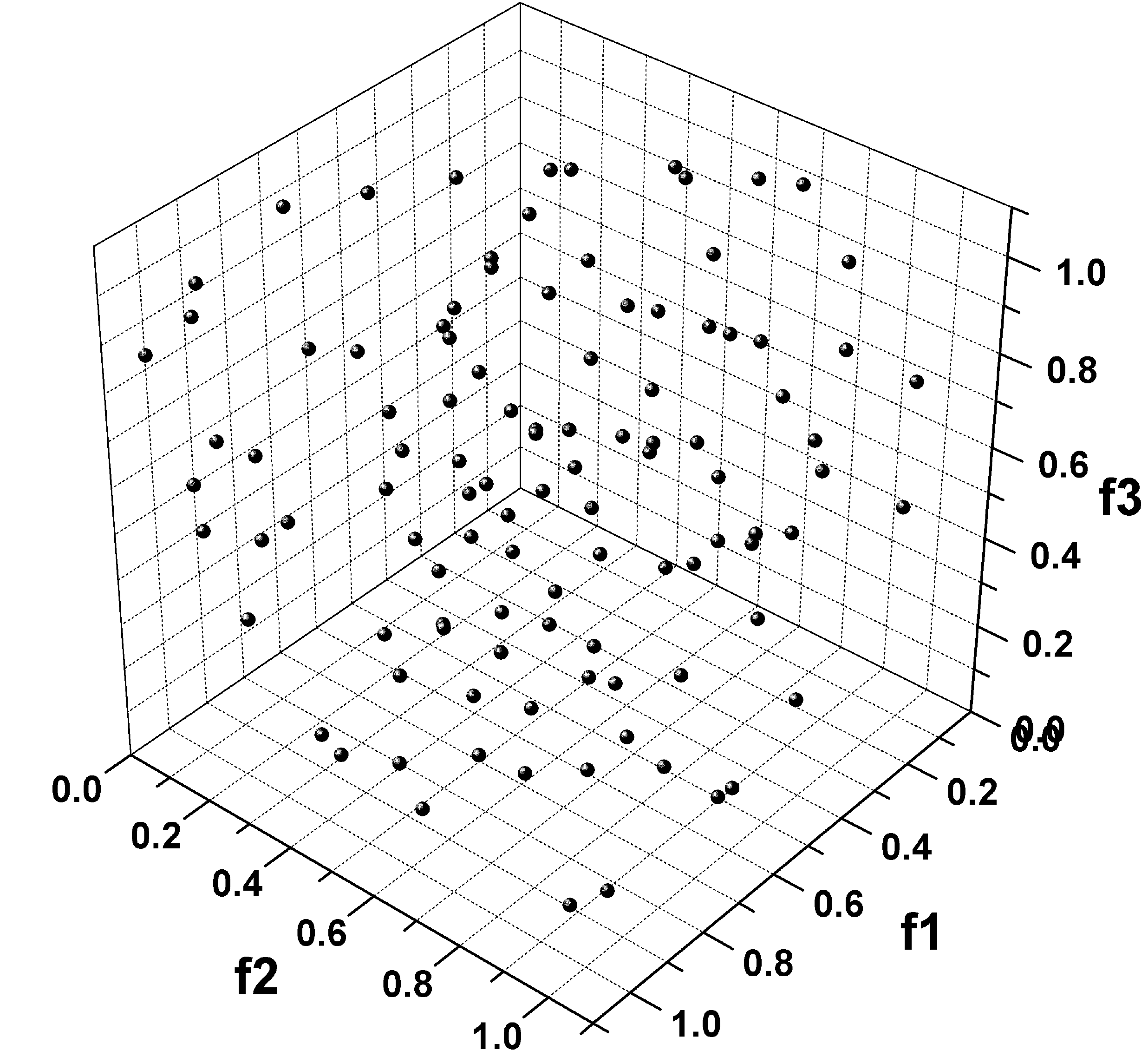}&
			\includegraphics[scale=0.14]{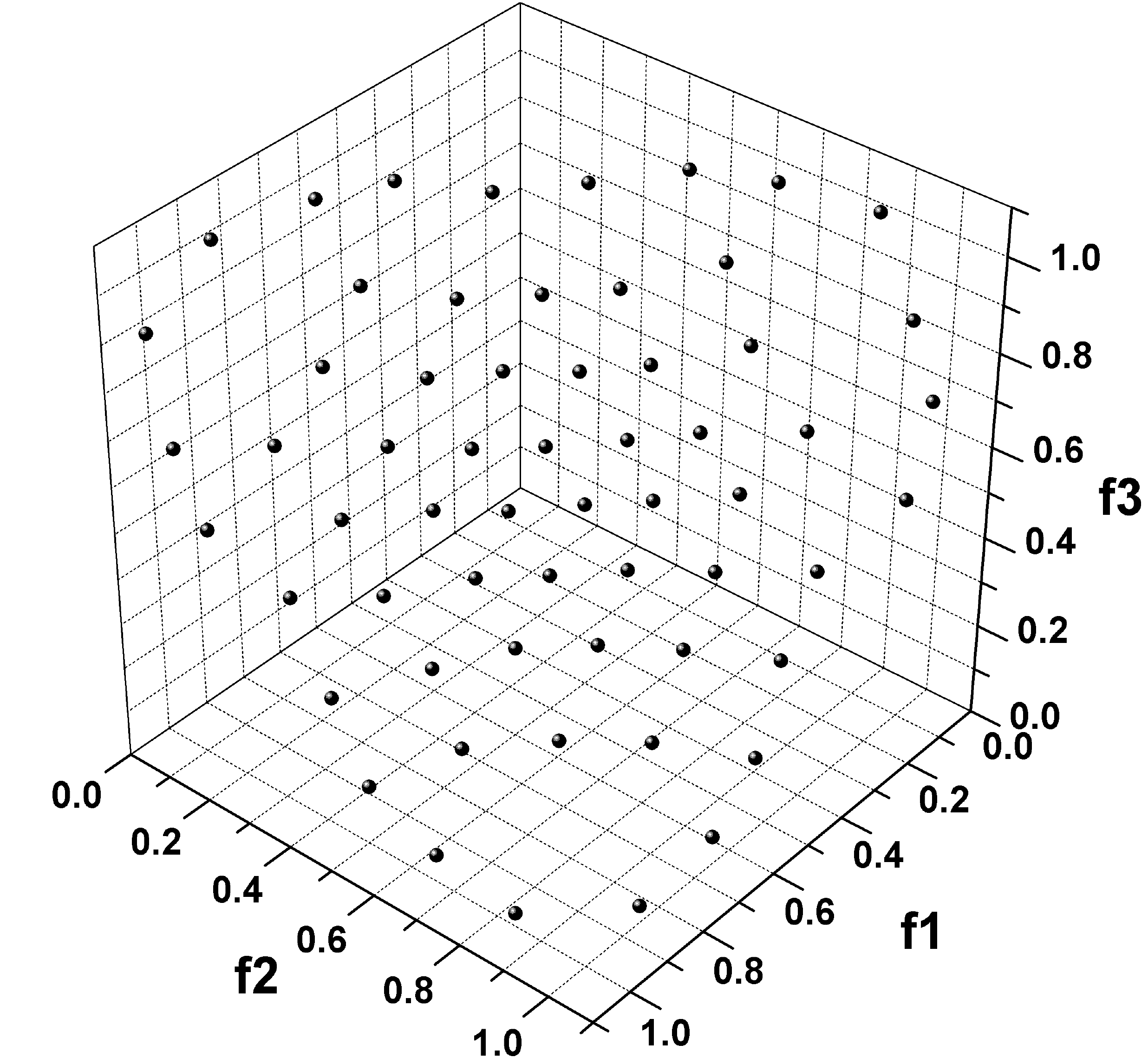}&
			\includegraphics[scale=0.14]{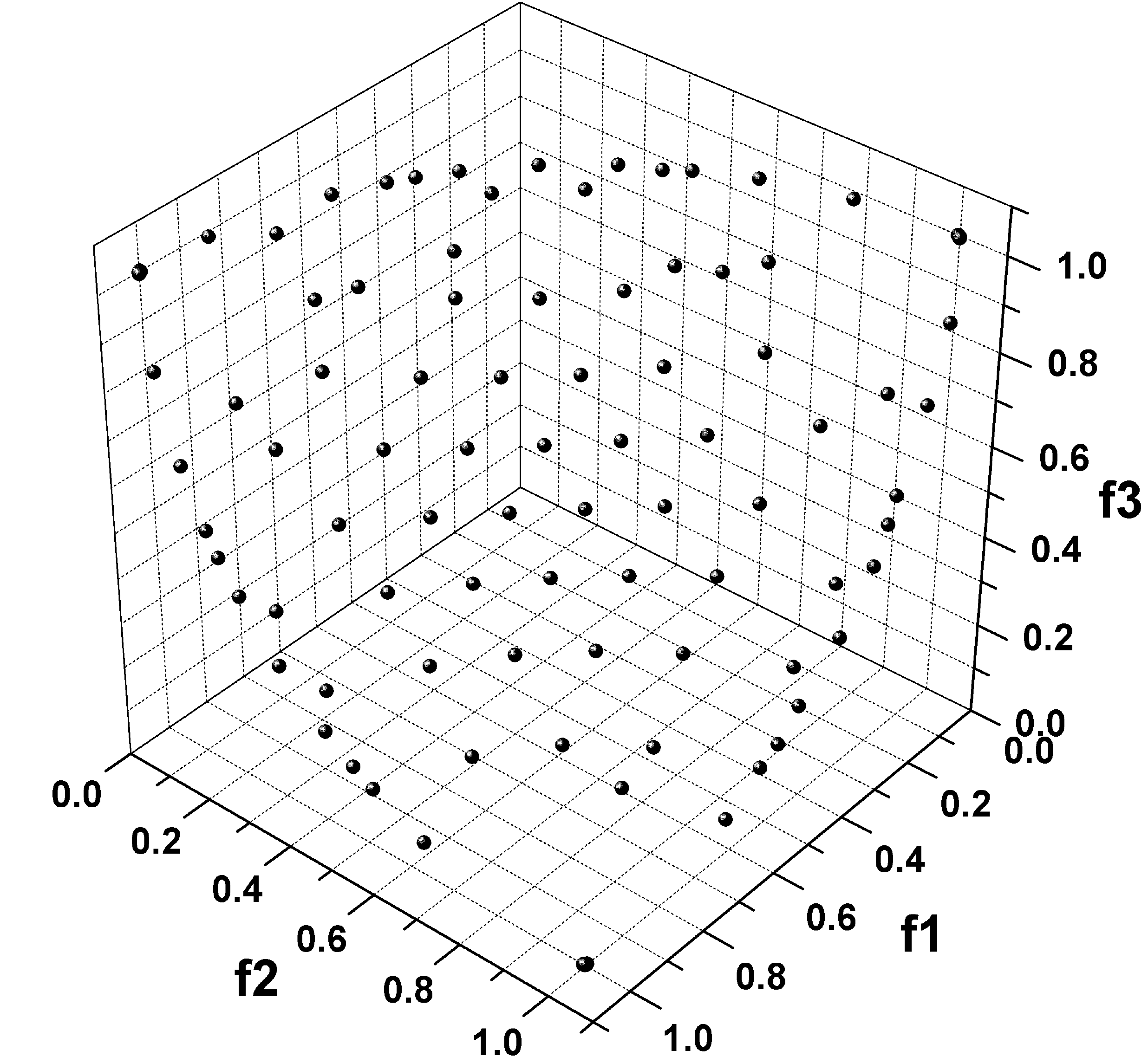}&
			\includegraphics[scale=0.14]{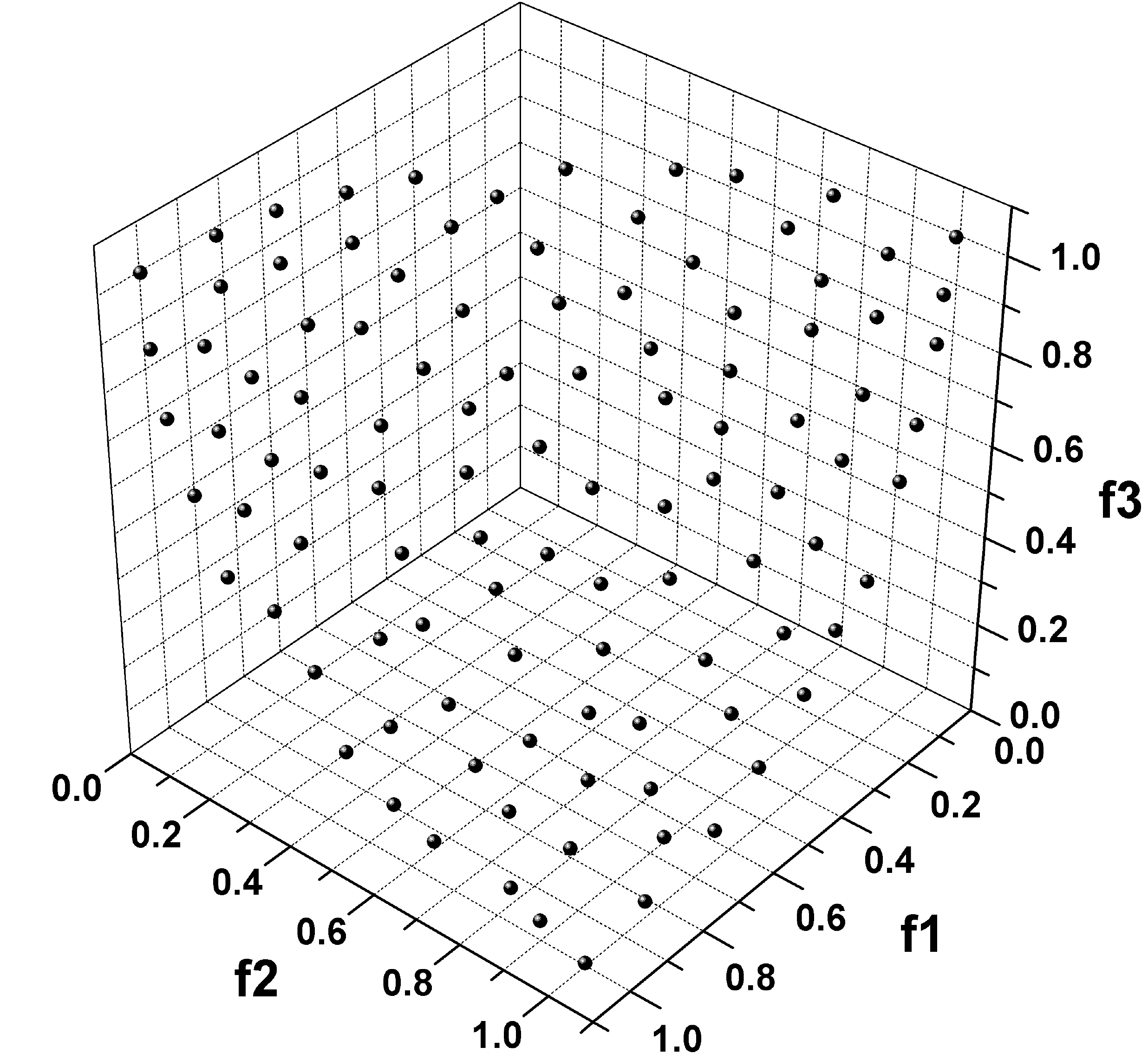}\\
			(a) MOEA/D & (b) A-NSGA-III & (c) RVEA & (d) MOEA/D-AWA & (e) AdaW \\
		\end{tabular}
	\end{center}
	\caption{The final solution set of the five algorithms on the inverted DTLZ2.}
	\label{Fig:IDTLZ2-3}
\end{figure*}

The proposed AdaW has shown a clear advantage over its competitors on this group. 
\mbox{Figures~\ref{Fig:IDTLZ1-3}--\ref{Fig:IDTLZ2-3}} plot the final solution set 
of the five algorithms on IDTLZ1 and IDTLZ2,
respectively.
As shown,
many solutions of MOEA/D and MOEA/D-AWA concentrate on the boundary of the Pareto front.
The solutions of A-NSGA-III have a good coverage but are not distributed very uniformly,
while the solutions of RVEA are distributed uniformly 
but their number is apparently less than the population size. 
For AdaW, 
an inverted simple-like Pareto front has no effect on the algorithm's performance, 
and the obtained solution set has a good coverage and uniformity over the whole front. 

\begin{figure*}[tbp]
	\begin{center}
		\footnotesize
		\begin{tabular}{@{}c@{}c@{}c@{}c@{}c@{}}
			\includegraphics[scale=0.15]{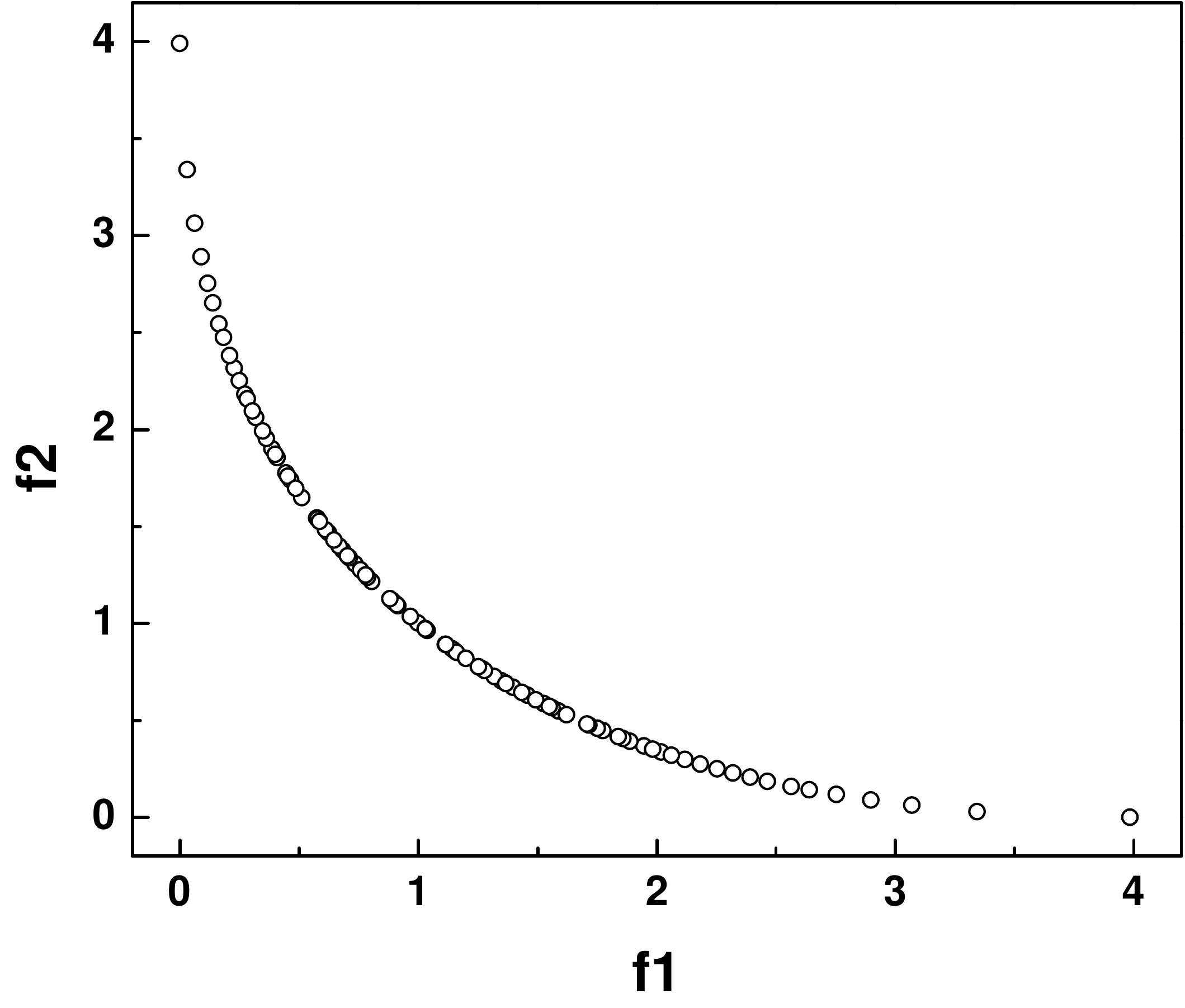}~&
			~\includegraphics[scale=0.15]{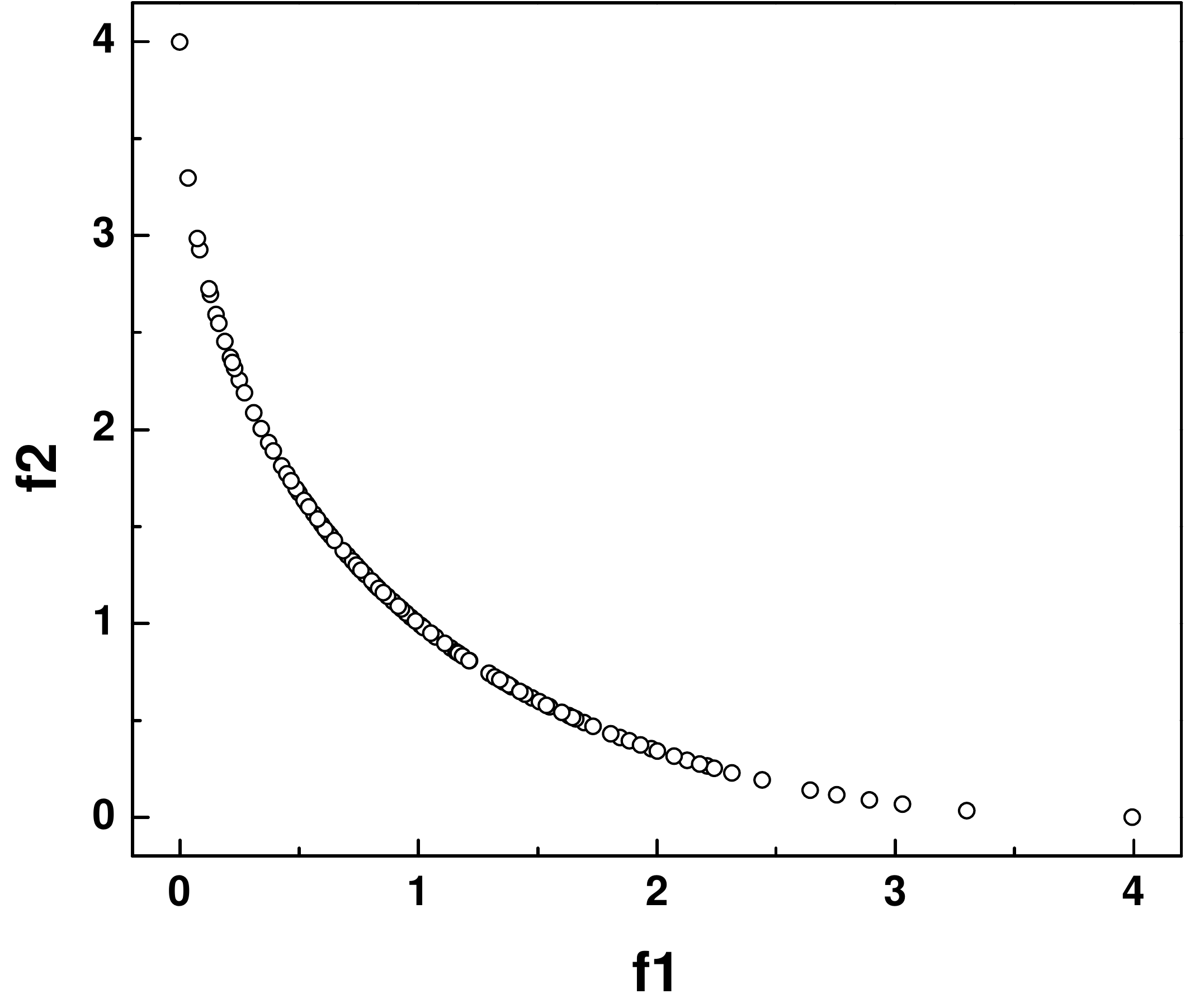}~&
			~\includegraphics[scale=0.15]{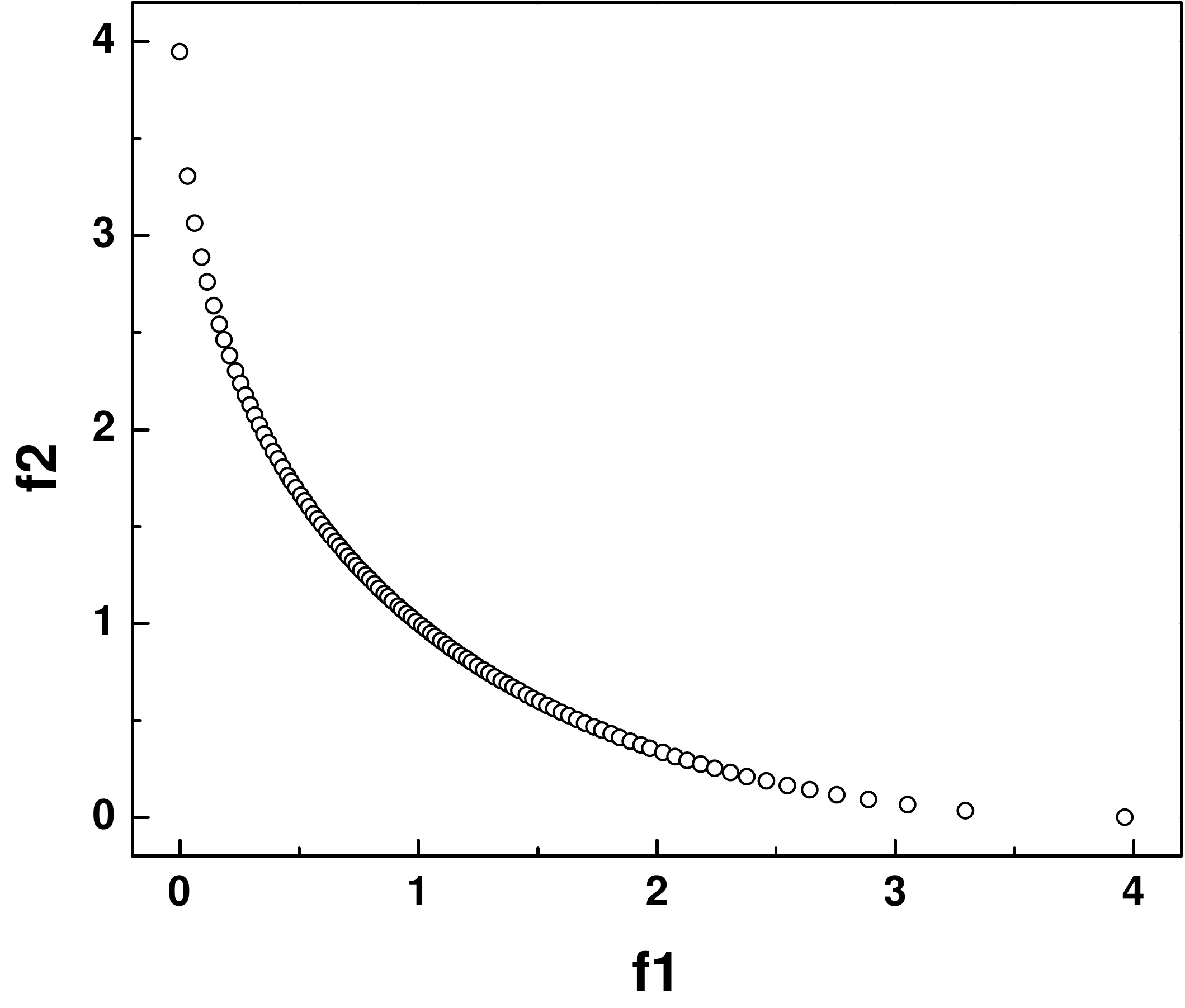}~&
			~\includegraphics[scale=0.15]{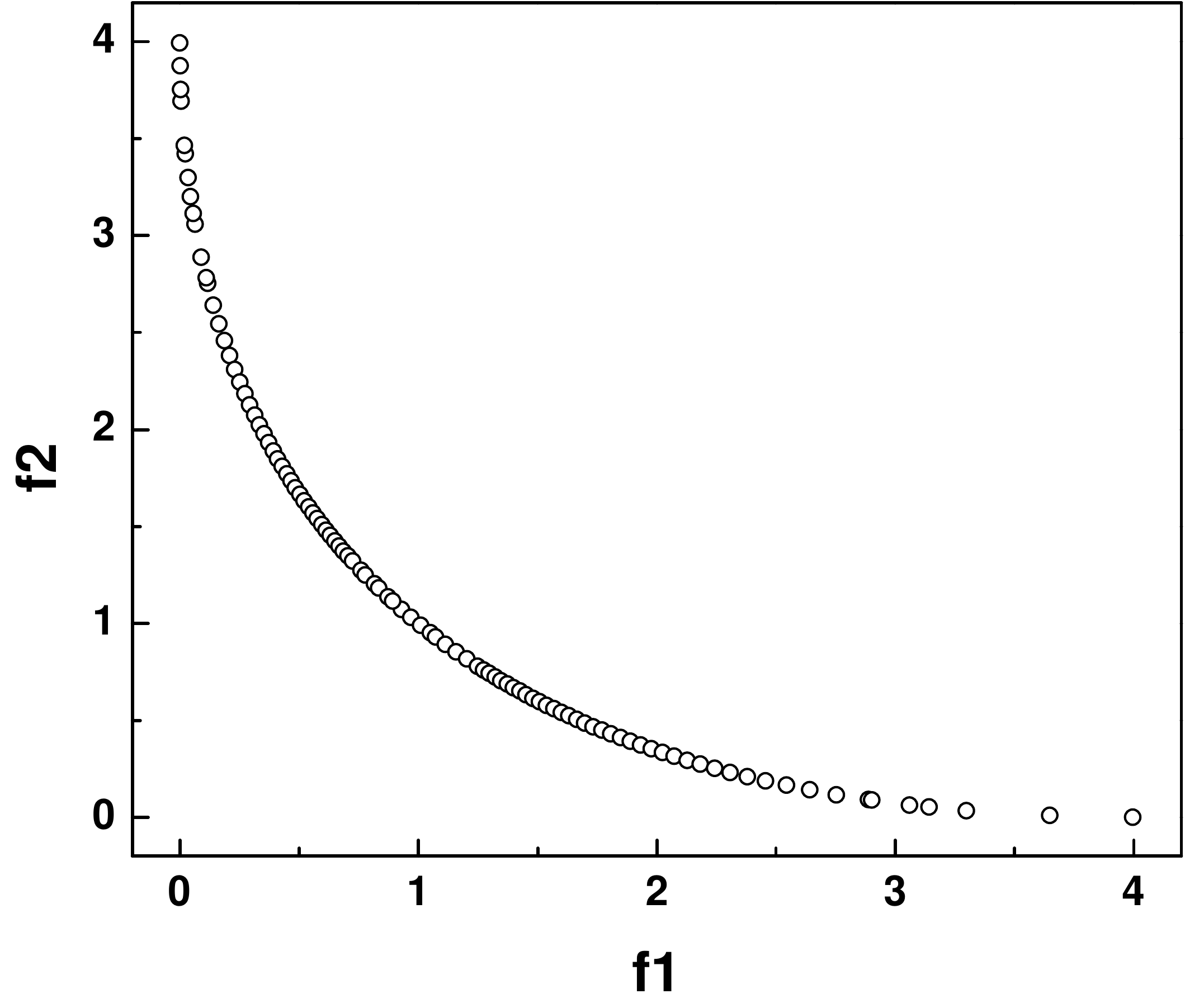}~&
			~\includegraphics[scale=0.15]{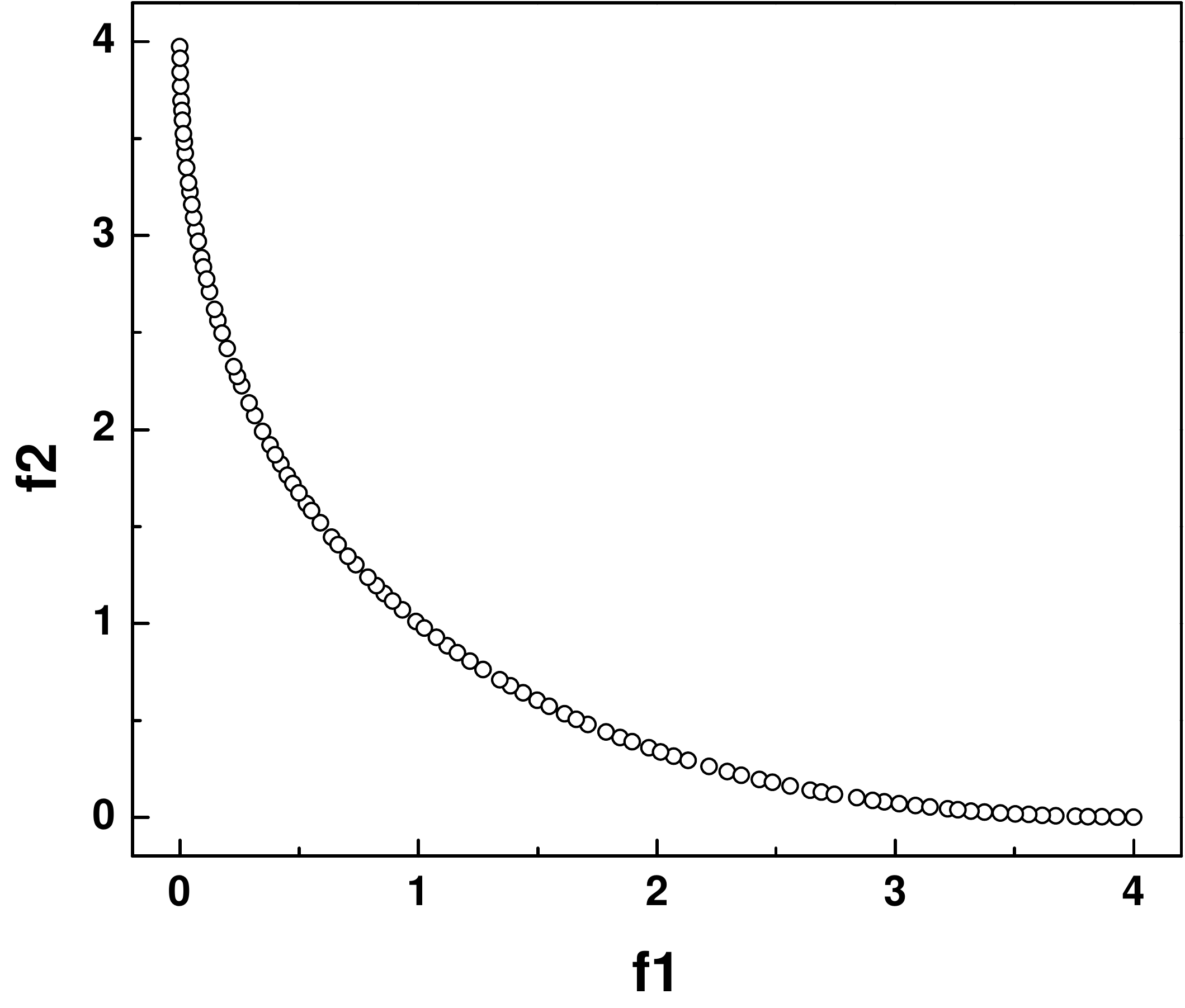}\\
			(a) MOEA/D & (b) A-NSGA-III & (c) RVEA & (d) MOEA/D-AWA & (e) AdaW \\
		\end{tabular}
	\end{center}
	\caption{The final solution set of the five algorithms on SCH1.}
	\label{Fig:SCH}
\end{figure*}
\begin{figure*}[!]
	\begin{center}
		\footnotesize
		\begin{tabular}{@{}c@{}c@{}c@{}c@{}c@{}}
			\includegraphics[scale=0.15]{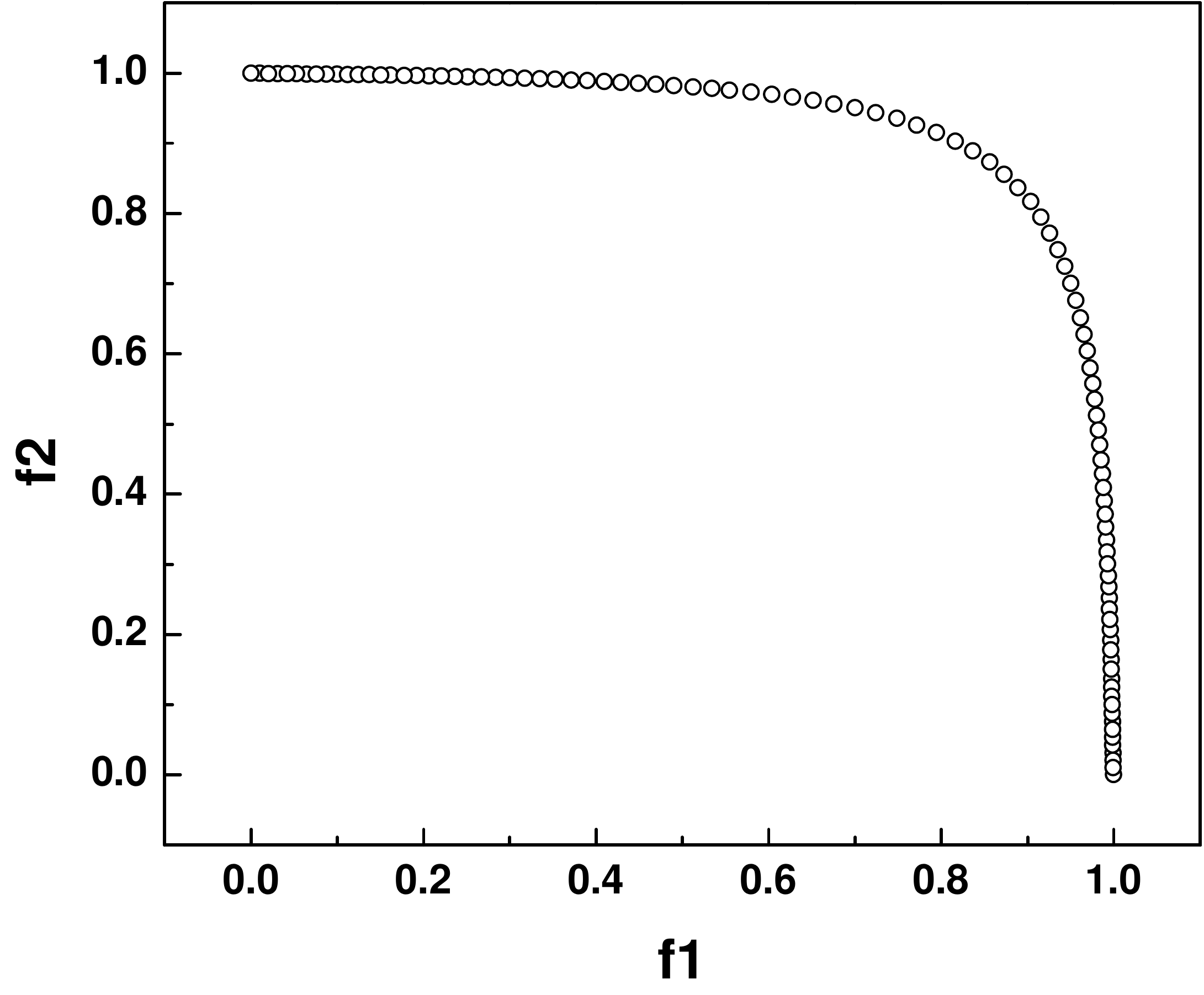}~&
			~\includegraphics[scale=0.15]{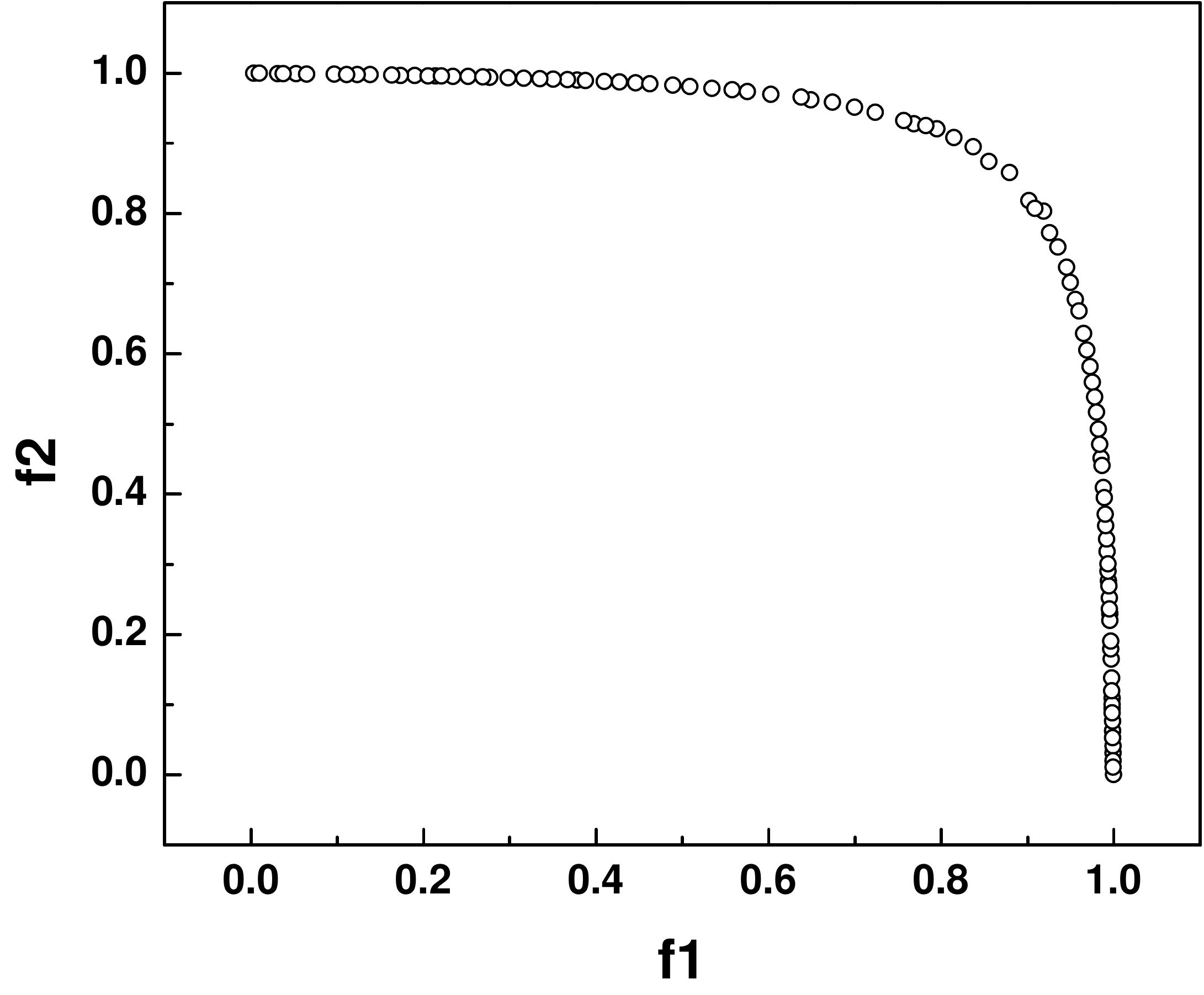}~&
			~\includegraphics[scale=0.15]{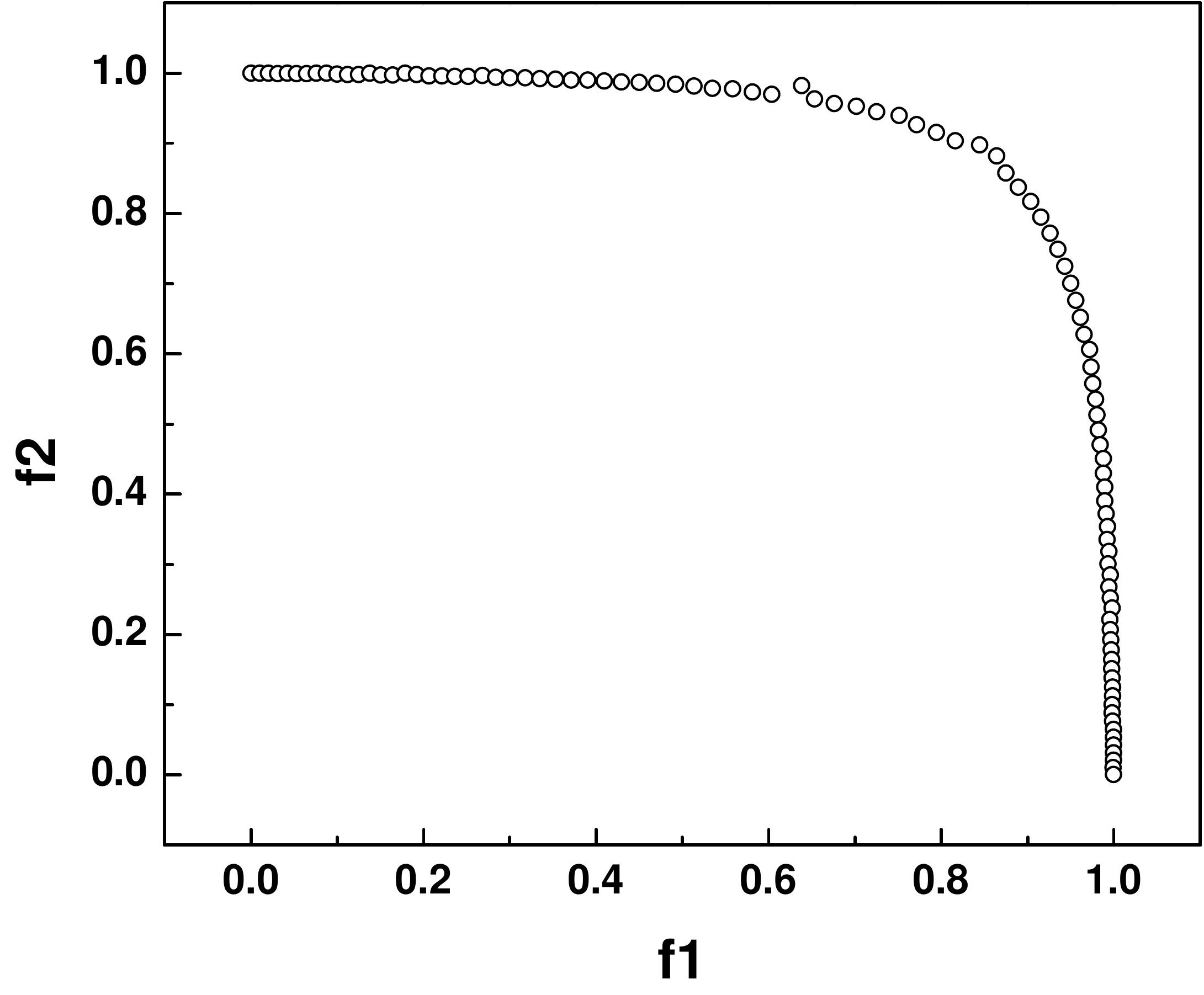}~&
			~\includegraphics[scale=0.15]{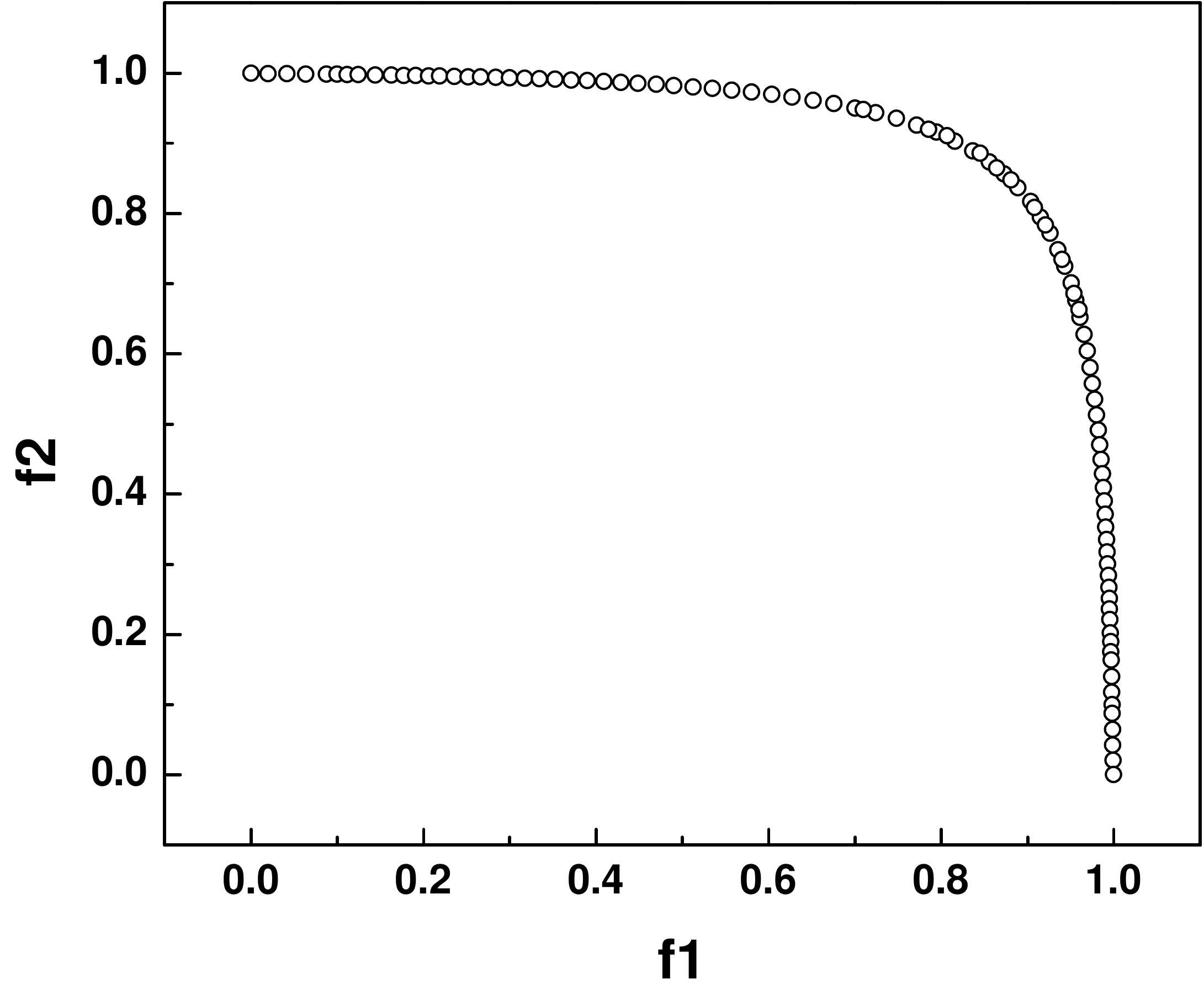}~&
			~\includegraphics[scale=0.15]{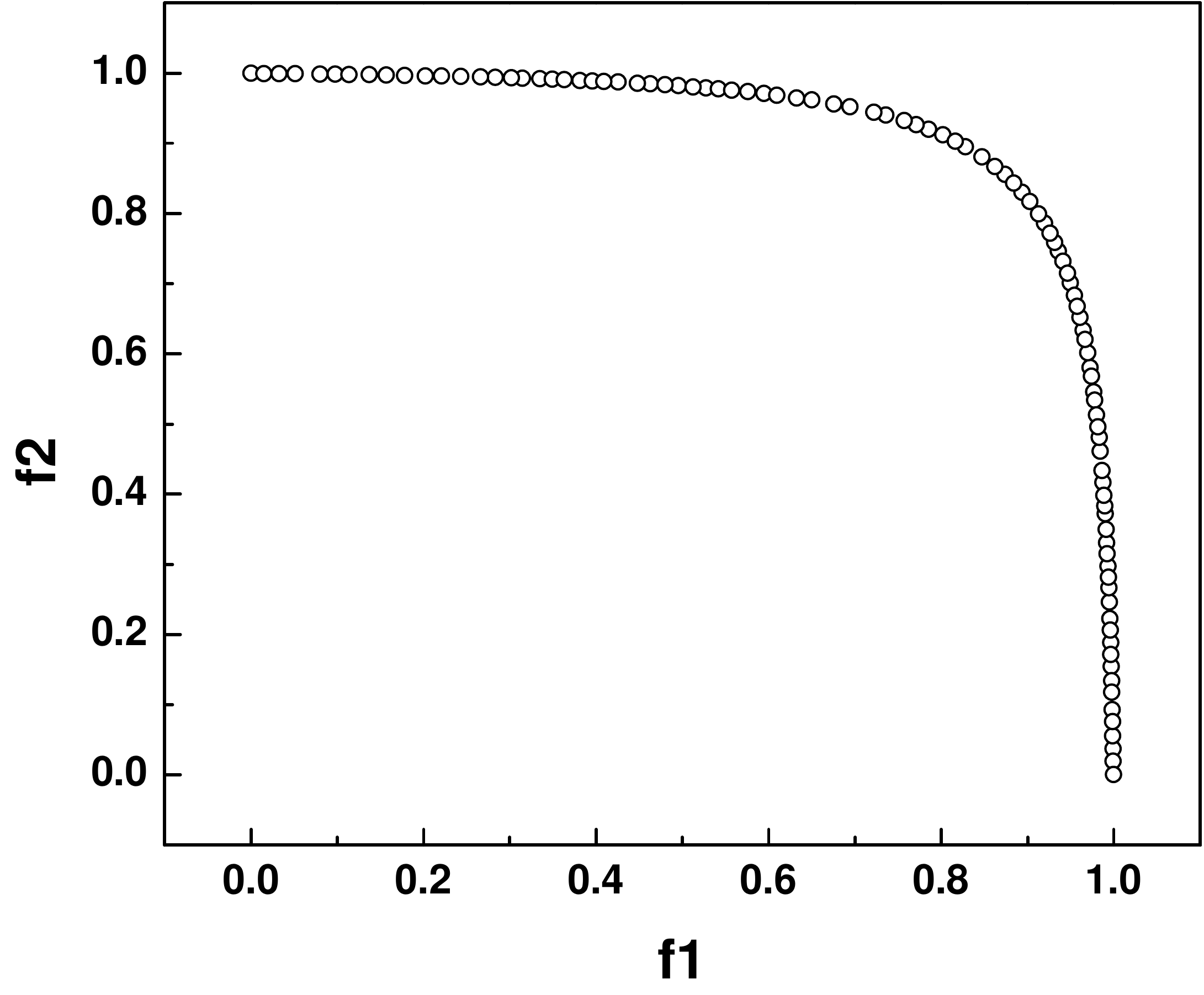}\\
			(a) MOEA/D & (b) A-NSGA-III & (c) RVEA & (d) MOEA/D-AWA & (e) AdaW \\
		\end{tabular}
	\end{center}
	\caption{The final solution set of the five algorithms on FON.}
	\label{Fig:FON}
\end{figure*} 

\subsection{On Highly Nonlinear Pareto Fronts}

The peer algorithms perform differently on the two instances of this group.
On the problem with a concave Pareto front (i.e., FON), 
all the algorithms work well (\mbox{Figure~\ref{Fig:FON}}), 
despite A-NSGA-II and RVEA performing slightly worse than the other three.
In contrast, 
on the problem with a convex Pareto front (i.e., SCH1),
only the proposed AdaW can obtain a well-distributed solution set,
and the others fail to extend their solutions to the boundary of the Pareto front (\mbox{Figure~\ref{Fig:SCH}}).
This indicates that the convex Pareto front still poses a challenge to decomposition-based approach
even if some weight vector adaptations are introduced.

\begin{figure*}[tbp]
	\begin{center}
		\footnotesize
		\begin{tabular}{@{}c@{}c@{}c@{}c@{}c@{}}
			\includegraphics[scale=0.15]{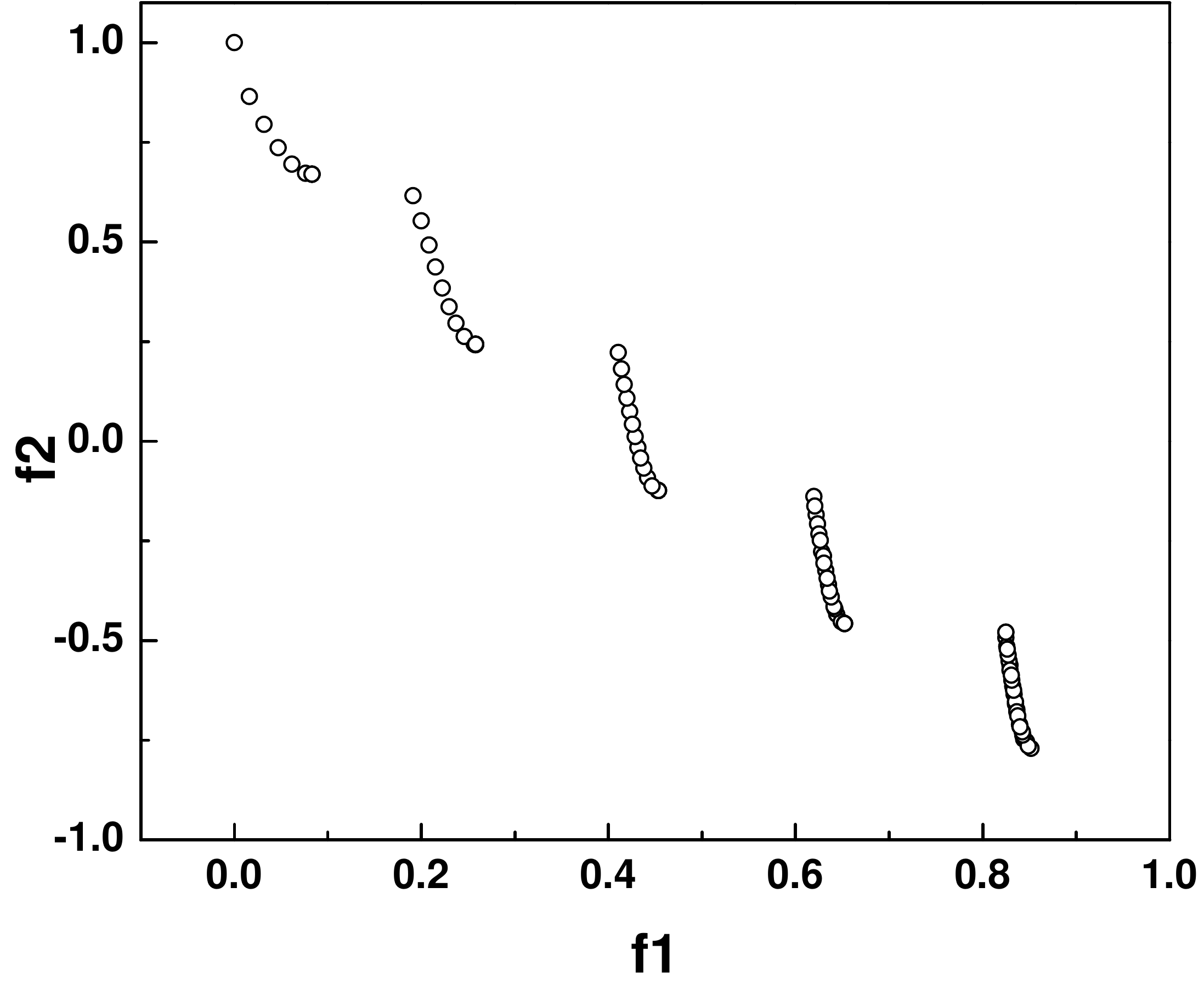}~&
			~\includegraphics[scale=0.15]{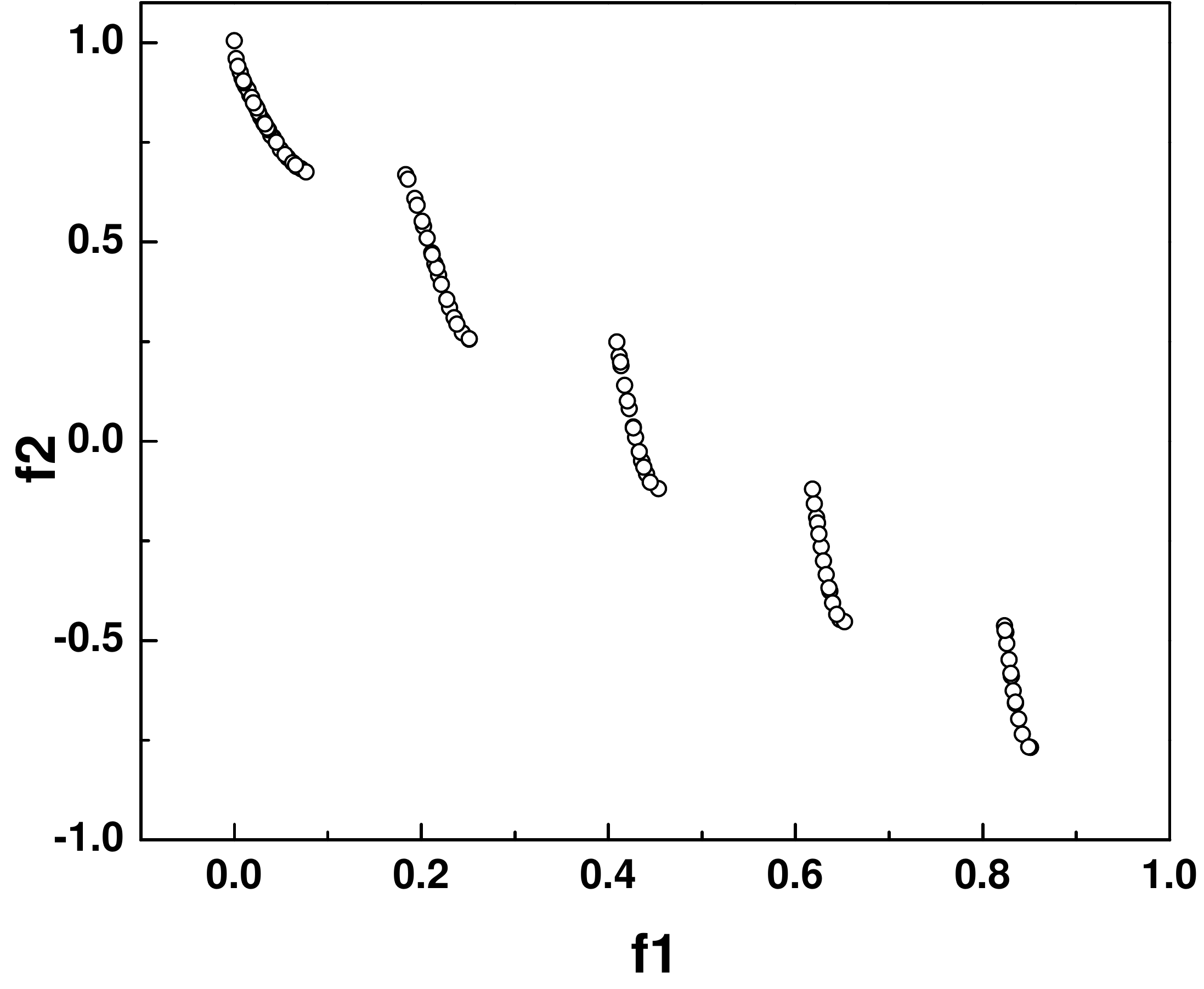}~&
			~\includegraphics[scale=0.15]{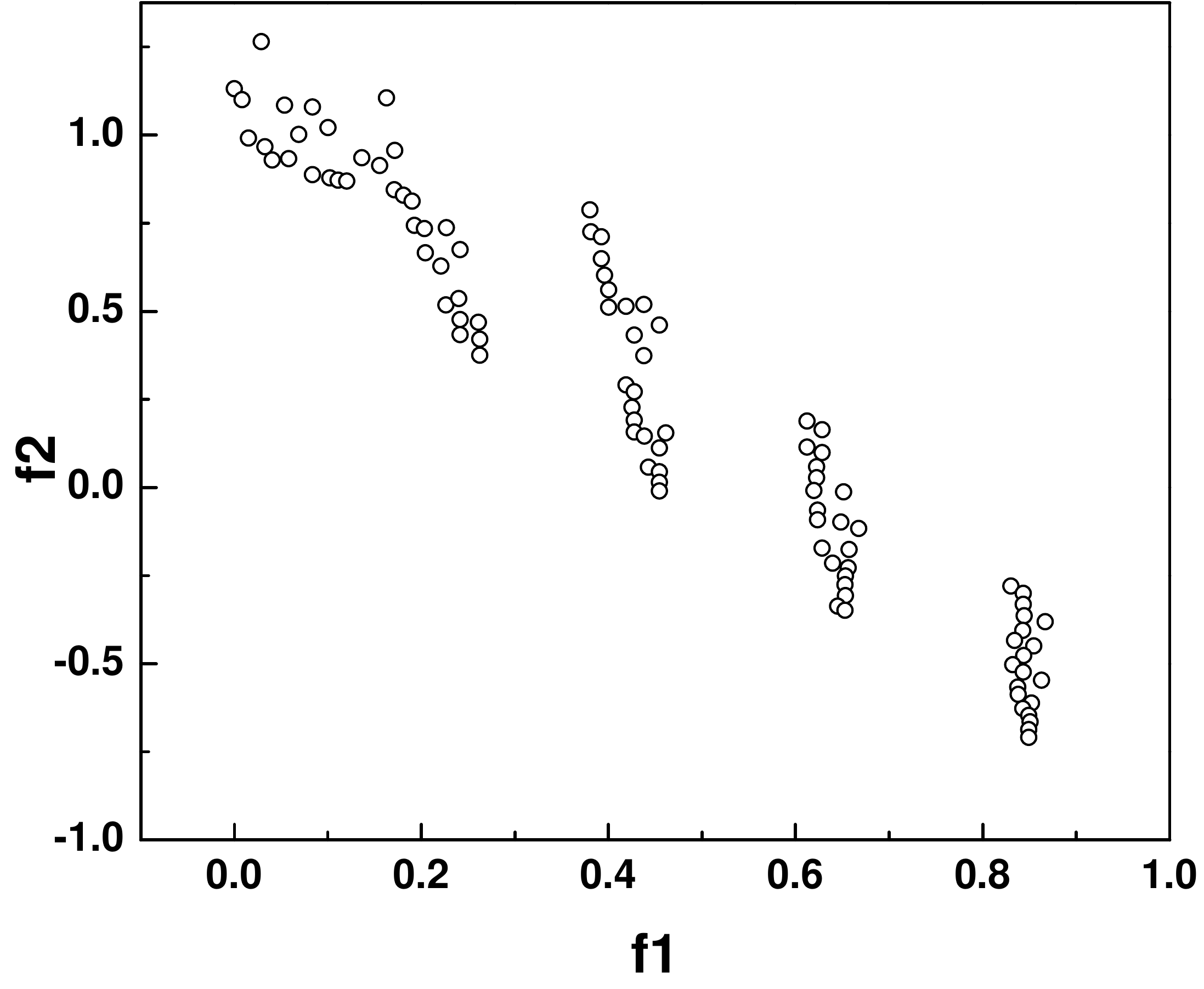}~&
			~\includegraphics[scale=0.15]{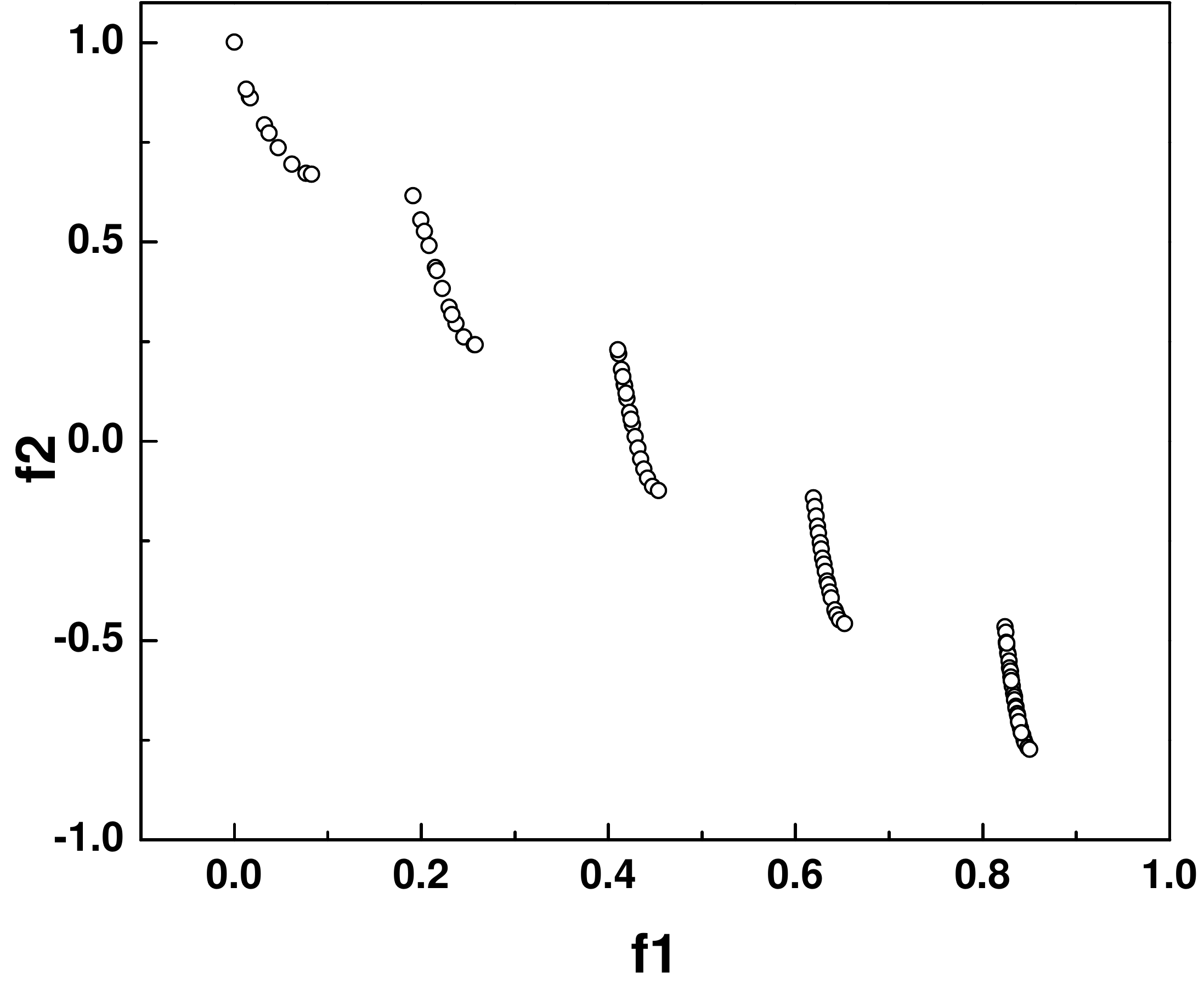}~&
			~\includegraphics[scale=0.15]{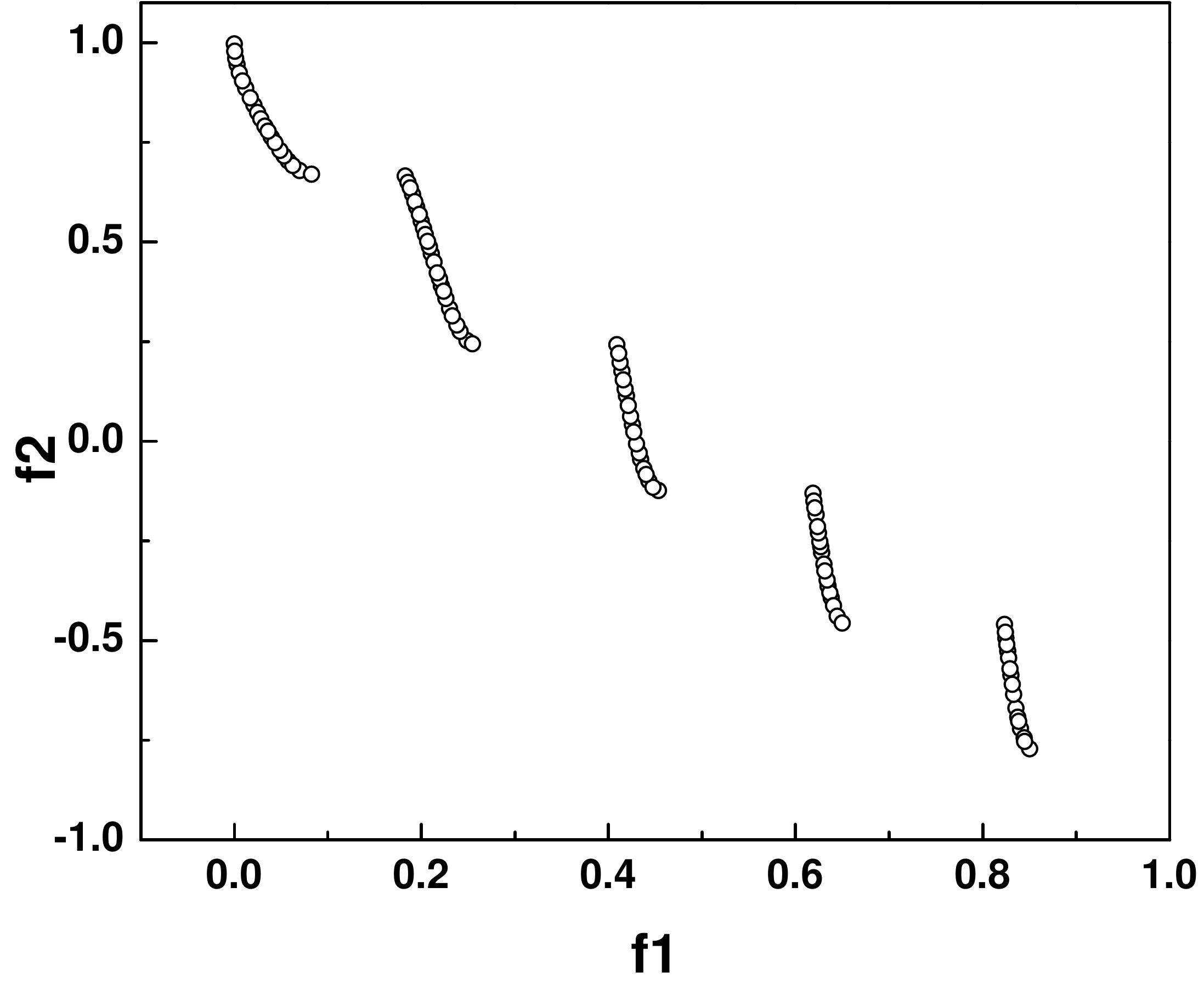}\\
			(a) MOEA/D & (b) A-NSGA-III & (c) RVEA & (d) MOEA/D-AWA & (e) AdaW \\
		\end{tabular}
	\end{center}
	\caption{The final solution set of the five algorithms on ZDT3.}
	\label{Fig:ZDT3}
\end{figure*}

\begin{figure*}[tbp]
	\begin{center}
		\footnotesize
		\begin{tabular}{@{}c@{}c@{}c@{}c@{}c@{}}
			\includegraphics[scale=0.14]{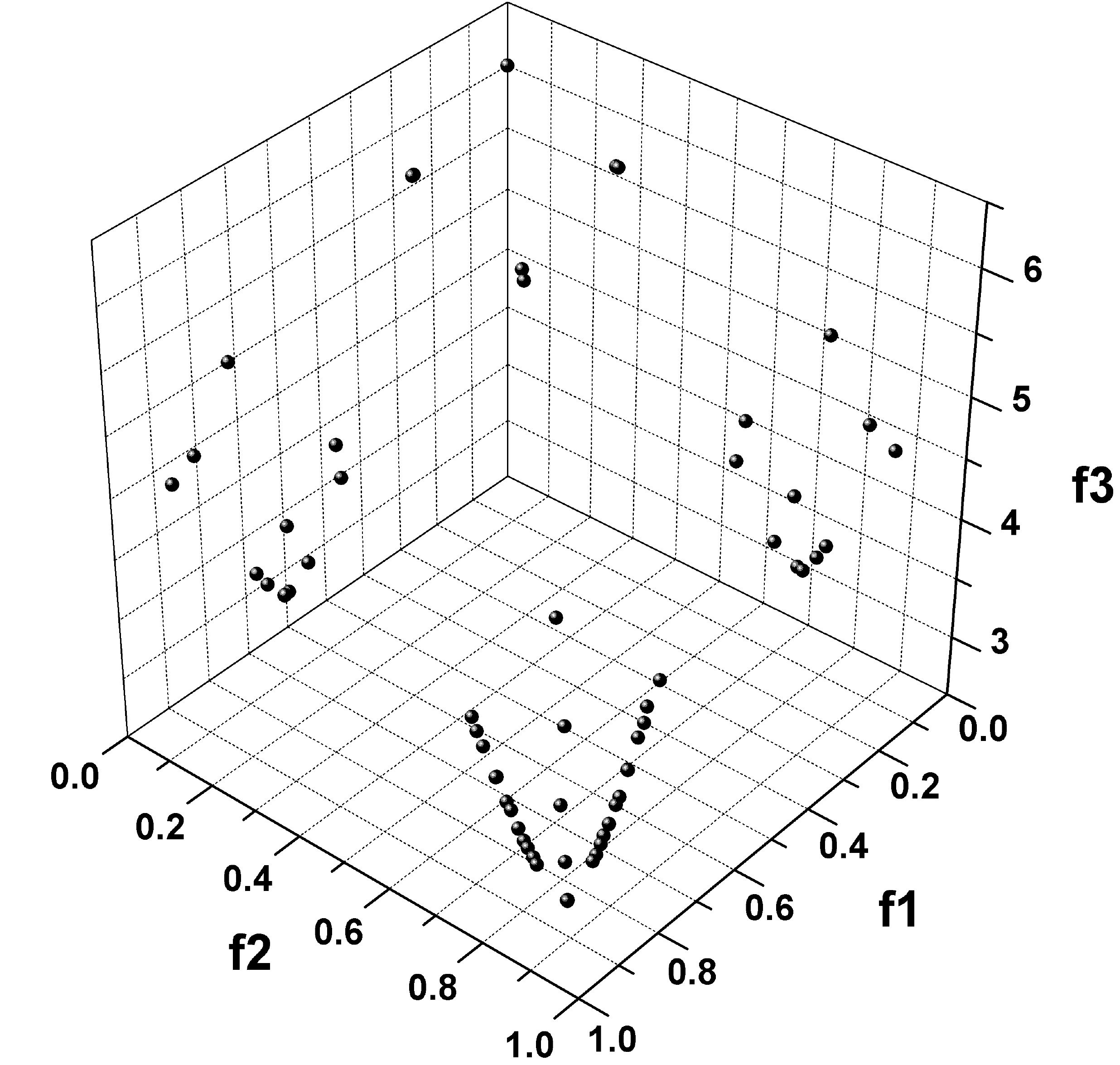}&
			\includegraphics[scale=0.14]{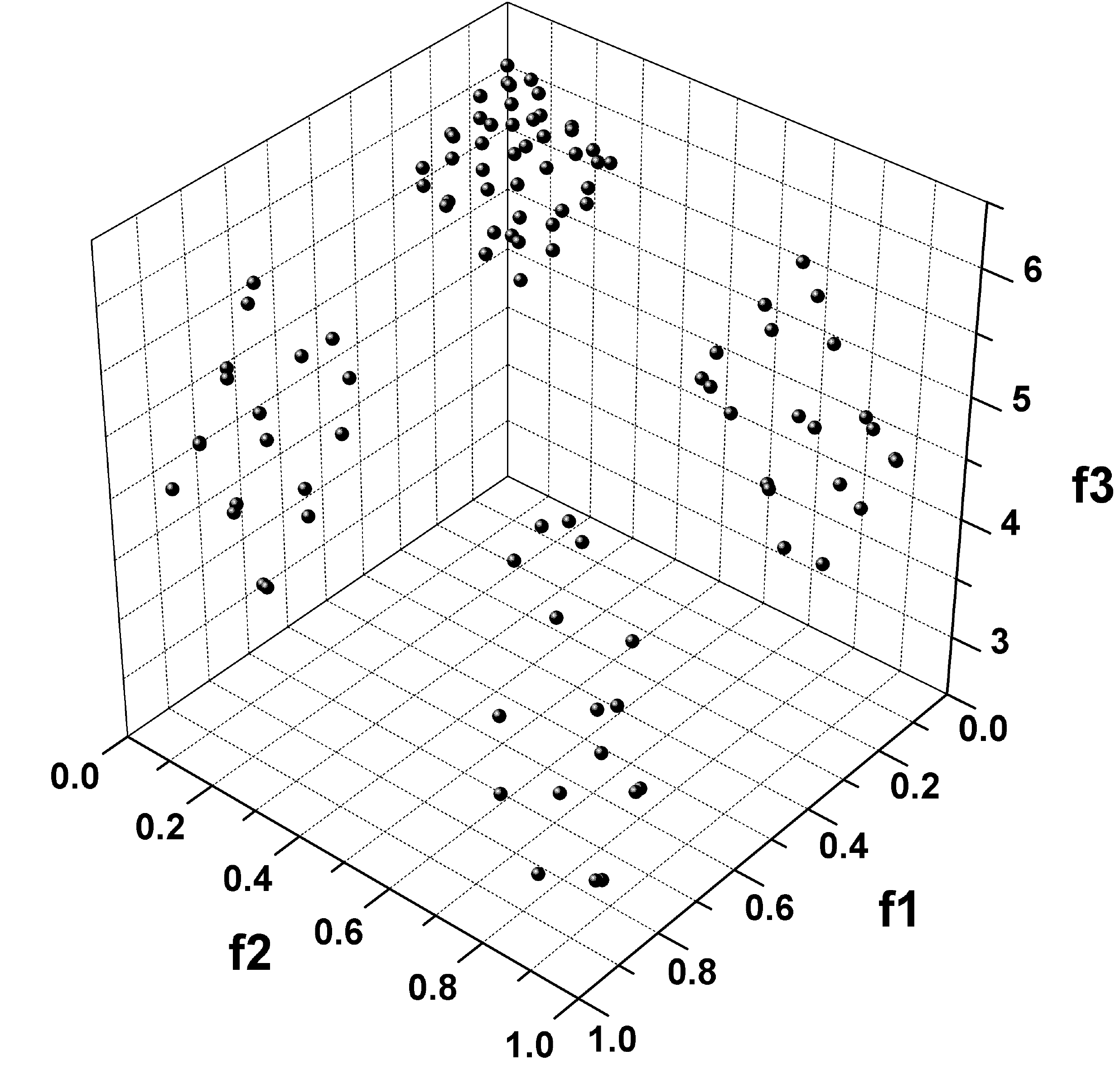}&
			\includegraphics[scale=0.14]{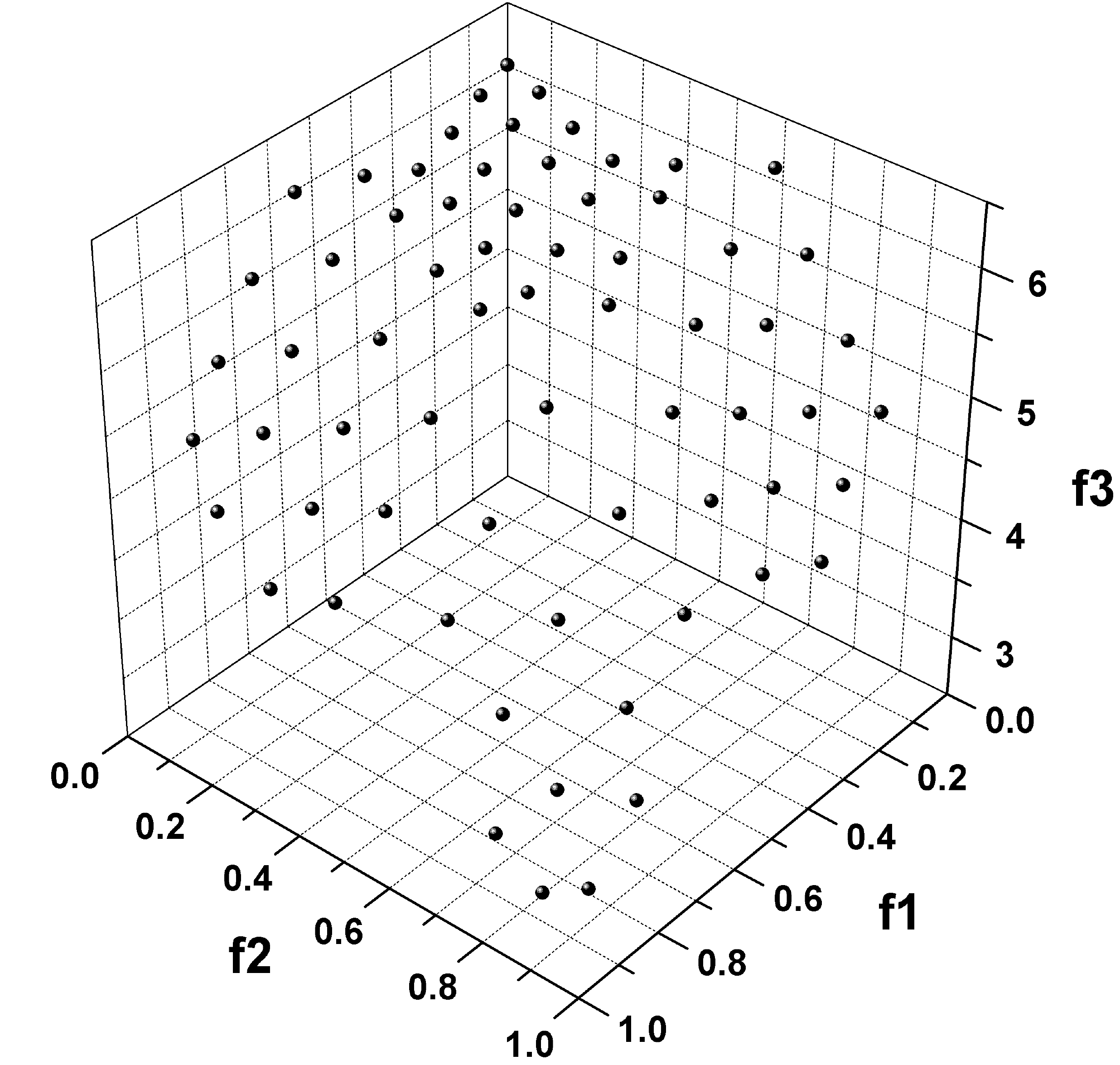}&
			\includegraphics[scale=0.14]{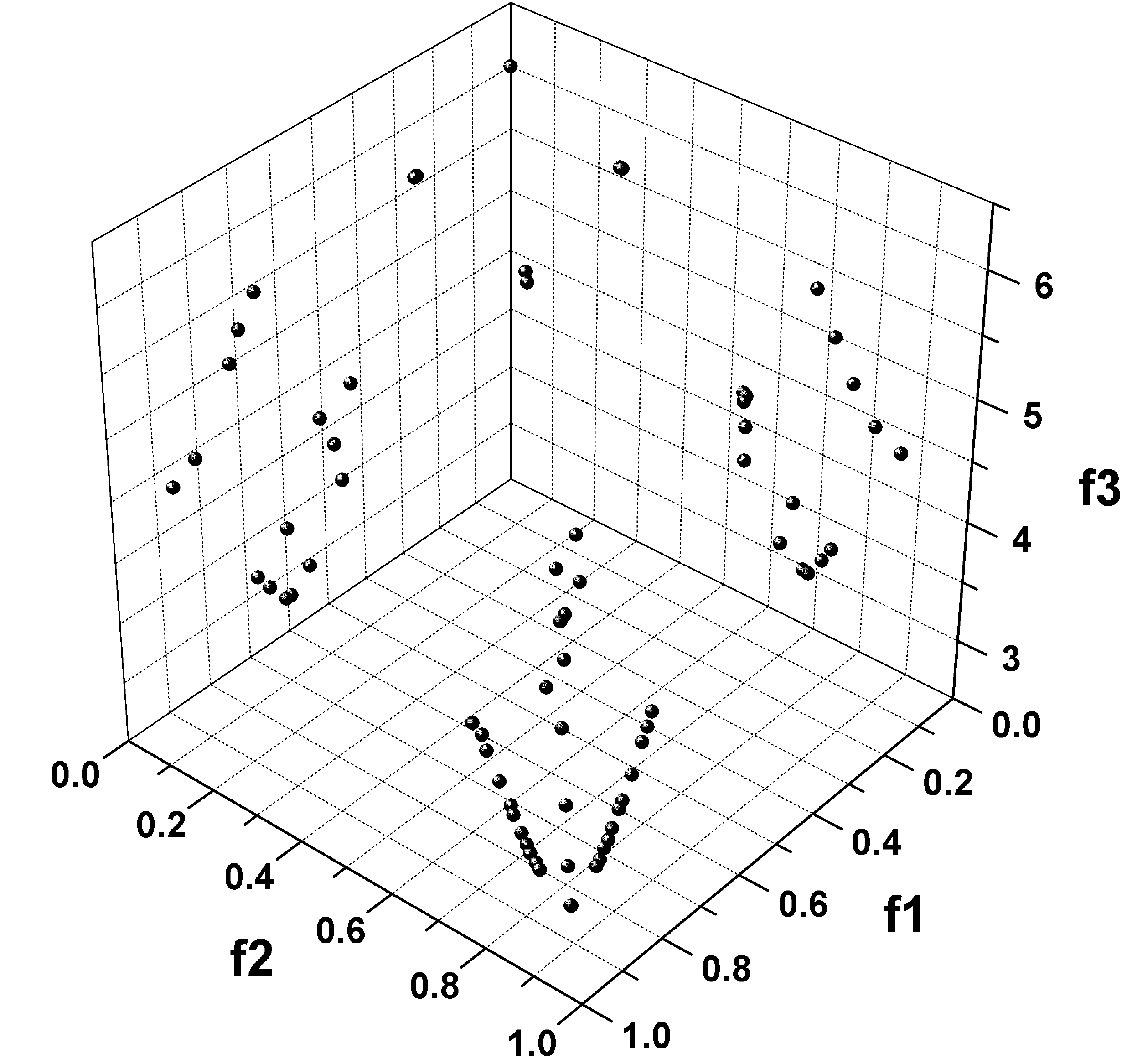}&
			\includegraphics[scale=0.14]{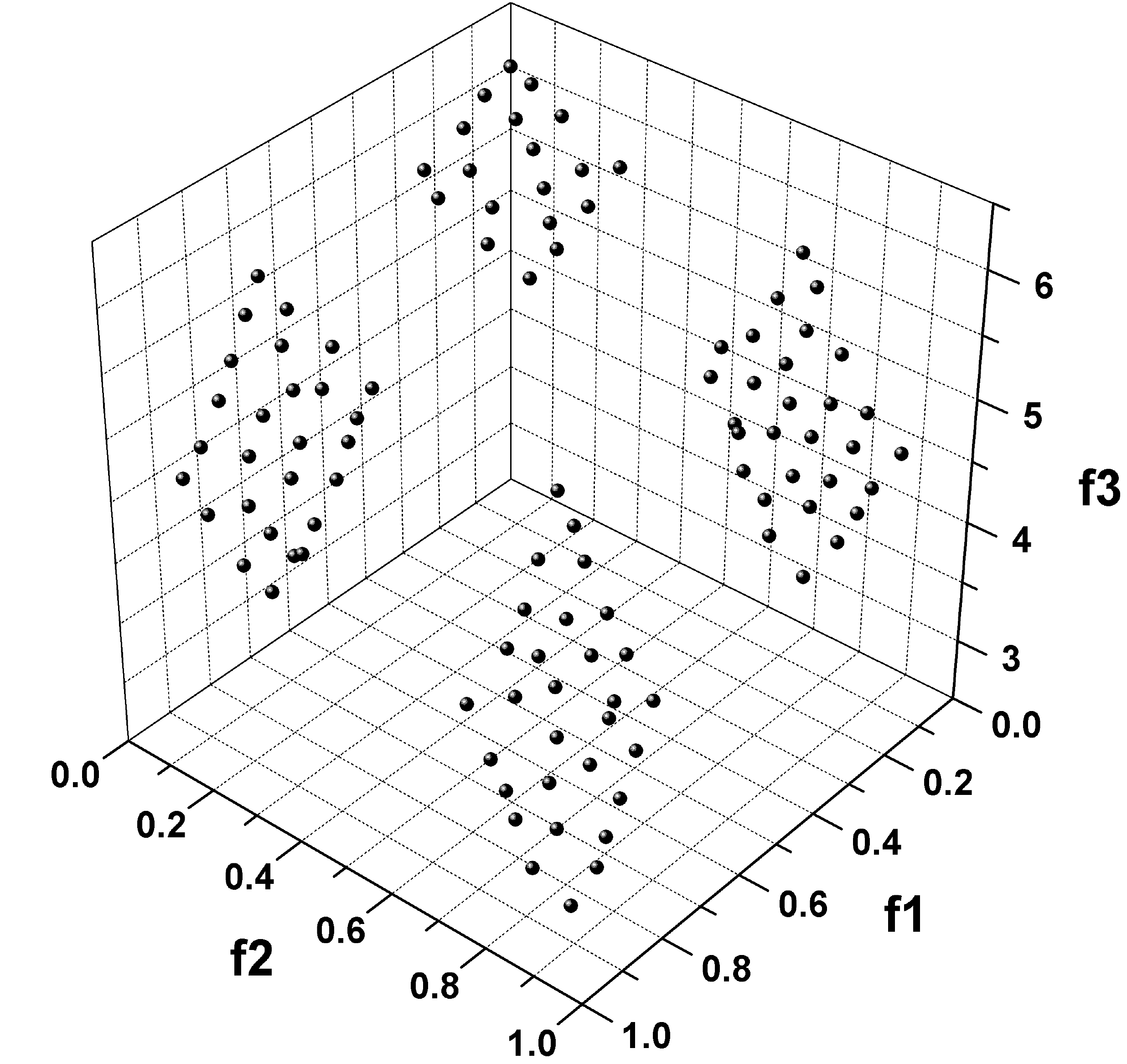}\\
			(a) MOEA/D & (b) A-NSGA-III & (c) RVEA & (d) MOEA/D-AWA & (e) AdaW \\
		\end{tabular}
	\end{center}
	\caption{The final solution set of the five algorithms on DTLZ7.}
	\label{Fig:DTLZ7}
\end{figure*}

\begin{figure*}[tbp]
	\begin{center}
		\footnotesize
		\begin{tabular}{@{}c@{}c@{}c@{}c@{}c@{}}
			\includegraphics[scale=0.15]{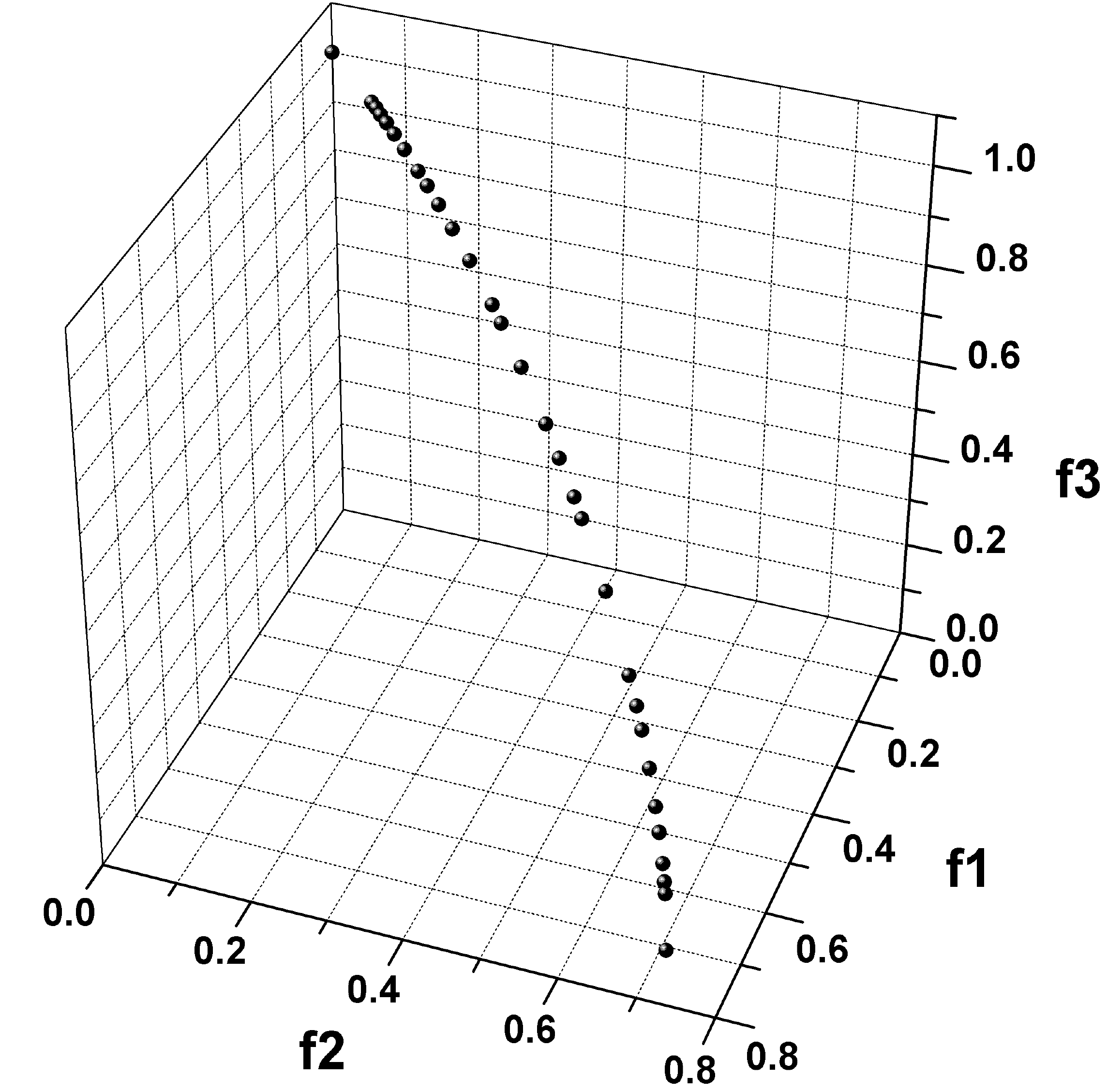}&
			~\includegraphics[scale=0.15]{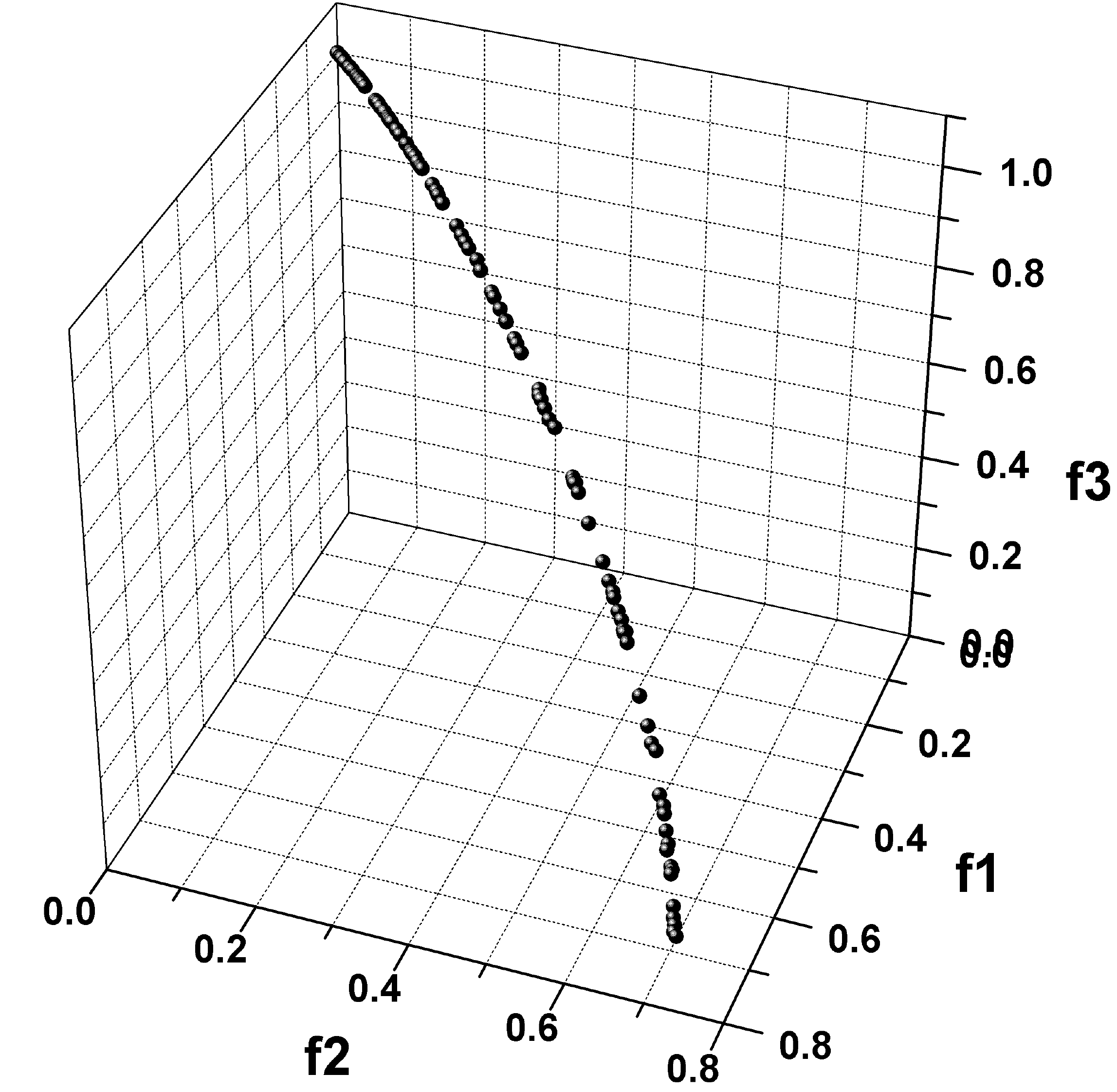}&
			~\includegraphics[scale=0.15]{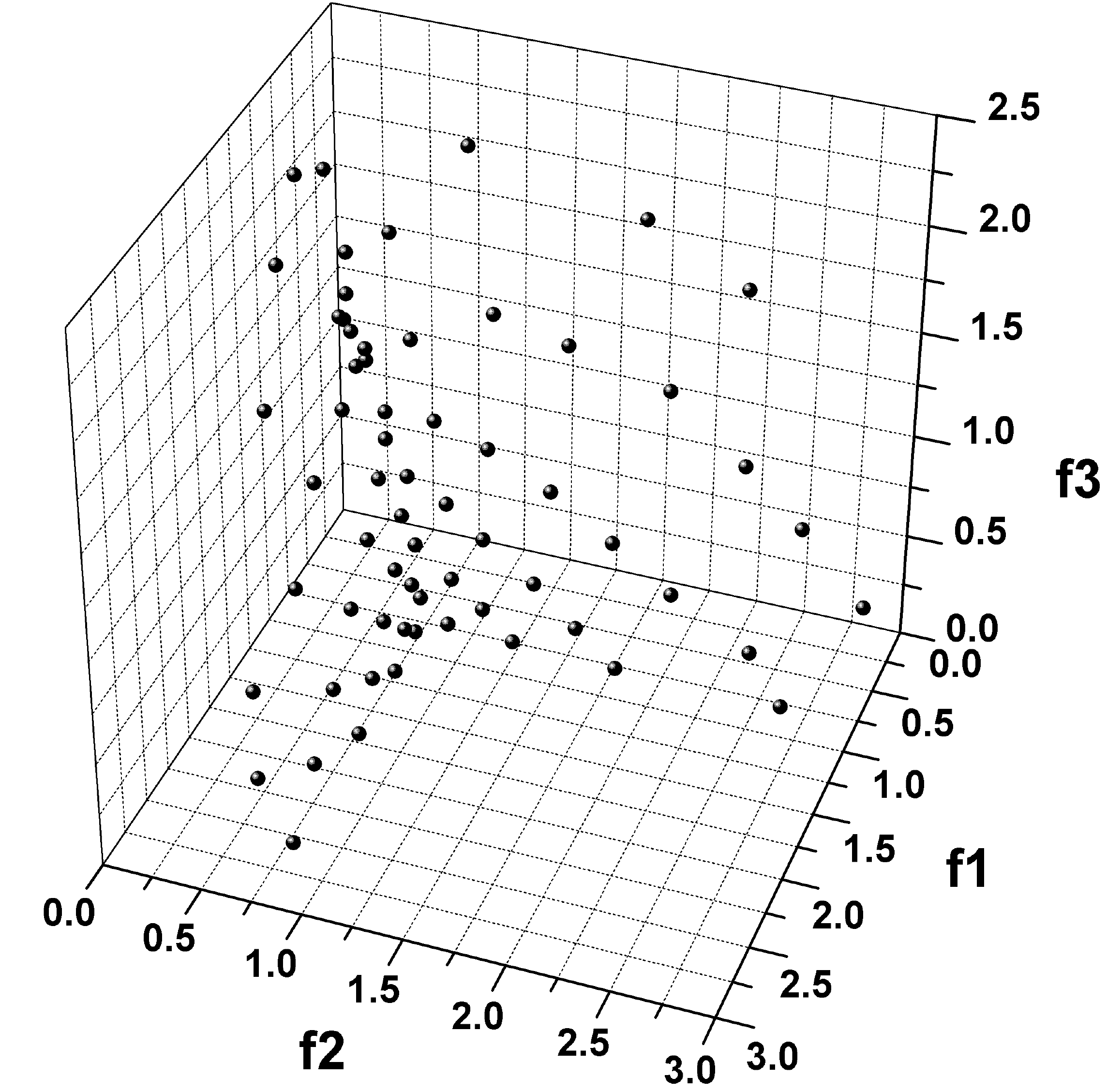}&
			~\includegraphics[scale=0.15]{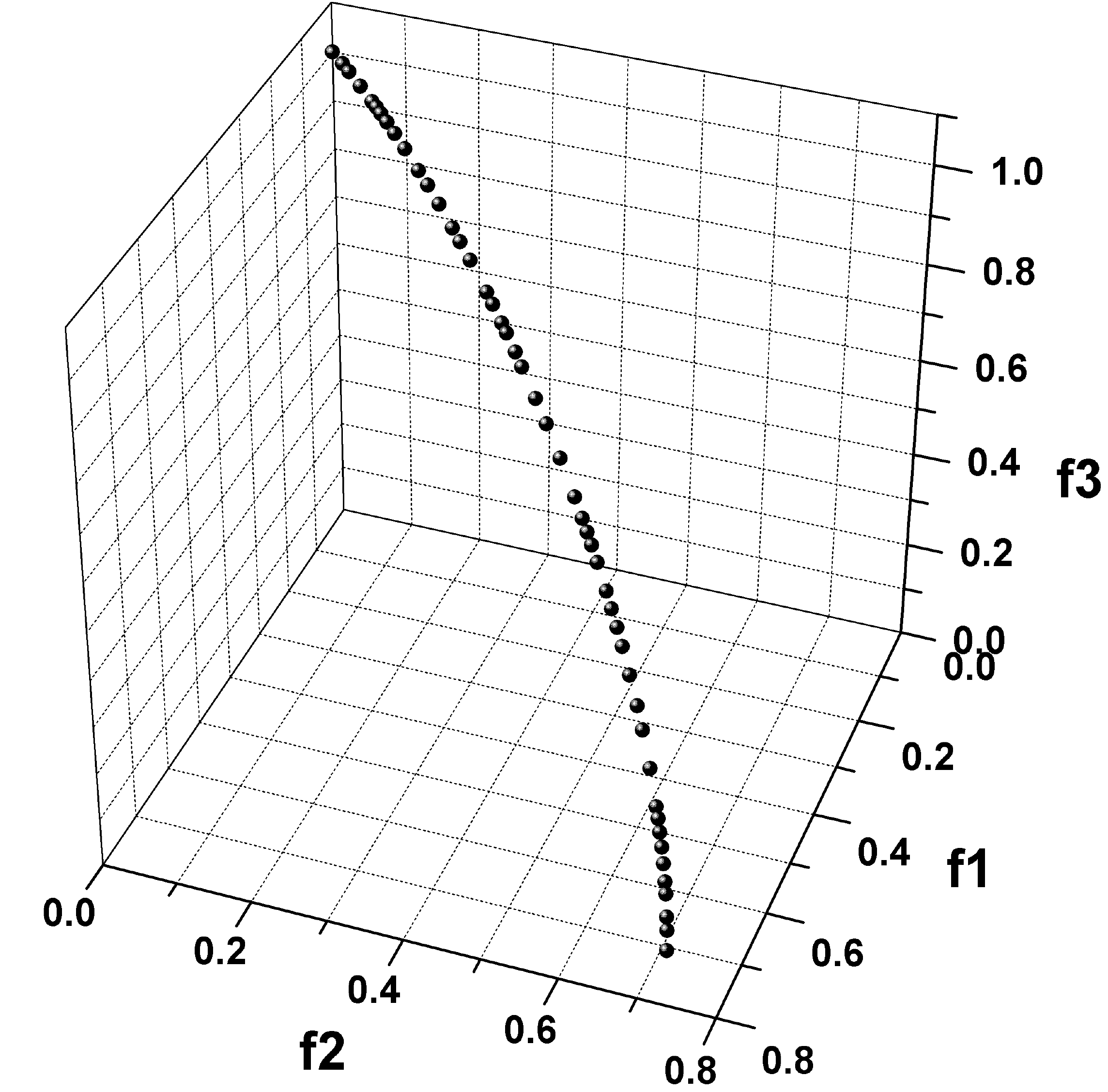}&
			~\includegraphics[scale=0.15]{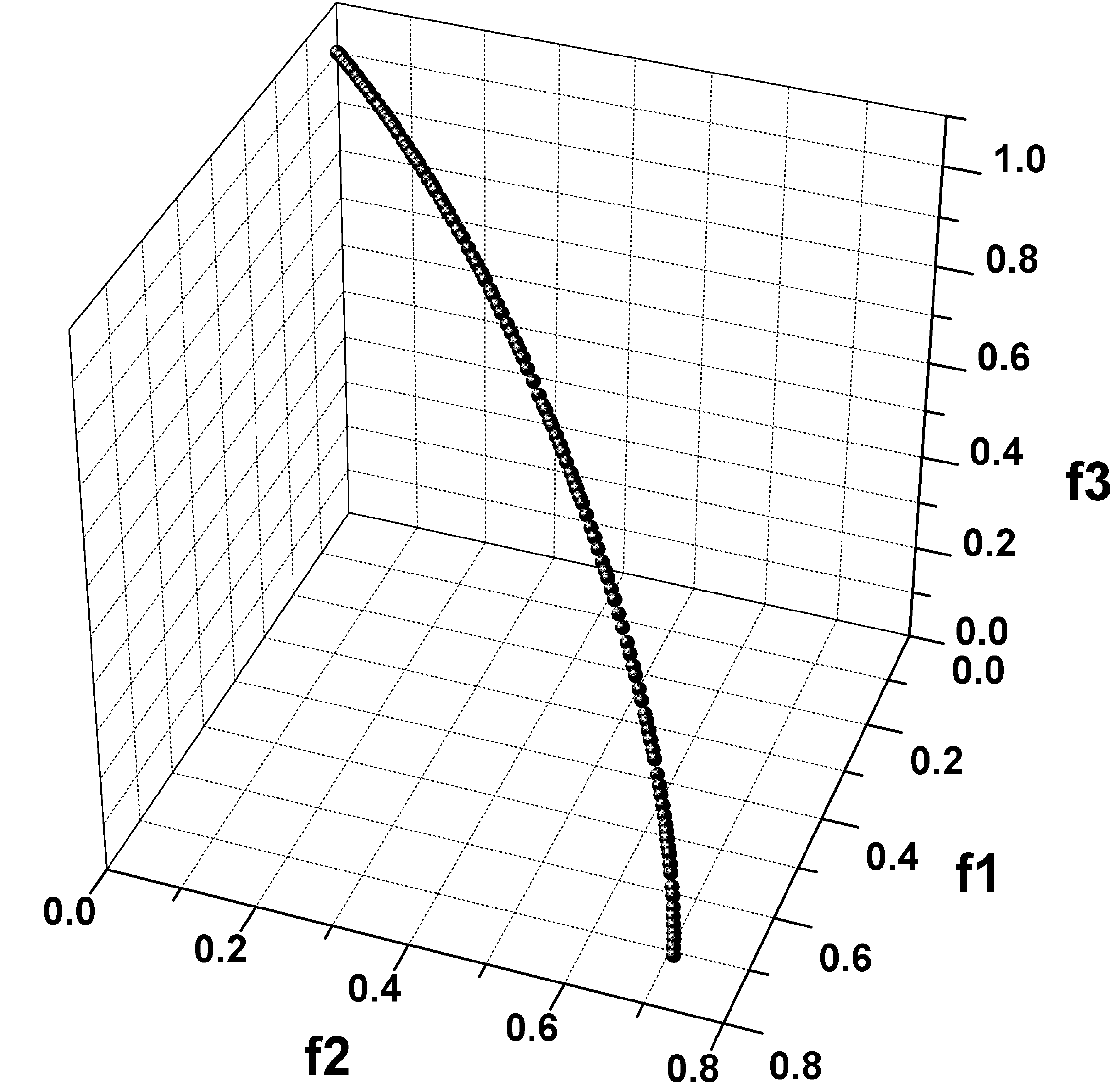}\\
			(a) MOEA/D & (b) A-NSGA-III & (c) RVEA & (d) MOEA/D-AWA & (e) AdaW \\
		\end{tabular}
	\end{center}
	\caption{The final solution set of the five algorithms on DTLZ5.}
	\label{Fig:DTLZ5}
\end{figure*}

\begin{figure*}[tbp]
	\begin{center}
		\footnotesize
		\begin{tabular}{@{}c@{}c@{}c@{}c@{}c@{}}
			\includegraphics[scale=0.14]{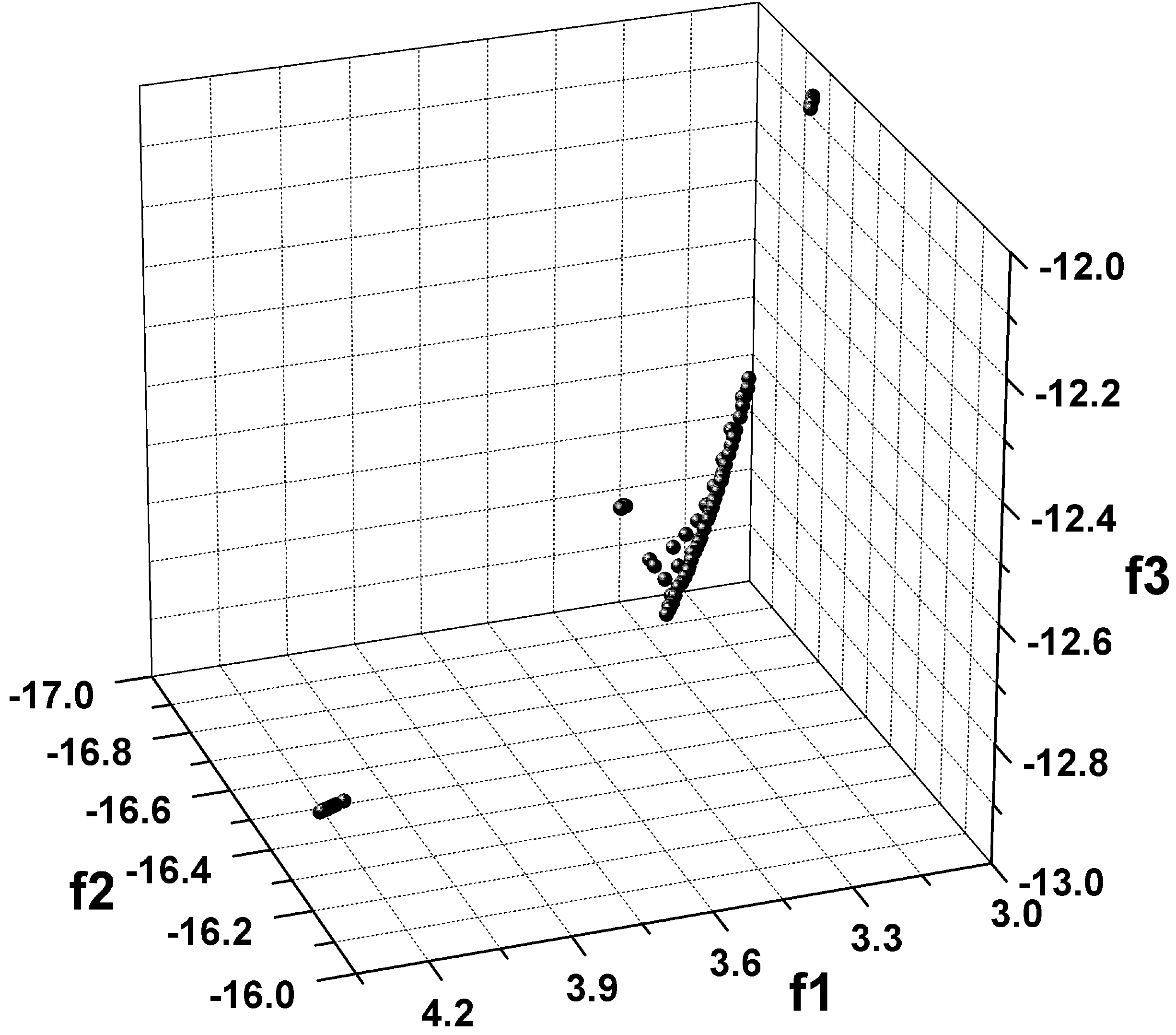}&
			\includegraphics[scale=0.14]{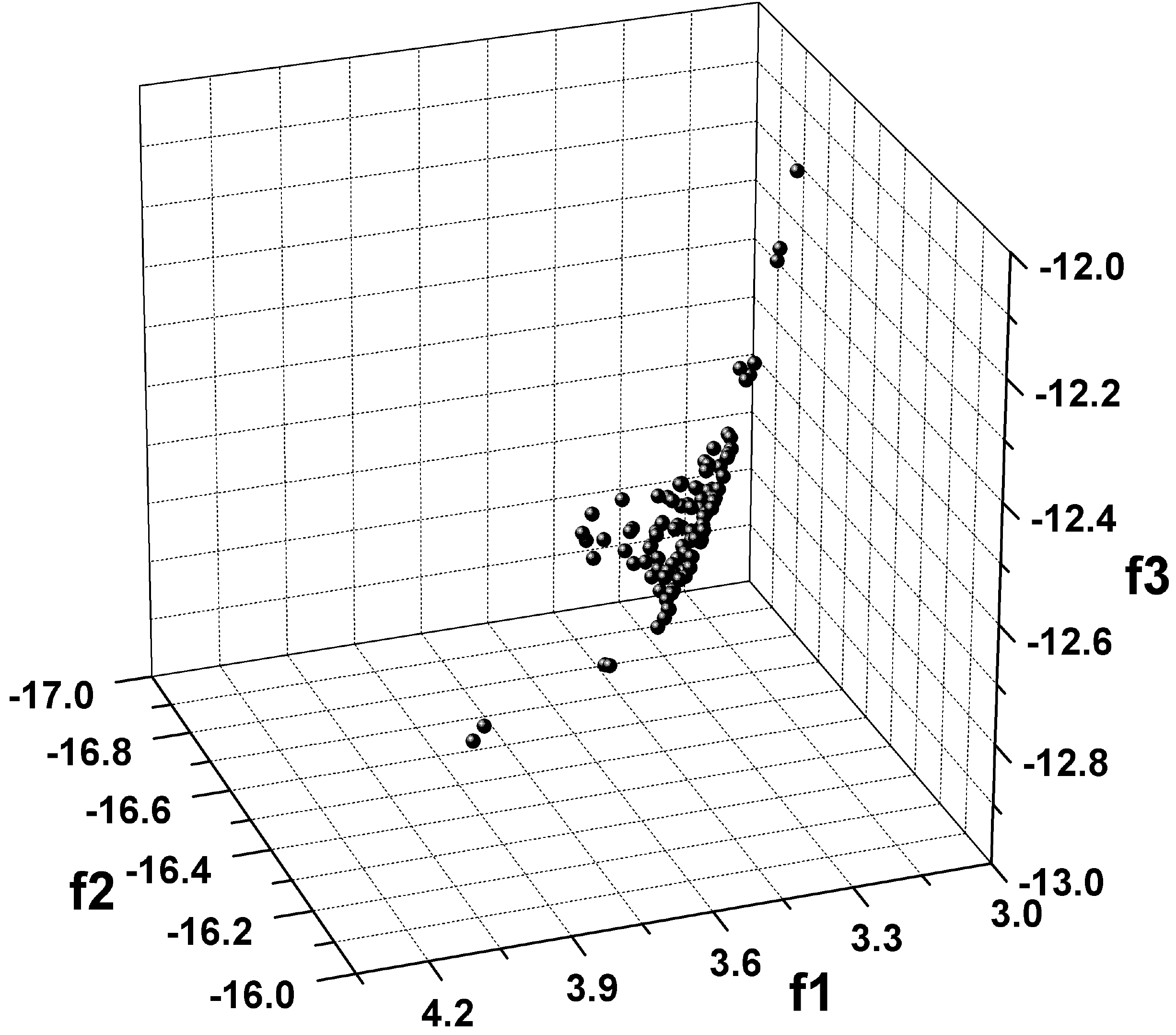}&
			\includegraphics[scale=0.14]{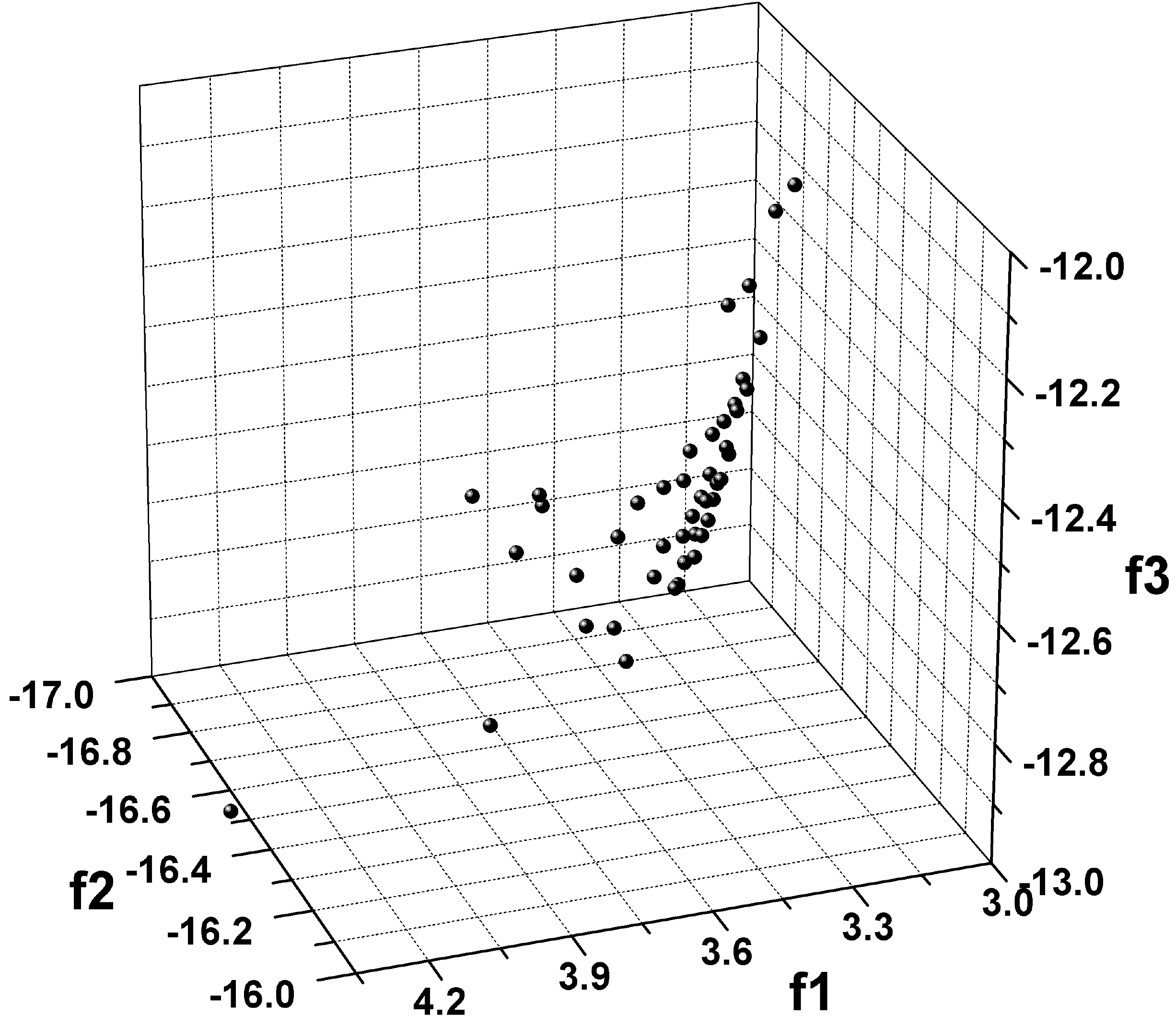}&
			\includegraphics[scale=0.14]{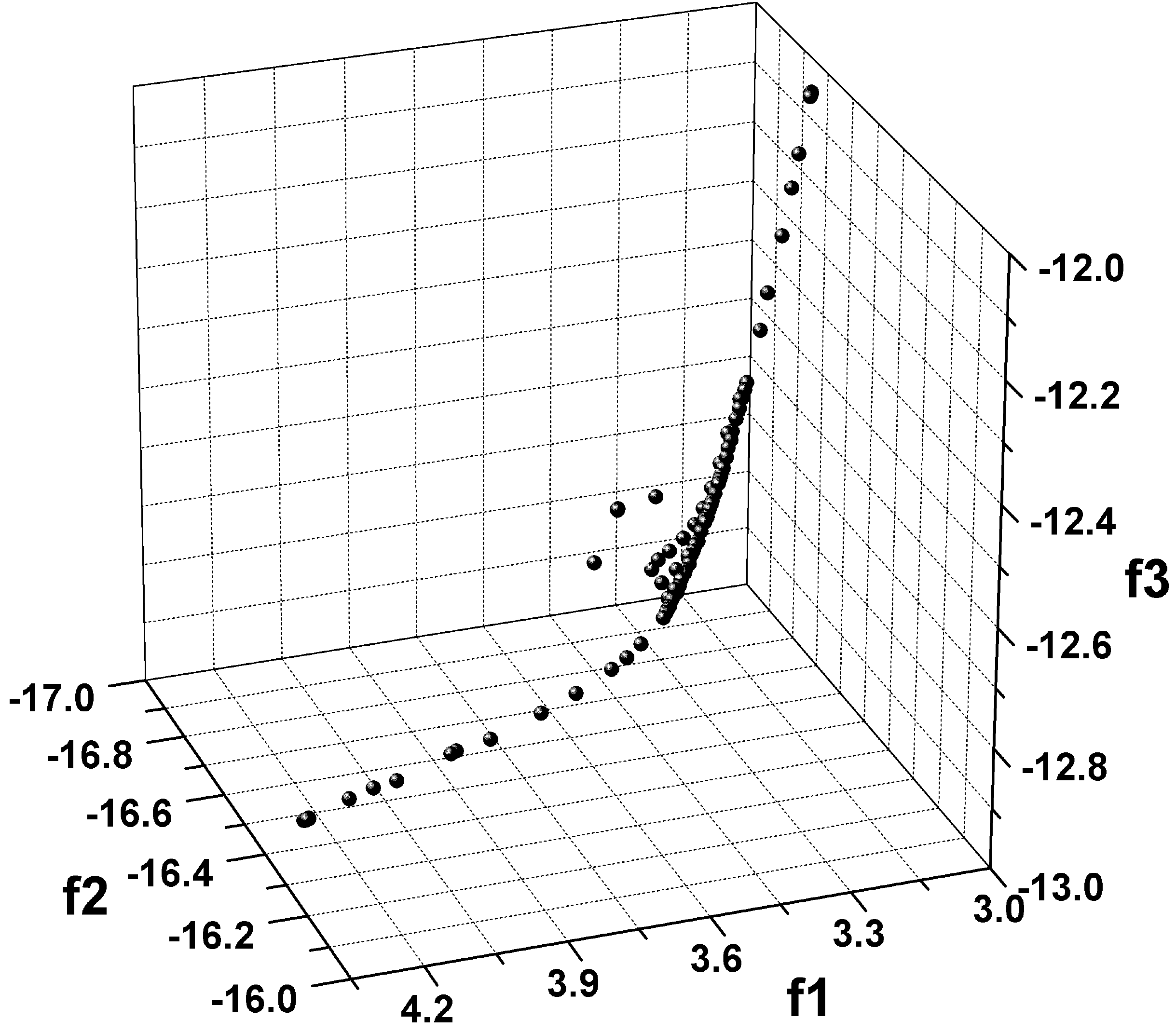}&
			\includegraphics[scale=0.14]{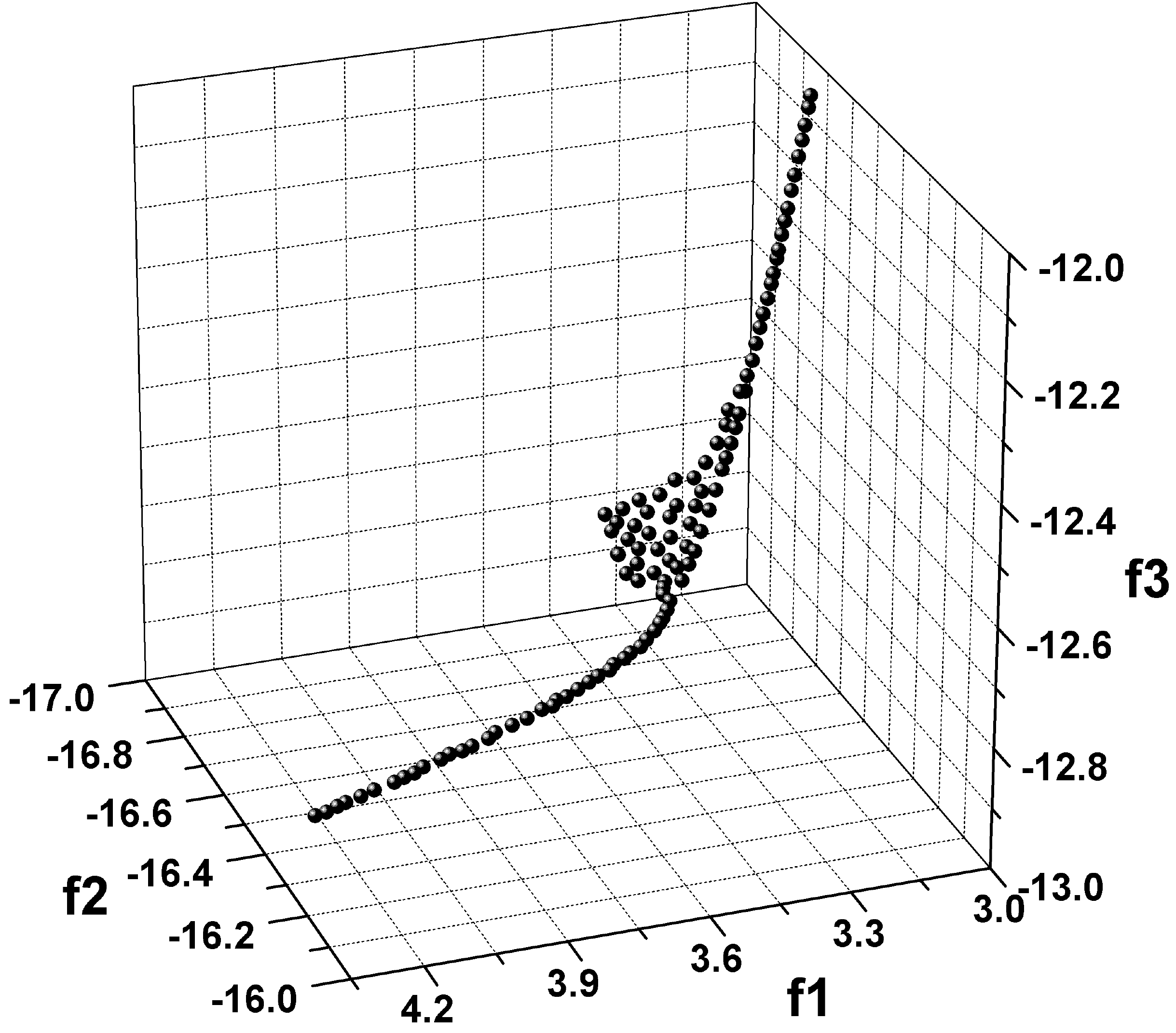}\\
			(a) MOEA/D & (b) A-NSGA-III & (c) RVEA & (d) MOEA/D-AWA & (e) AdaW \\
		\end{tabular}
	\end{center}
	\caption{The final solution set of the five algorithms on VNT2.}
	\label{Fig:VNT2}
\end{figure*}

\subsection{On Disconnected Pareto Fronts}

\mbox{Figures~\ref{Fig:ZDT3} and \ref{Fig:DTLZ7}} plot the final solution set 
of the five algorithms on ZDT3 and DTLZ7,
respectively.
On ZDT3, 
only AdaW and A-NSGA-III can maintain a good distribution of the solution set. 
MOEA/D and MOEA/D-AWA show a similar pattern, 
with their solutions distributed sparsely on the upper-left part of the Pareto front.
The set obtained by RVEA has many dominated solutions.
On DTLZ7, 
only the proposed algorithm works well.
The peer algorithms either fail to lead their solutions to cover the Pareto front 
( MOEA/D and MOEA/D-AWA), 
struggle to maintain the uniformity (A-NSGA-III), 
or produce some dominated solutions (RVEA).

\subsection{On Degenerate Pareto Fronts}

Problems with a degenerate Pareto front poses a big challenge to decomposition-based approaches 
since the ideal weight vector set is located in a lower-dimensional manifold than its initial setting~\cite{Li2017}.
On this group of problems,
the proposed algorithm has shown a significant advantage over its competitors (see \mbox{Figures~\ref{Fig:DTLZ5} and \ref{Fig:VNT2}}).
It is worth noting that VNT2 has a mixed Pareto front, 
with both ends degenerating into two curves and the middle part being a triangle-like plane.
As can be seen from \mbox{Figure~\ref{Fig:VNT2}},
the solution set of AdaW has a good distribution over the whole Pareto front.

\subsection{On Badly-Scaled Pareto Fronts}

\begin{figure*}[tbp]
	\begin{center}
		\footnotesize
		\begin{tabular}{@{}c@{}c@{}c@{}c@{}c@{}}
			\includegraphics[scale=0.14]{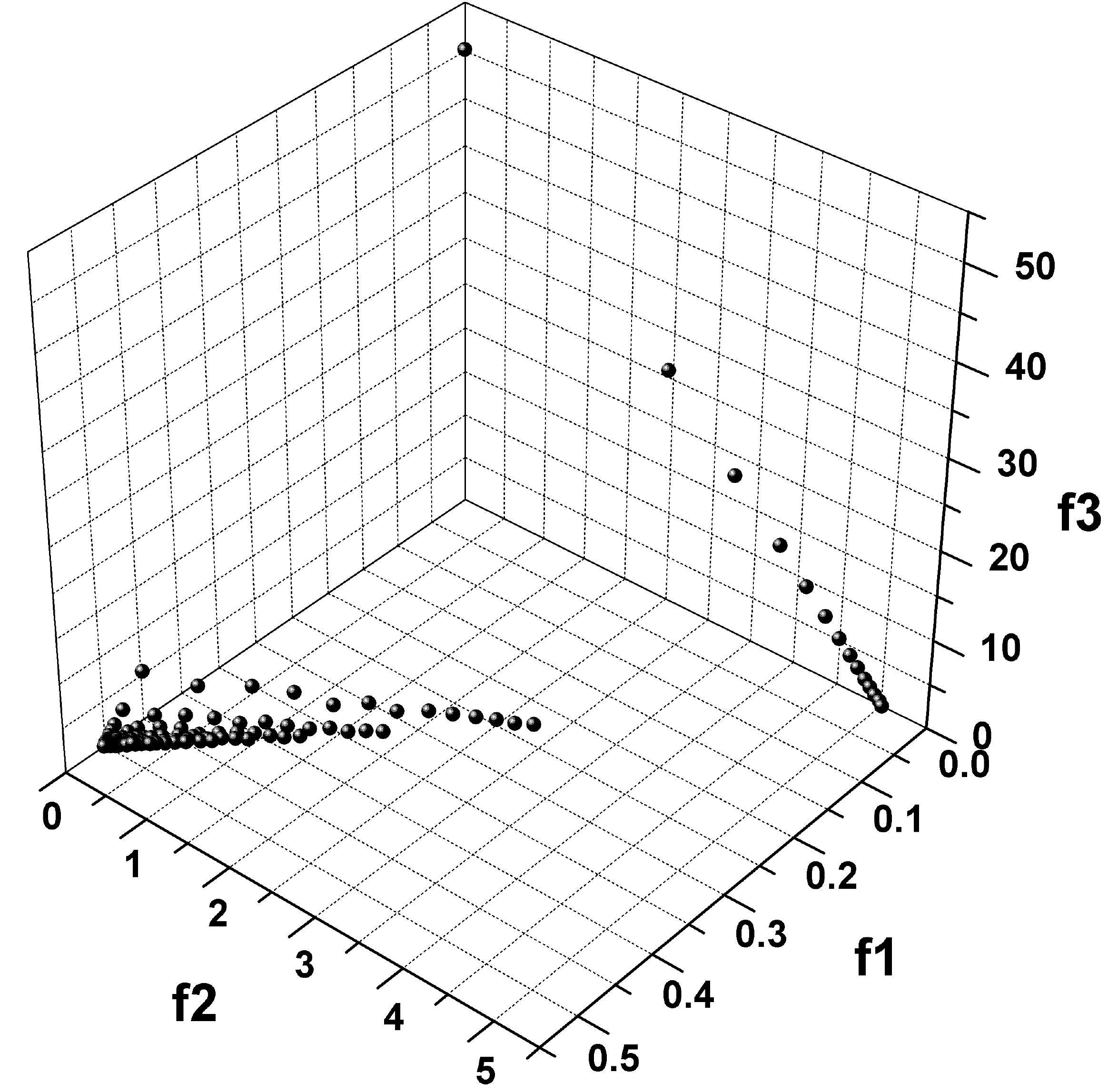}~~&
			~\includegraphics[scale=0.14]{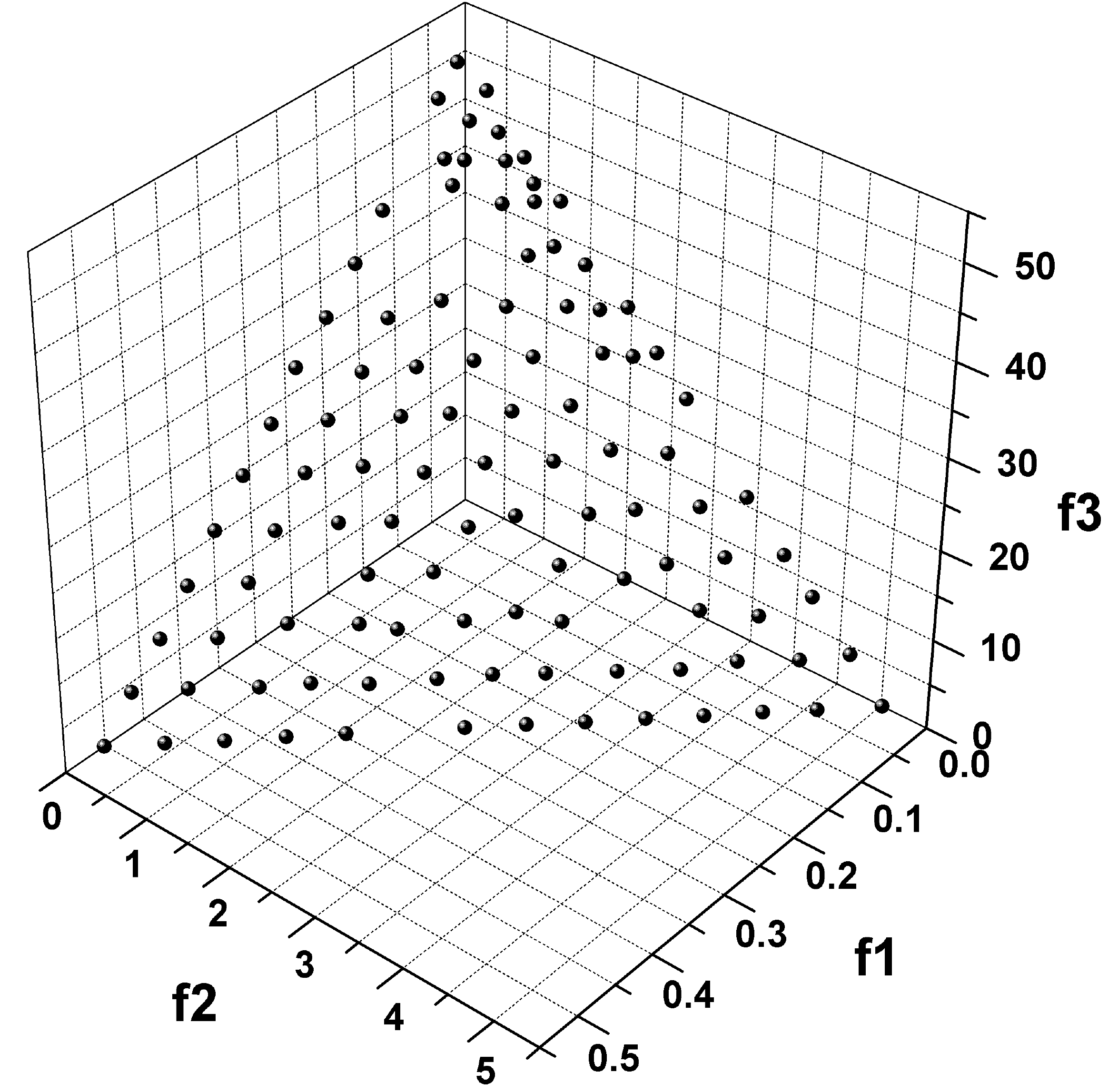}~~&
			~\includegraphics[scale=0.14]{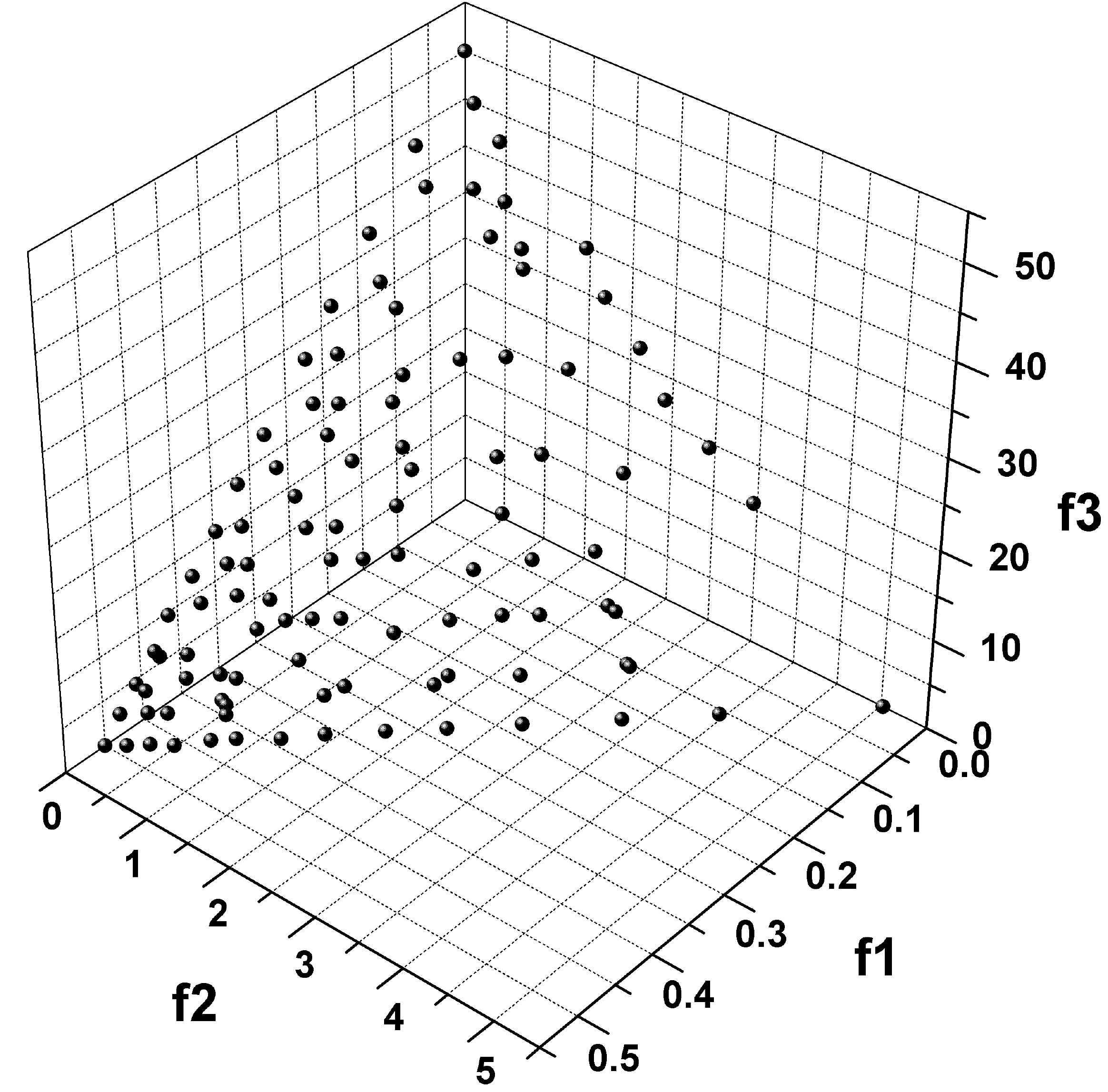}~~&
			~\includegraphics[scale=0.14]{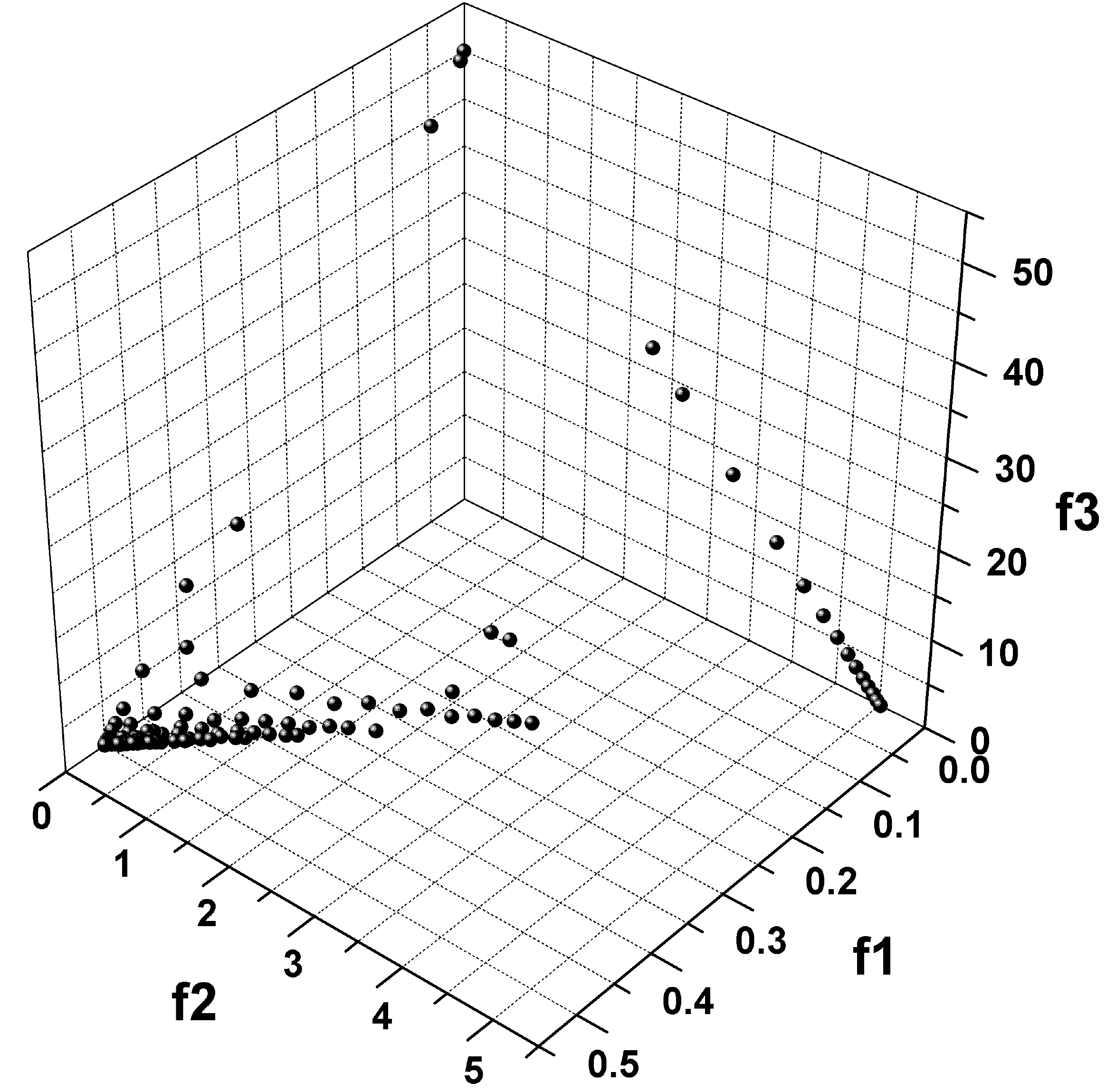}~~&
			~\includegraphics[scale=0.14]{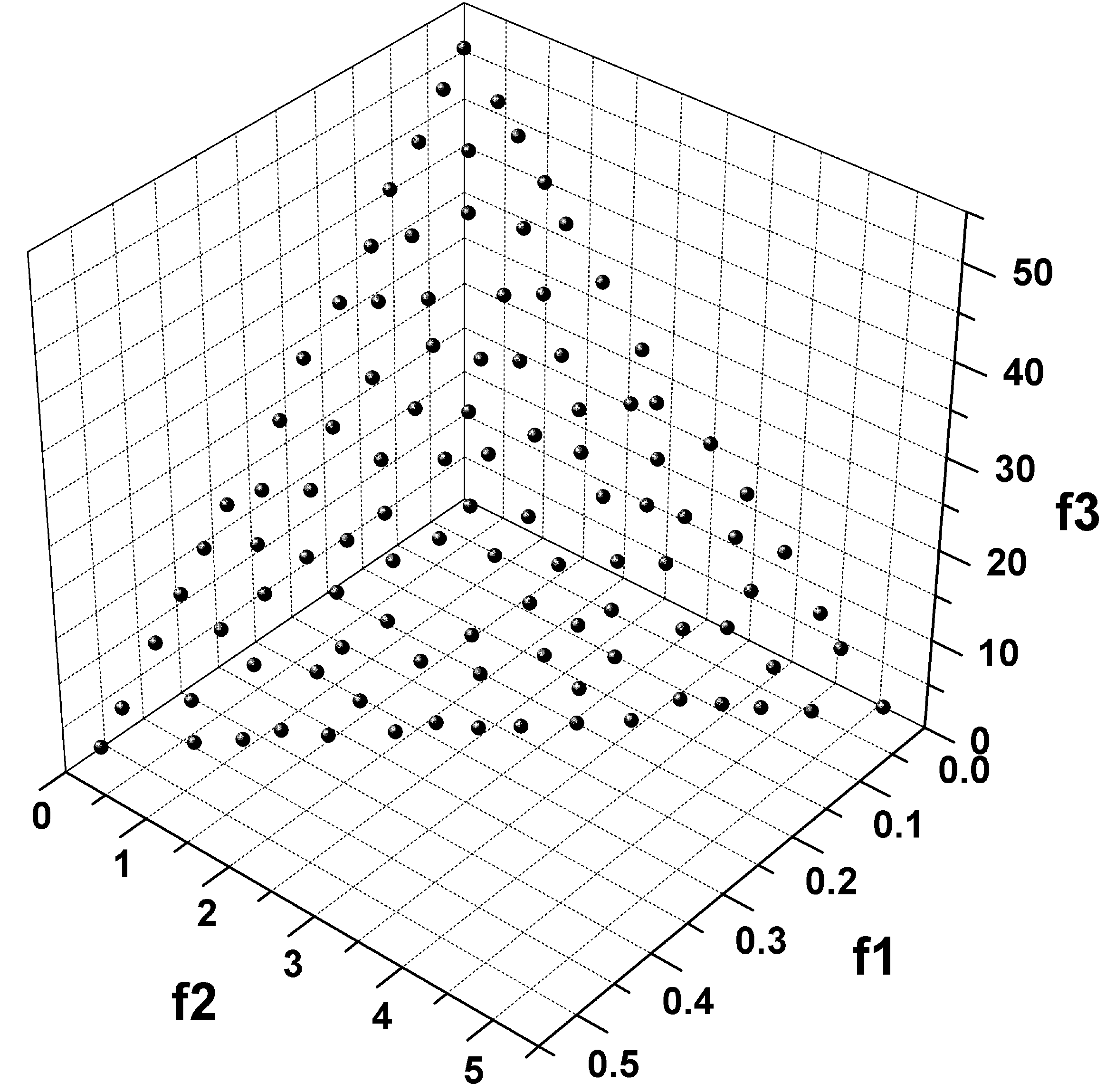}\\
			(a) MOEA/D & (b) A-NSGA-III & (c) RVEA & (d) MOEA/D-AWA & (e) AdaW \\
		\end{tabular}
	\end{center}
	\caption{The final solution set of the five algorithms on the scaled DTLZ1.}
	\label{Fig:SDTLZ1}
\end{figure*}

\begin{figure*}[tbp]
	\begin{center}
		\footnotesize
		\begin{tabular}{@{}c@{}c@{}c@{}c@{}c@{}}
			\includegraphics[scale=0.14]{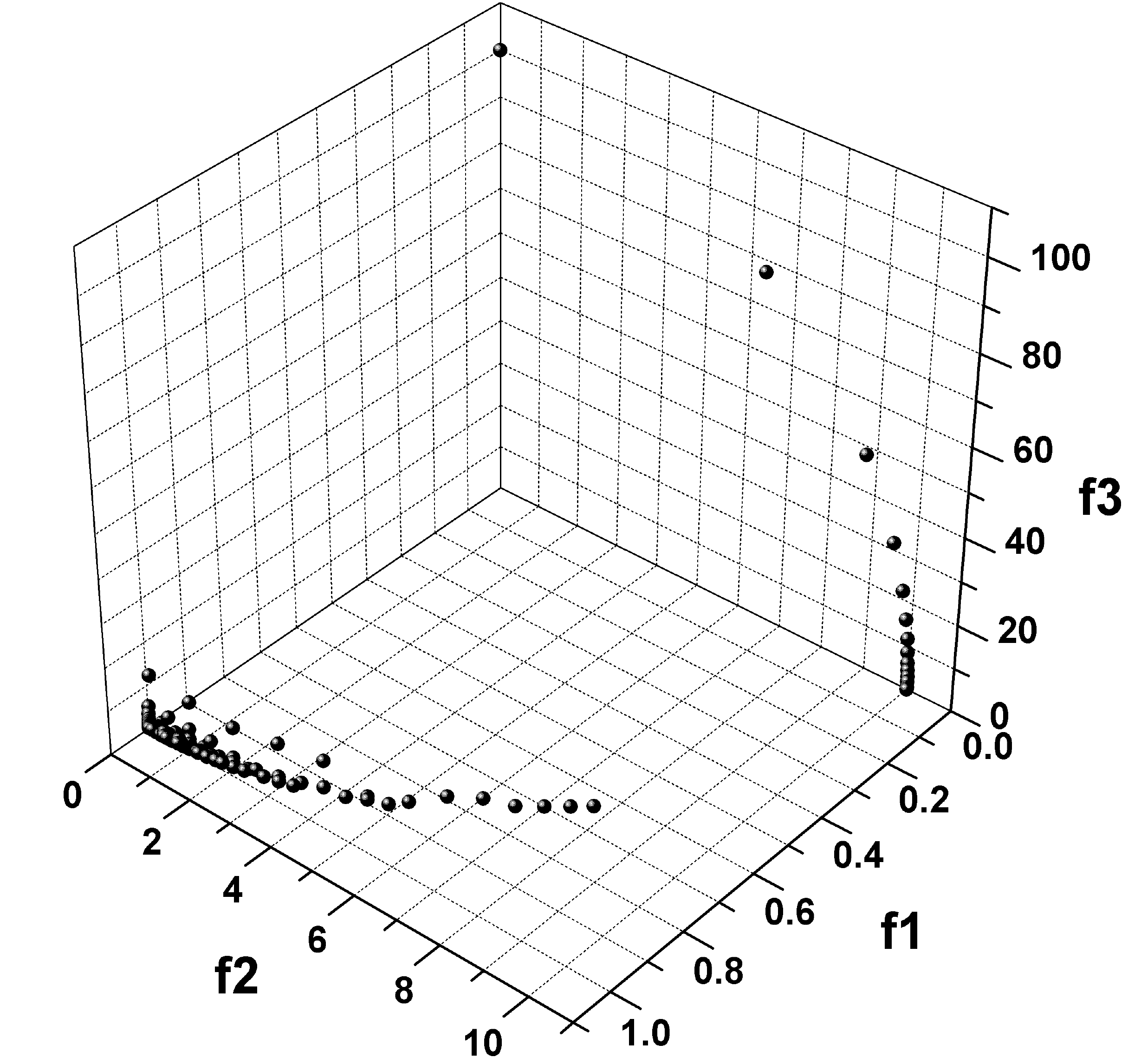}~&
			~\includegraphics[scale=0.14]{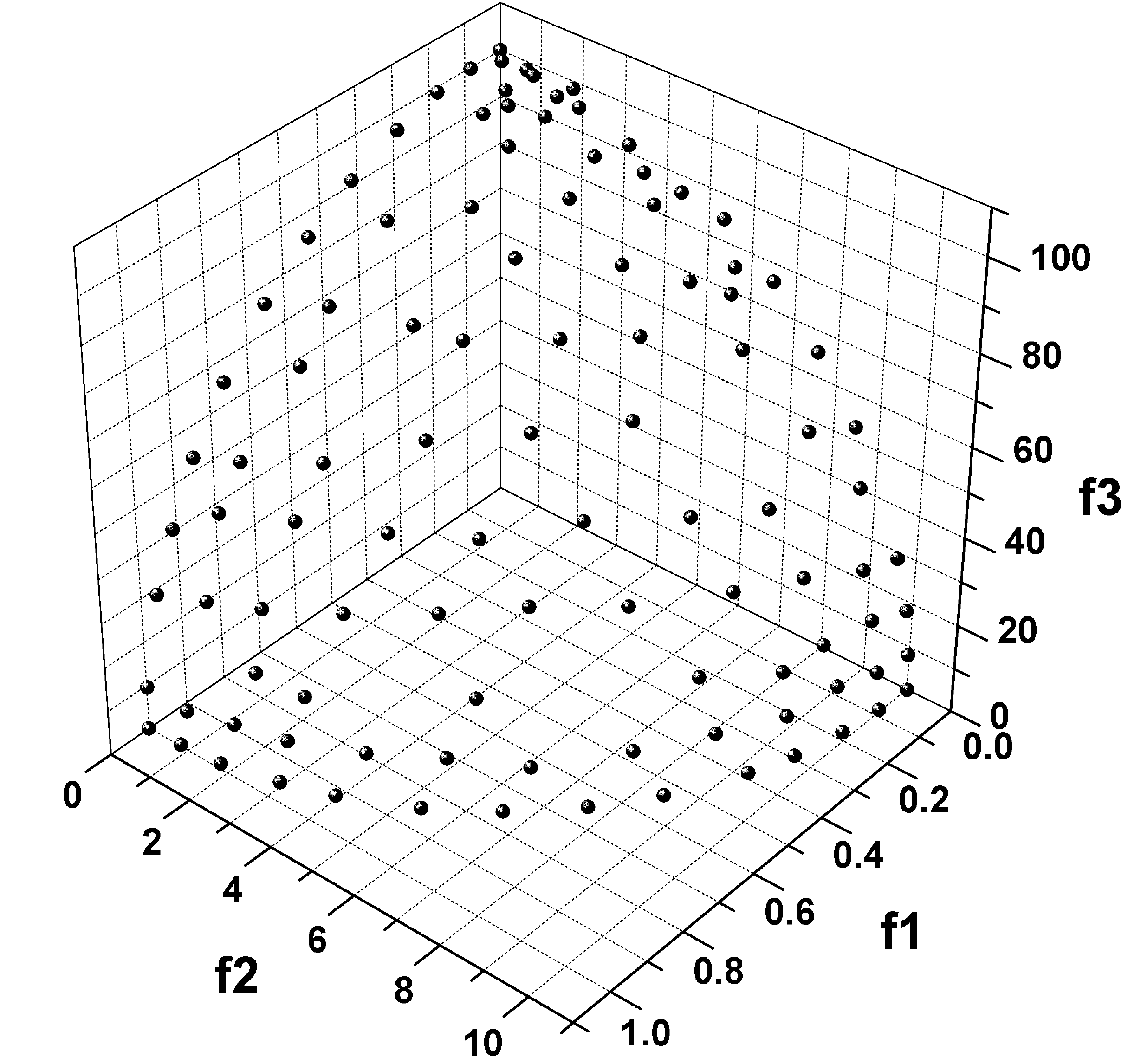}~&
			~\includegraphics[scale=0.14]{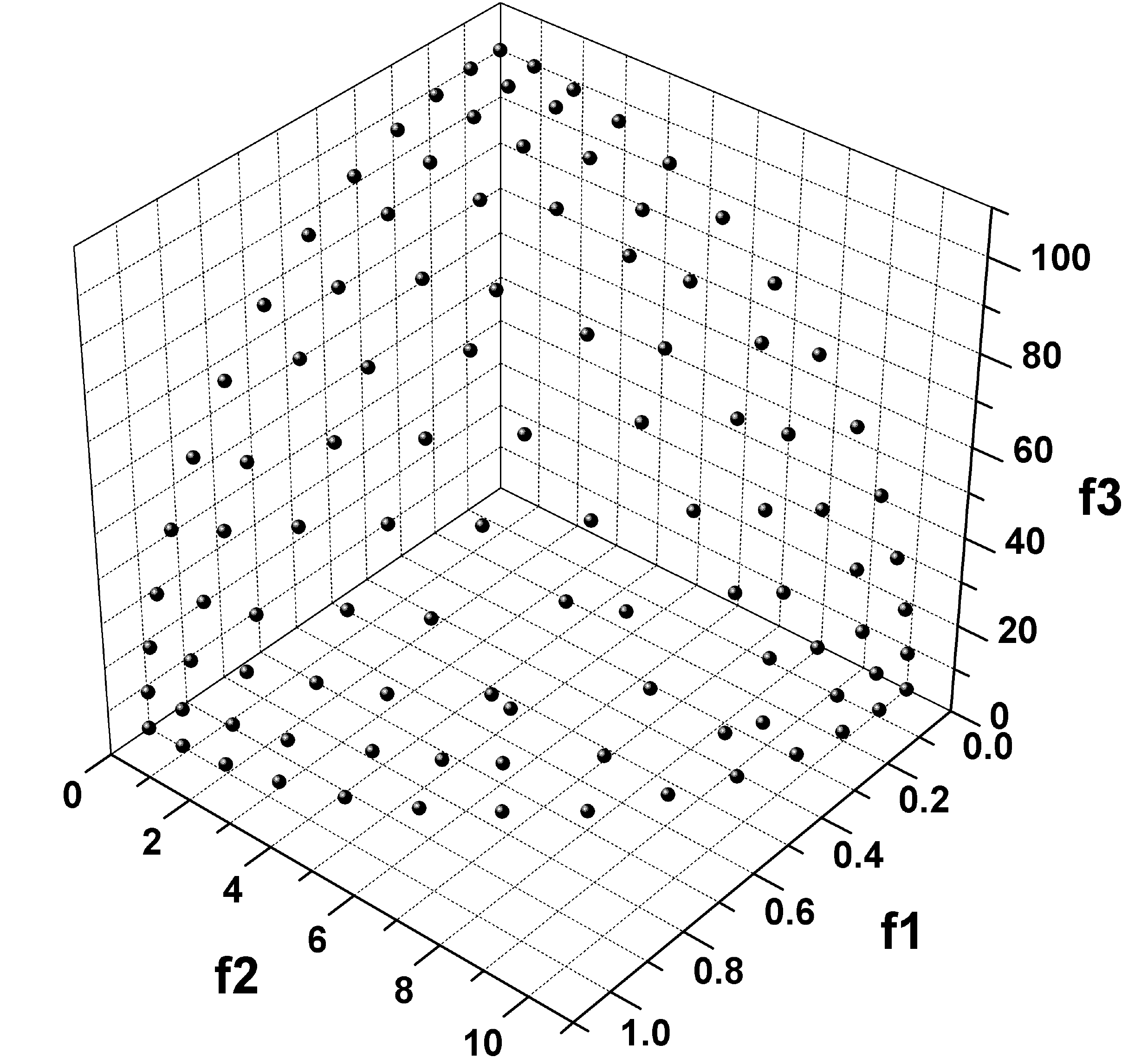}~&
			~\includegraphics[scale=0.14]{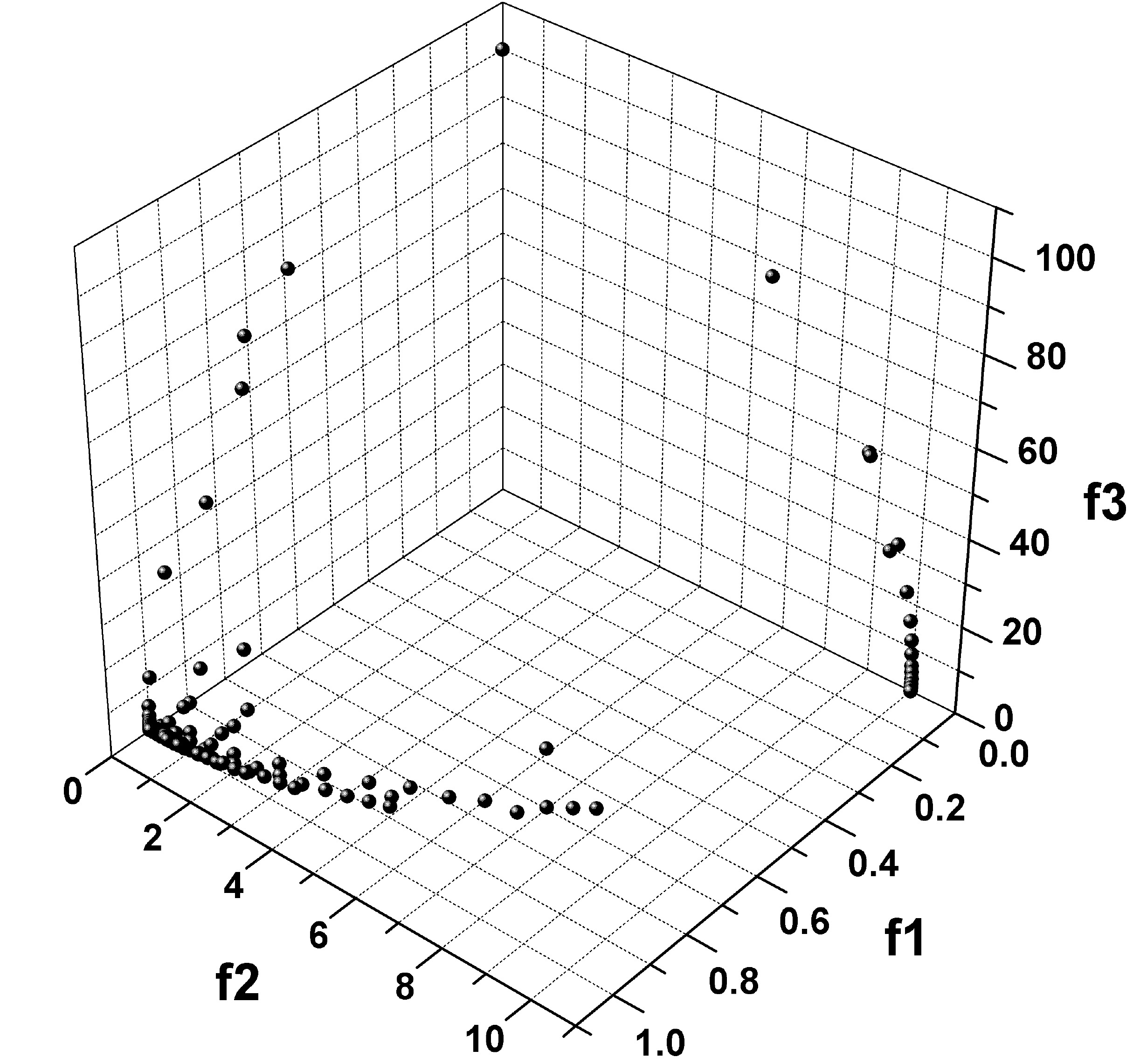}~&
			~\includegraphics[scale=0.14]{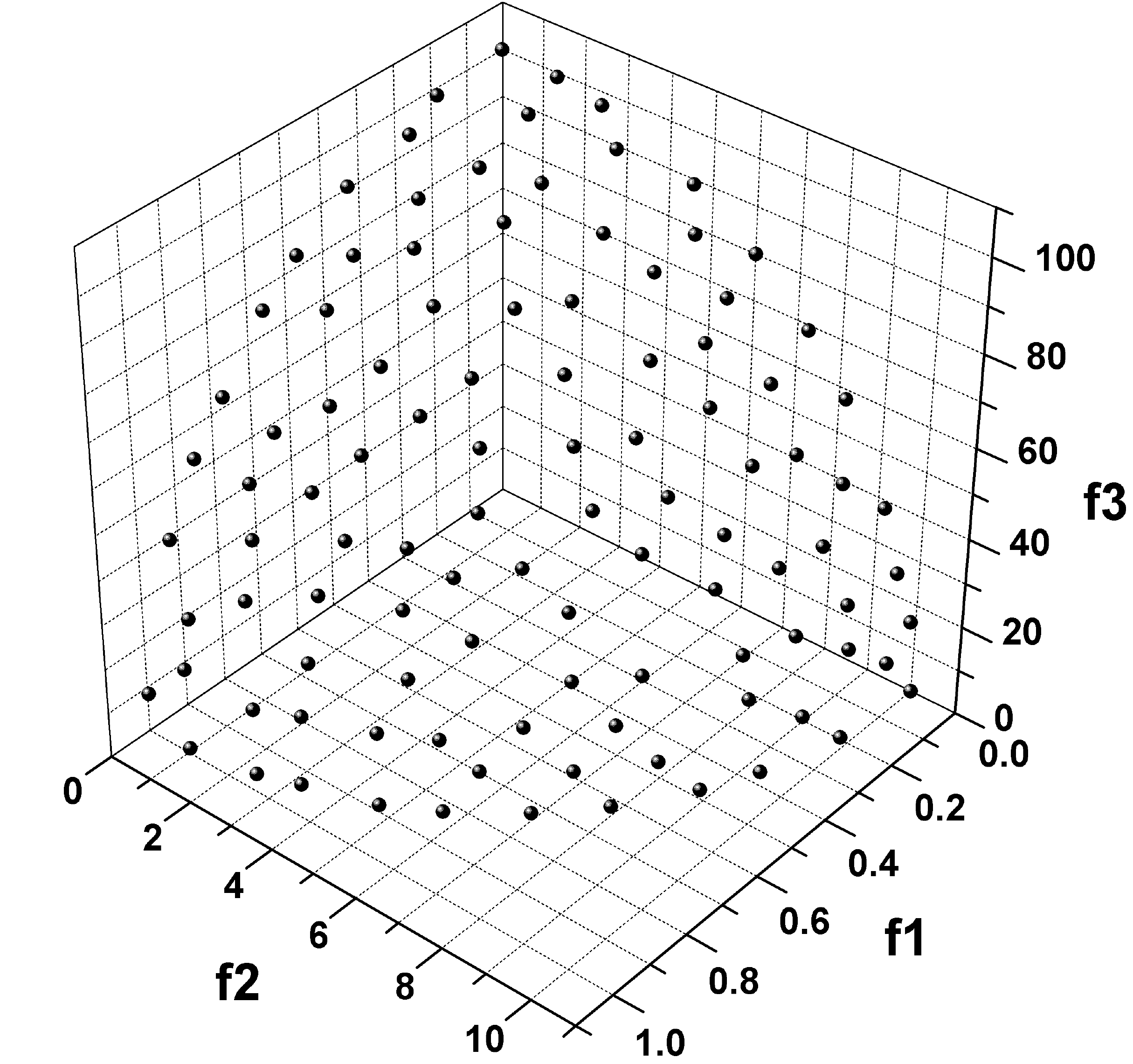}\\
			(a) MOEA/D & (b) A-NSGA-III & (c) RVEA & (d) MOEA/D-AWA & (e) AdaW \\
		\end{tabular}
	\end{center}
	\caption{The final solution set of the five algorithms on the scaled DTLZ2.}
	\label{Fig:SDTLZ2}
\end{figure*}

\begin{figure*}[!]
	\begin{center}
		\footnotesize
		\begin{tabular}{@{}c@{}c@{}c@{}c@{}c@{}}
			\includegraphics[scale=0.15]{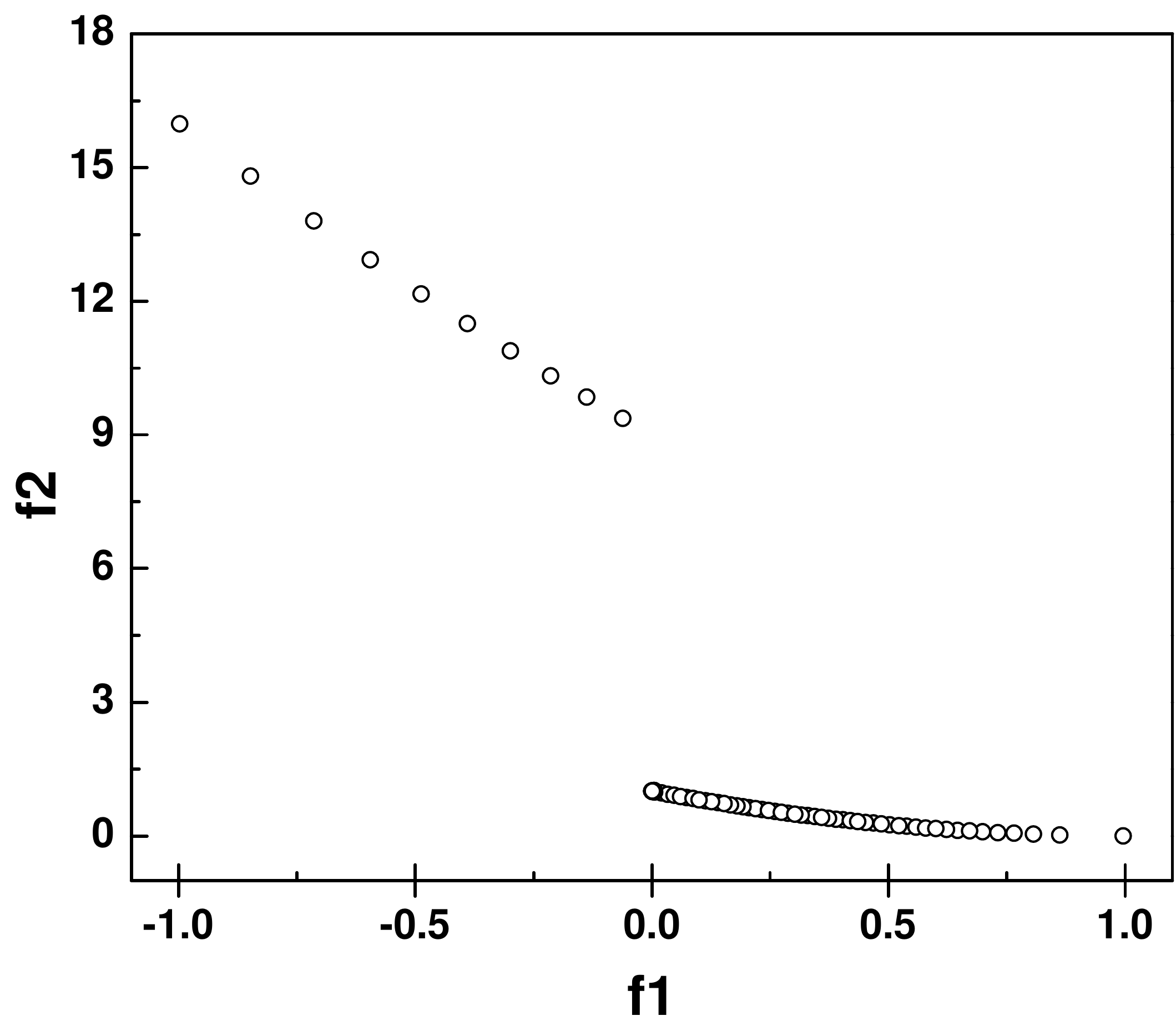}~~&
			~~\includegraphics[scale=0.15]{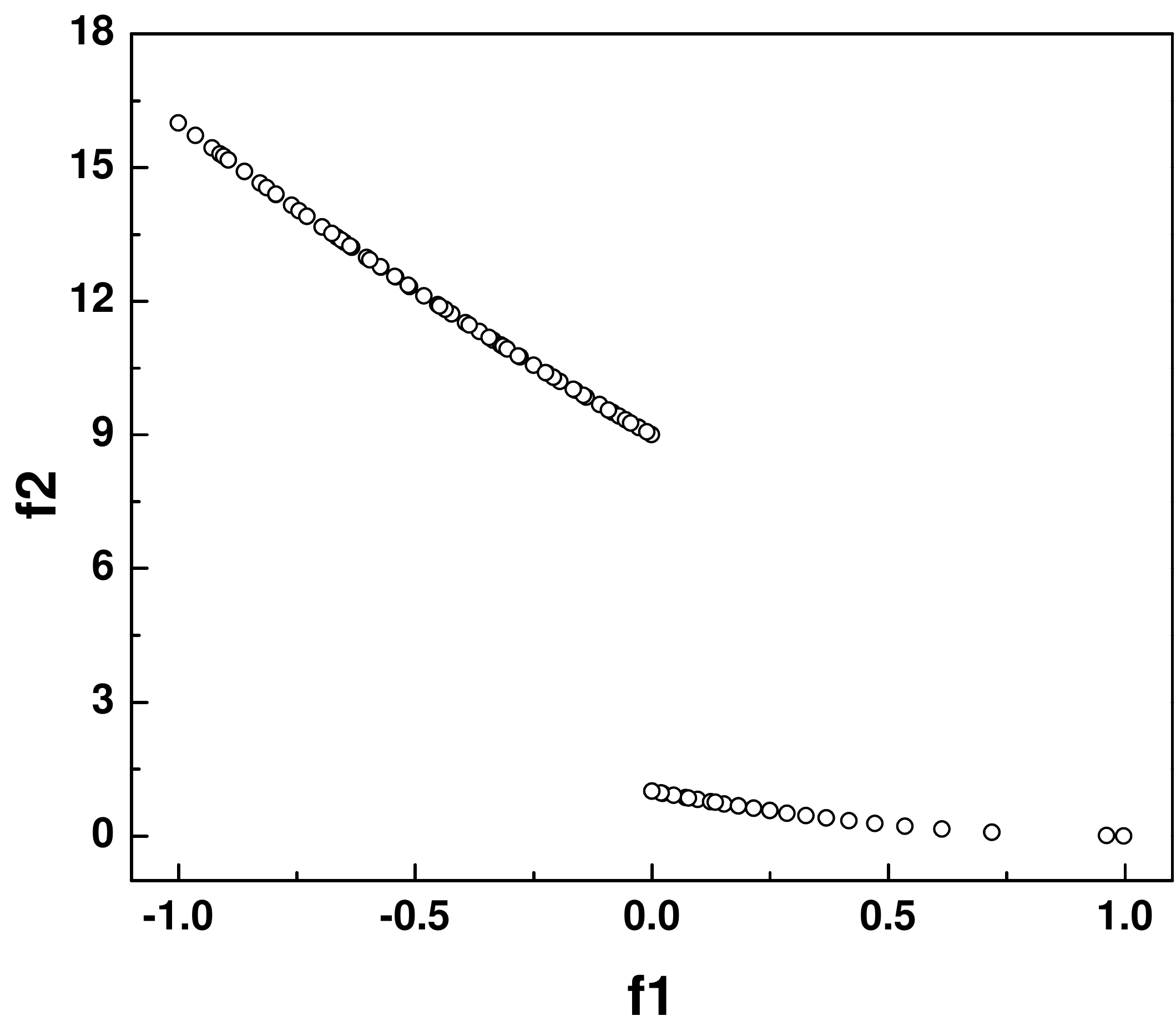}~~&
			~~\includegraphics[scale=0.15]{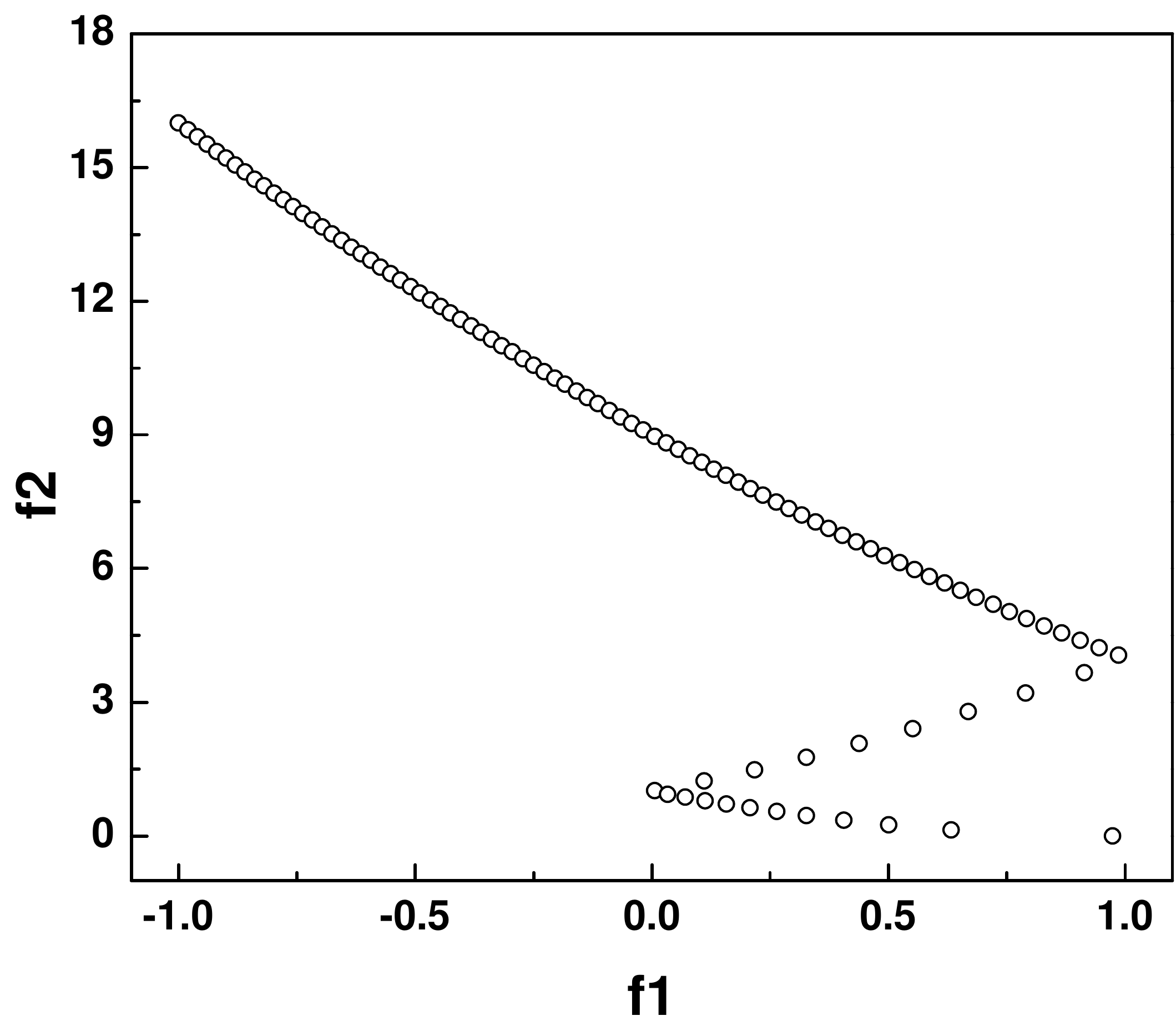}~~&
			~~\includegraphics[scale=0.15]{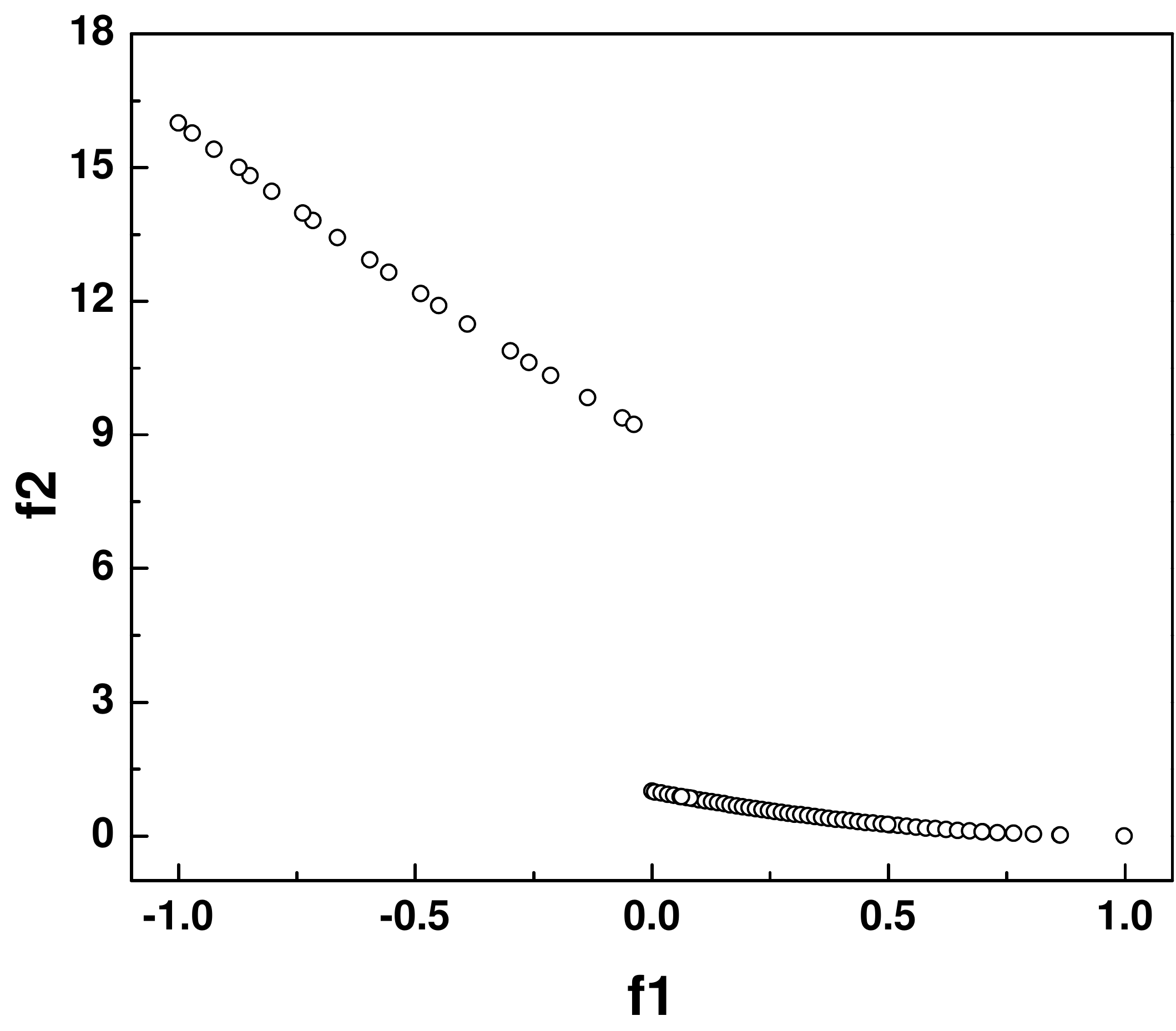}~~&
			~~\includegraphics[scale=0.15]{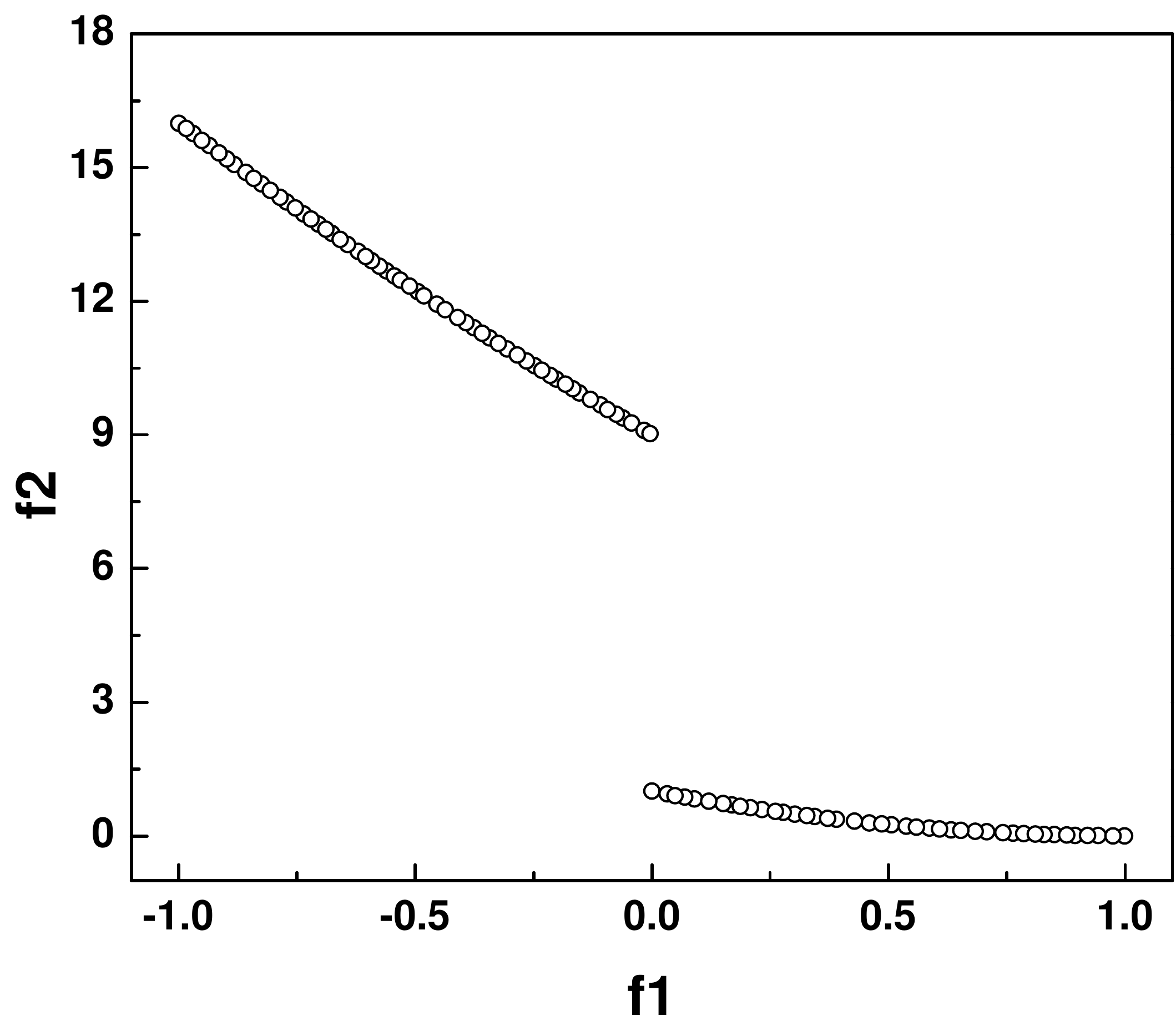}\\
			(a) MOEA/D & (b) A-NSGA-III & (c) RVEA & (d) MOEA/D-AWA & (e) AdaW \\
		\end{tabular}
	\end{center}
	\caption{The final solution set of the five algorithms on SCH2.}
	\label{Fig:SCH2}
\end{figure*}

\mbox{Figures~\ref{Fig:SDTLZ1}--\ref{Fig:SCH2}} plot the final solution set 
of the five algorithms on SDTLZ1, SDTLZ2 and SCH2,
respectively.
For the first two problems,
AdaW, A-NSGA-III and RVEA work fairly well, 
but the solutions obtained by RVEA are not so uniform as those obtained by the other two algorithms on SDTLZ1.
For SCH2 which also has a disconnected Pareto front,
AdaW significantly outperforms its competitors, 
with the solution set being uniformly distributed over the two parts of the Pareto front.

\subsection{On Many-Objective Problems}

This section evaluates the performance of the proposed AdaW on many-objective problems 
by considering three instances, the 10-objective DTLZ2, 10-objective IDTLZ1, 
and DTLZ5(2,10) where the number of objectives is 10 and 
the true Pareto front's dimensionality is 2. 

\begin{figure*}[tbp]
	\begin{center}
		\footnotesize
		\begin{tabular}{@{}c@{}c@{}c@{}c@{}c@{}}
			\includegraphics[scale=0.15]{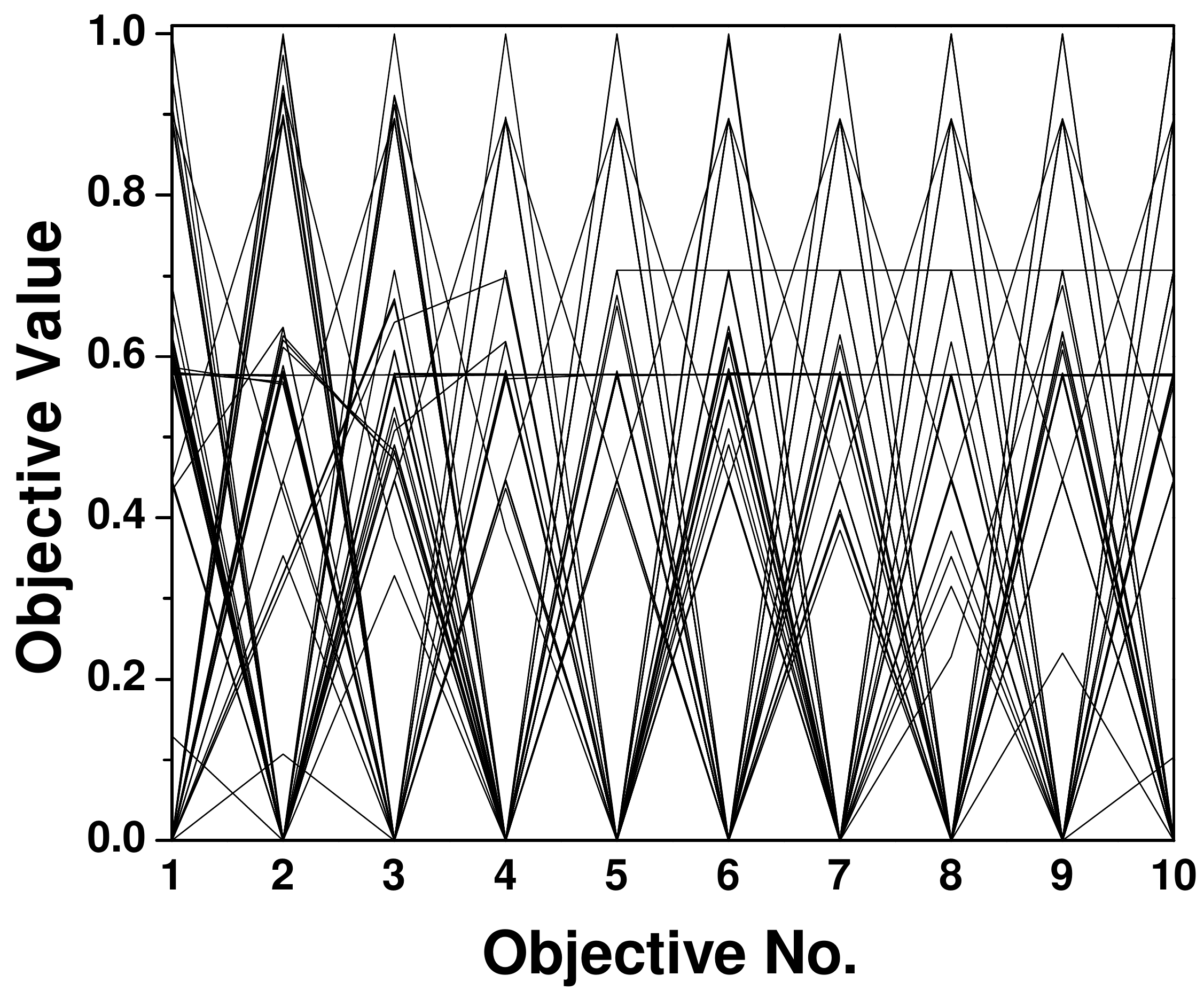}~&
			~\includegraphics[scale=0.15]{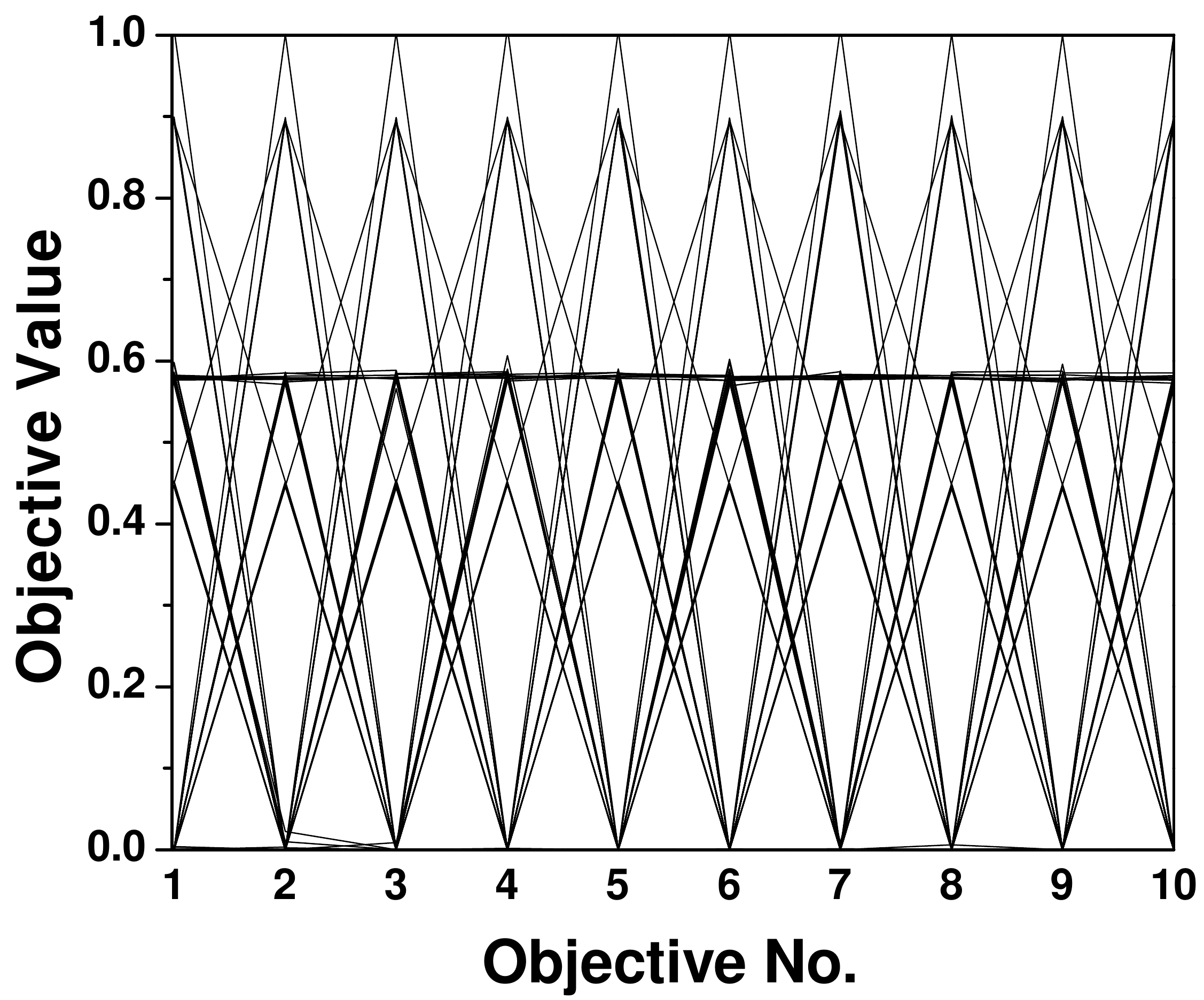}~&
			~\includegraphics[scale=0.15]{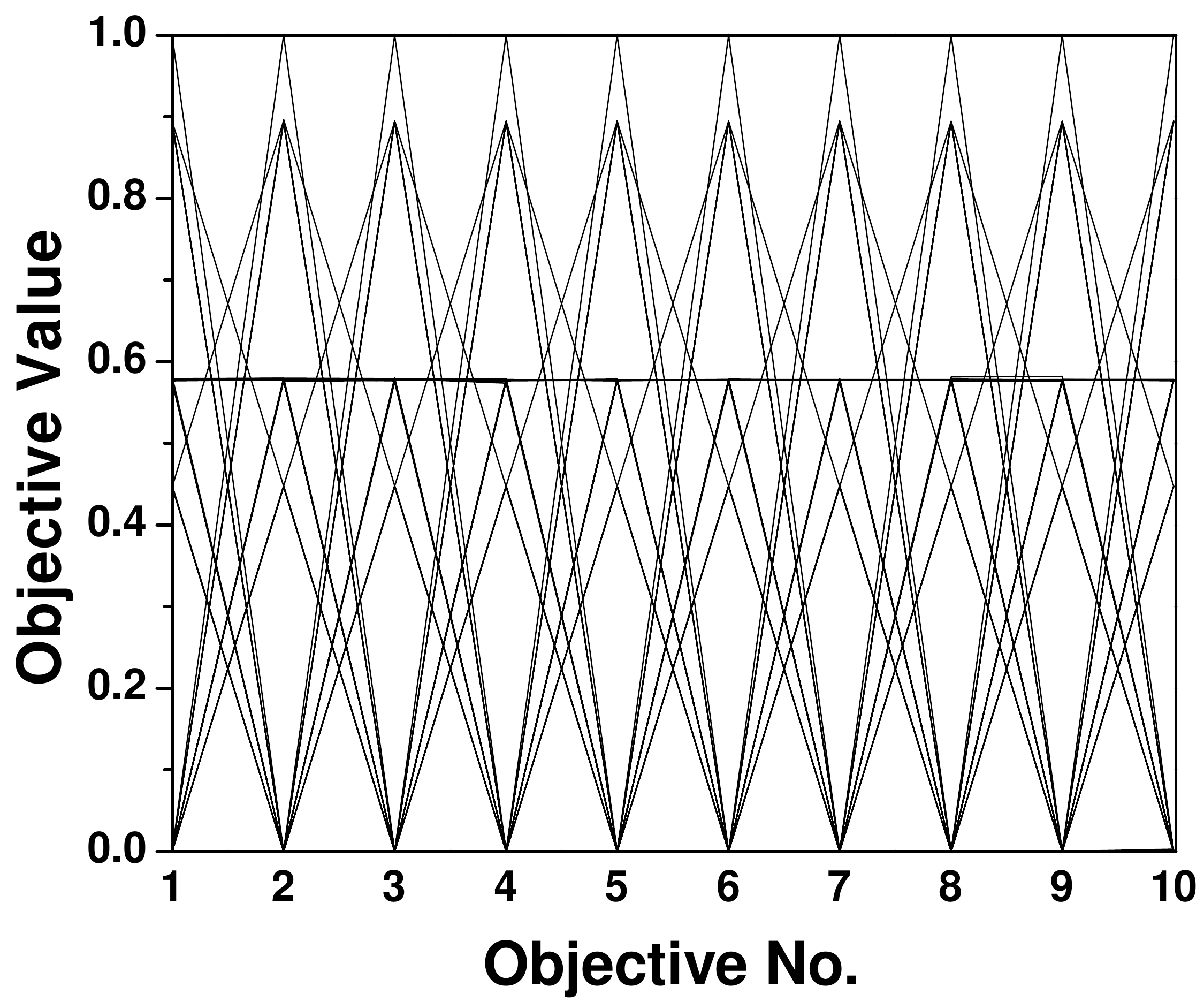}~&
			~\includegraphics[scale=0.15]{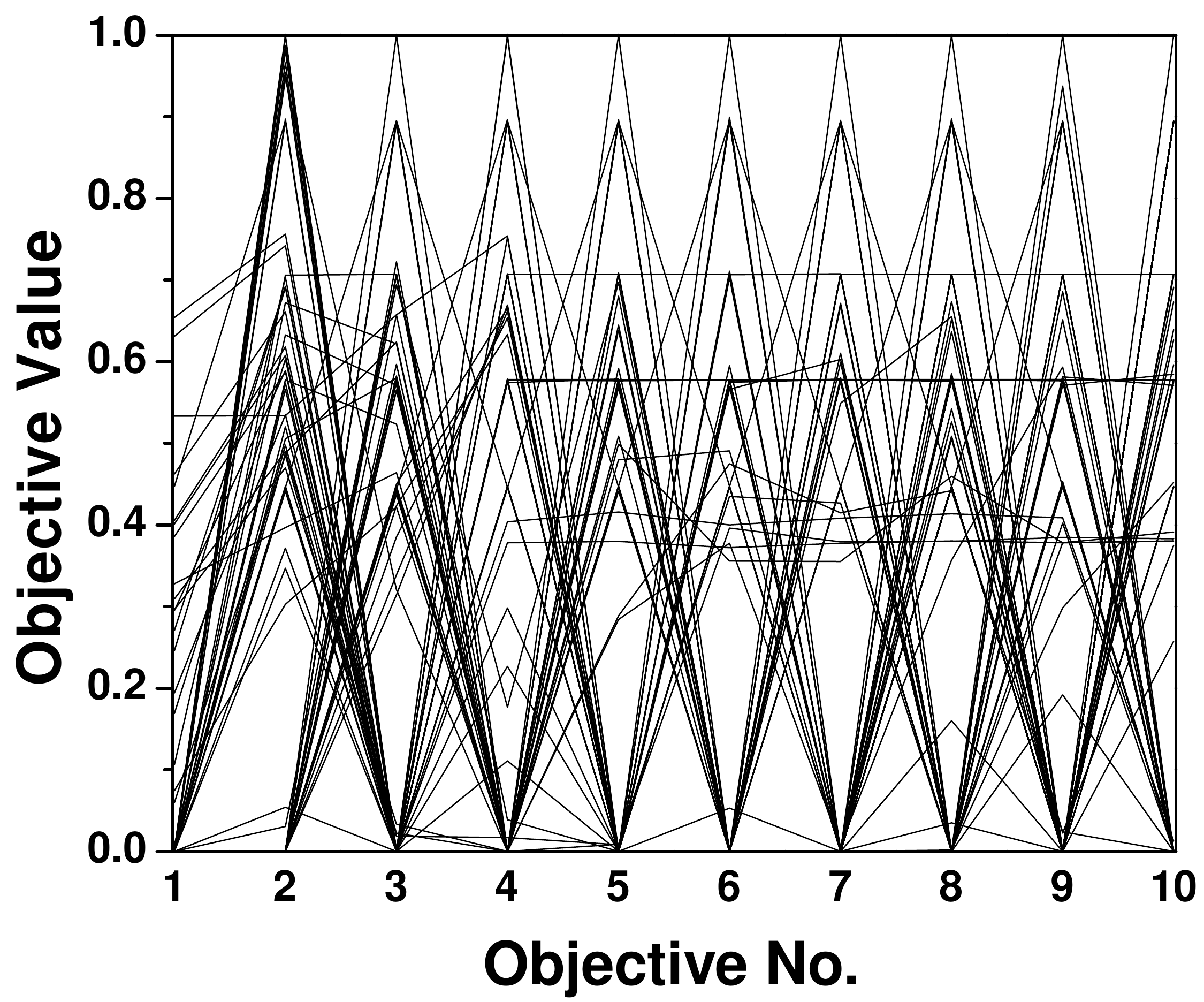}~&
			~\includegraphics[scale=0.15]{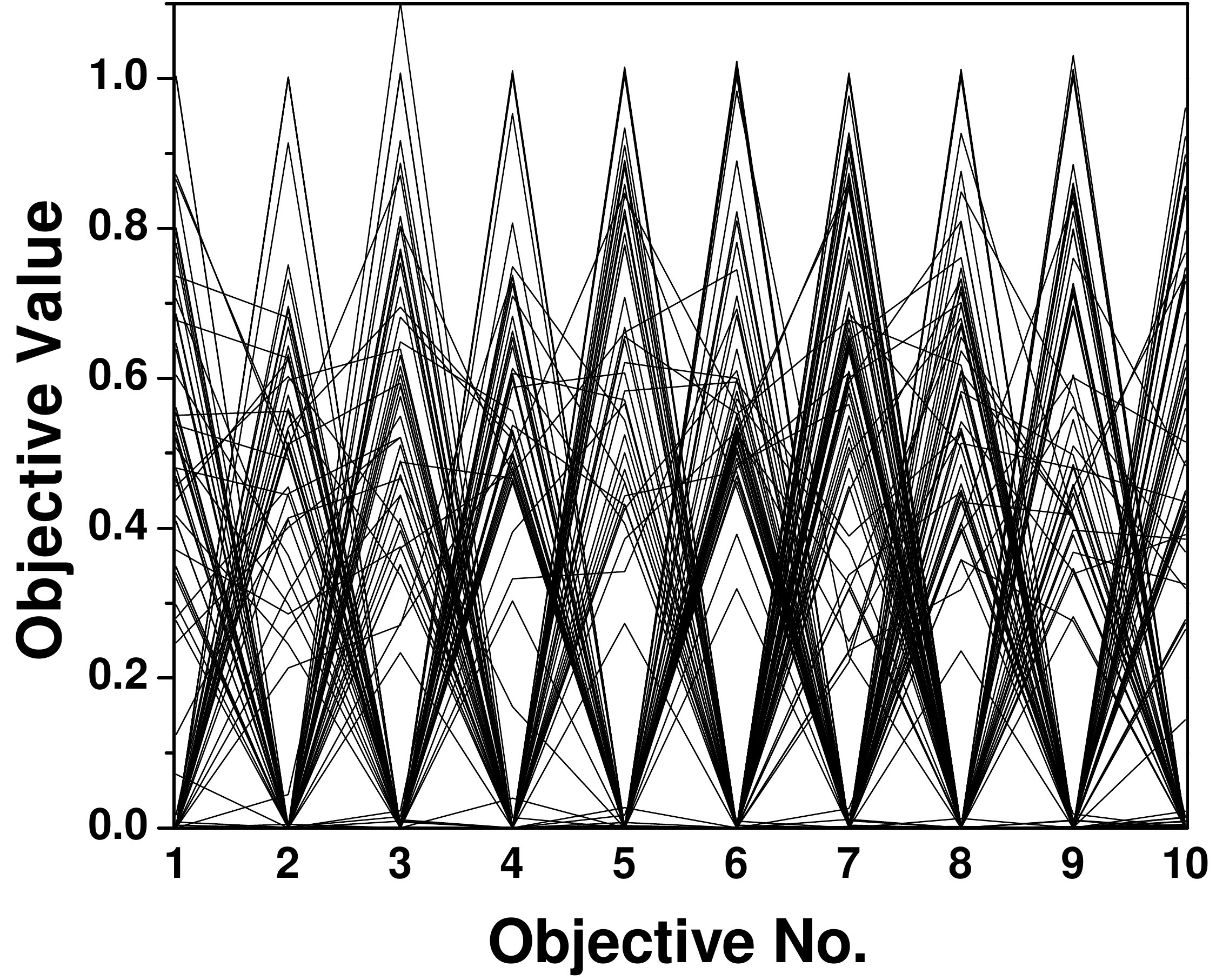}\\
			(a) MOEA/D & (b) A-NSGA-III & (c) RVEA & (d) MOEA/D-AWA & (e) AdaW \\
		\end{tabular}
	\end{center}
	\caption{The final solution set of the five algorithms on the 10-objective DTLZ2.}
	\label{Fig:DTLZ2-10}
\end{figure*}

For the 10-objective DTLZ2 which has a simplex-like Pareto front,
all the five algorithms appear to work well (\mbox{Figure~\ref{Fig:DTLZ2-10}}) 
despite that there exists one solution of AdaW not converging into the Pareto front. 
We may not be able to conclude the distribution difference of the algorithms 
by the parallel coordinates plots~\cite{Li2017a}, 
but all the algorithms seem to perform similarly according to the IGD results 
in \mbox{Table~\ref{Table:IGDresults}}.

\begin{figure*}[tbp]
	\begin{center}
		\footnotesize
		\begin{tabular}{@{}c@{}c@{}c@{}c@{}c@{}}
			\includegraphics[scale=0.15]{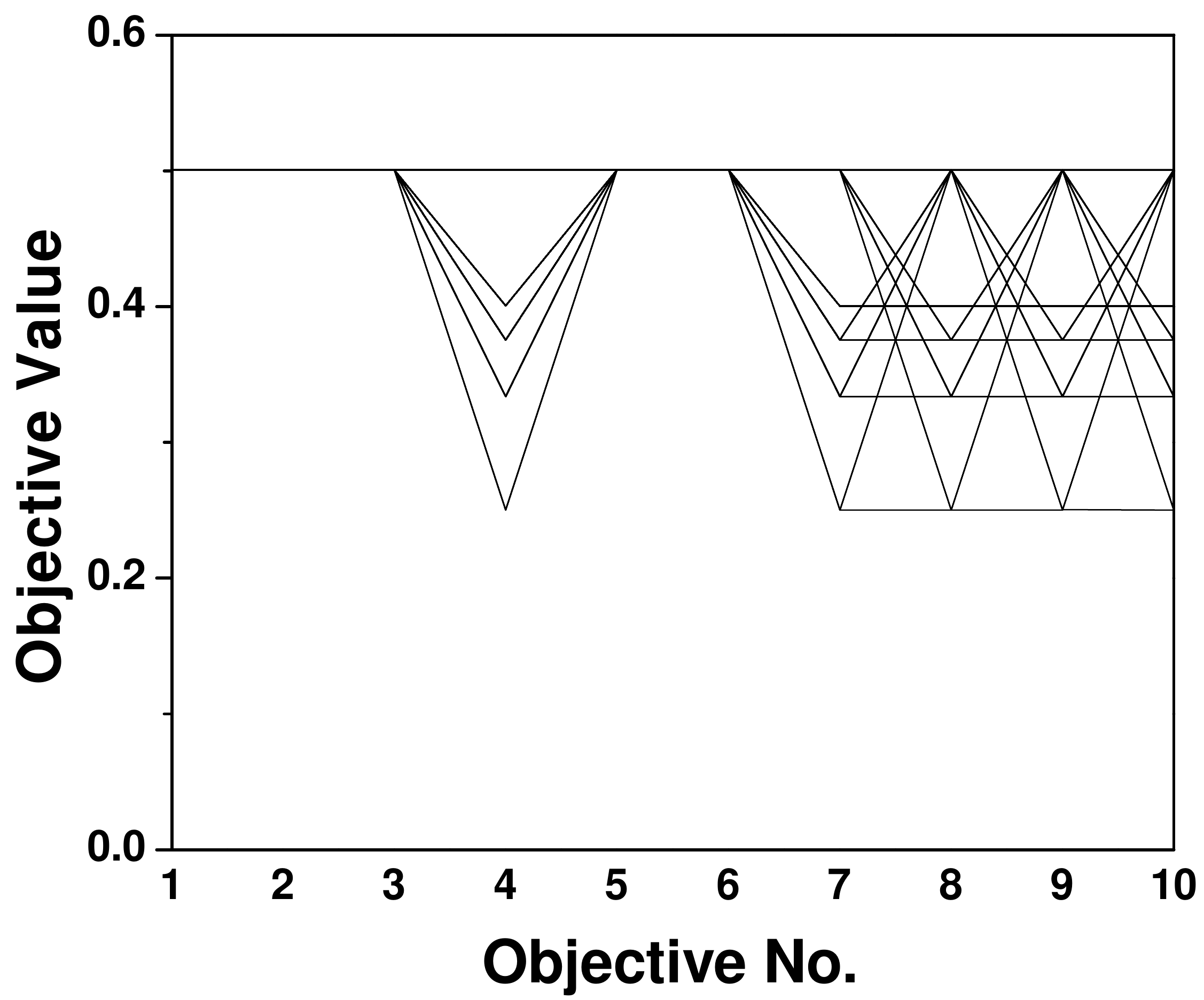}~&
			~\includegraphics[scale=0.15]{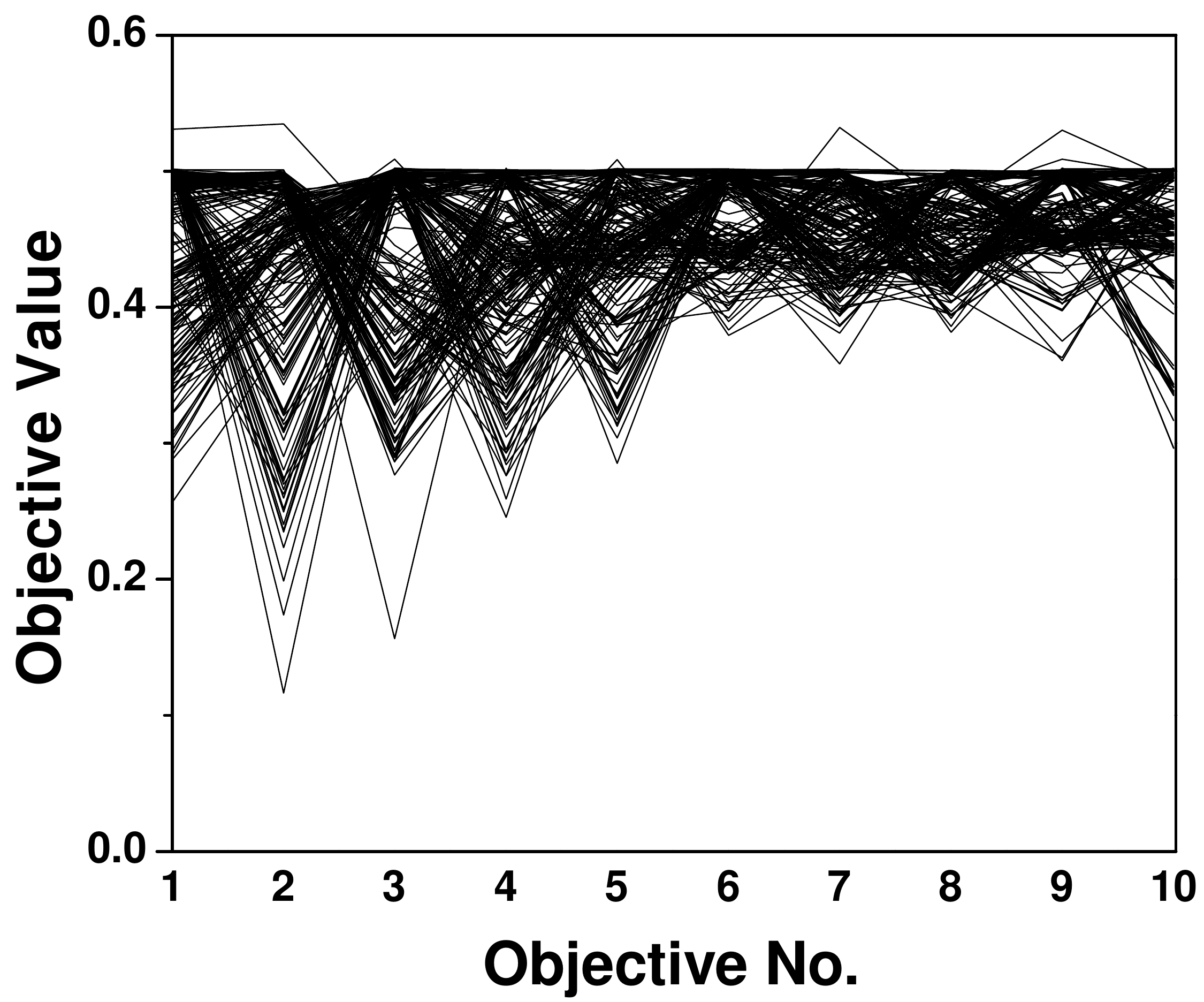}~&
			~\includegraphics[scale=0.15]{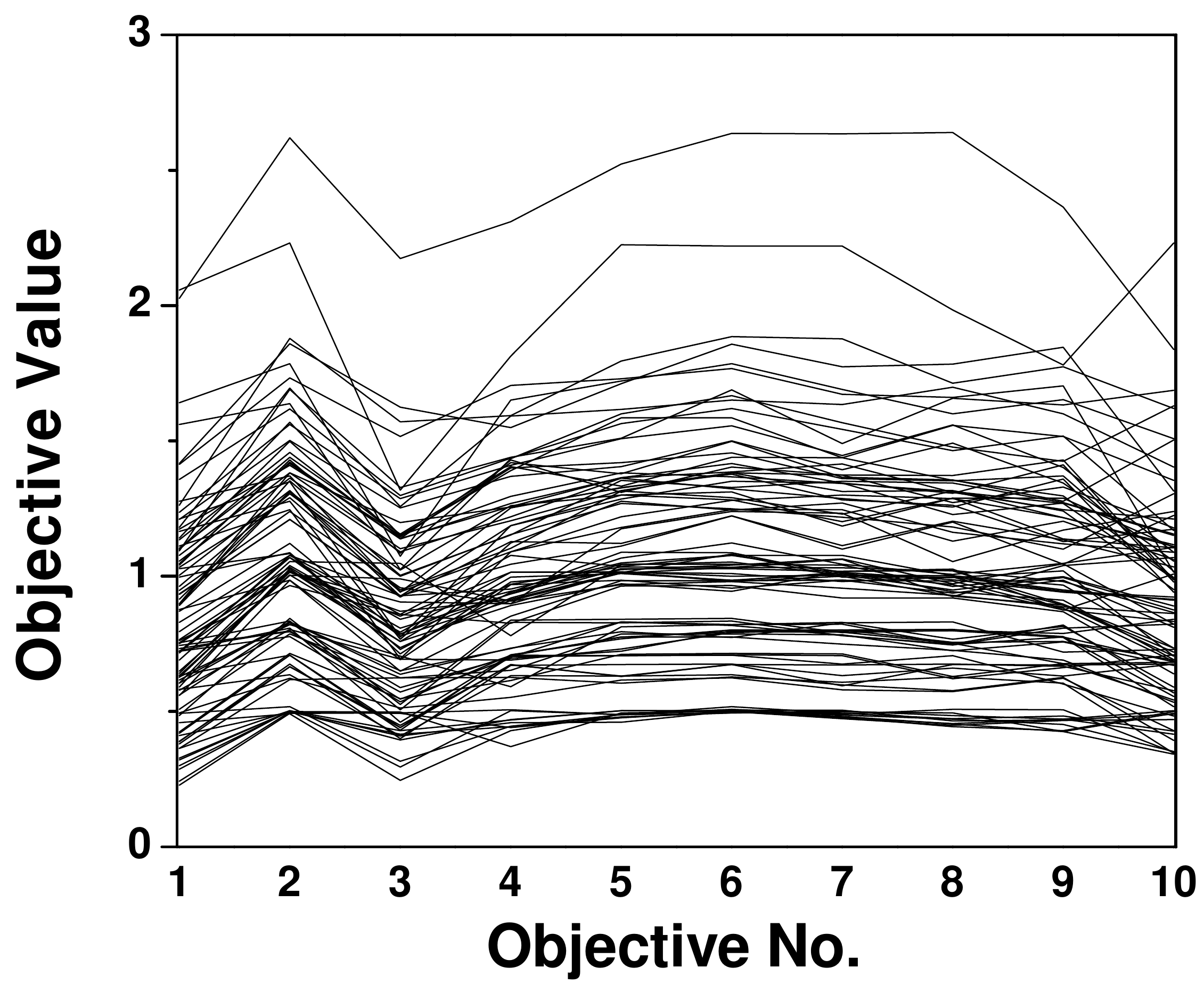}~&
			~\includegraphics[scale=0.15]{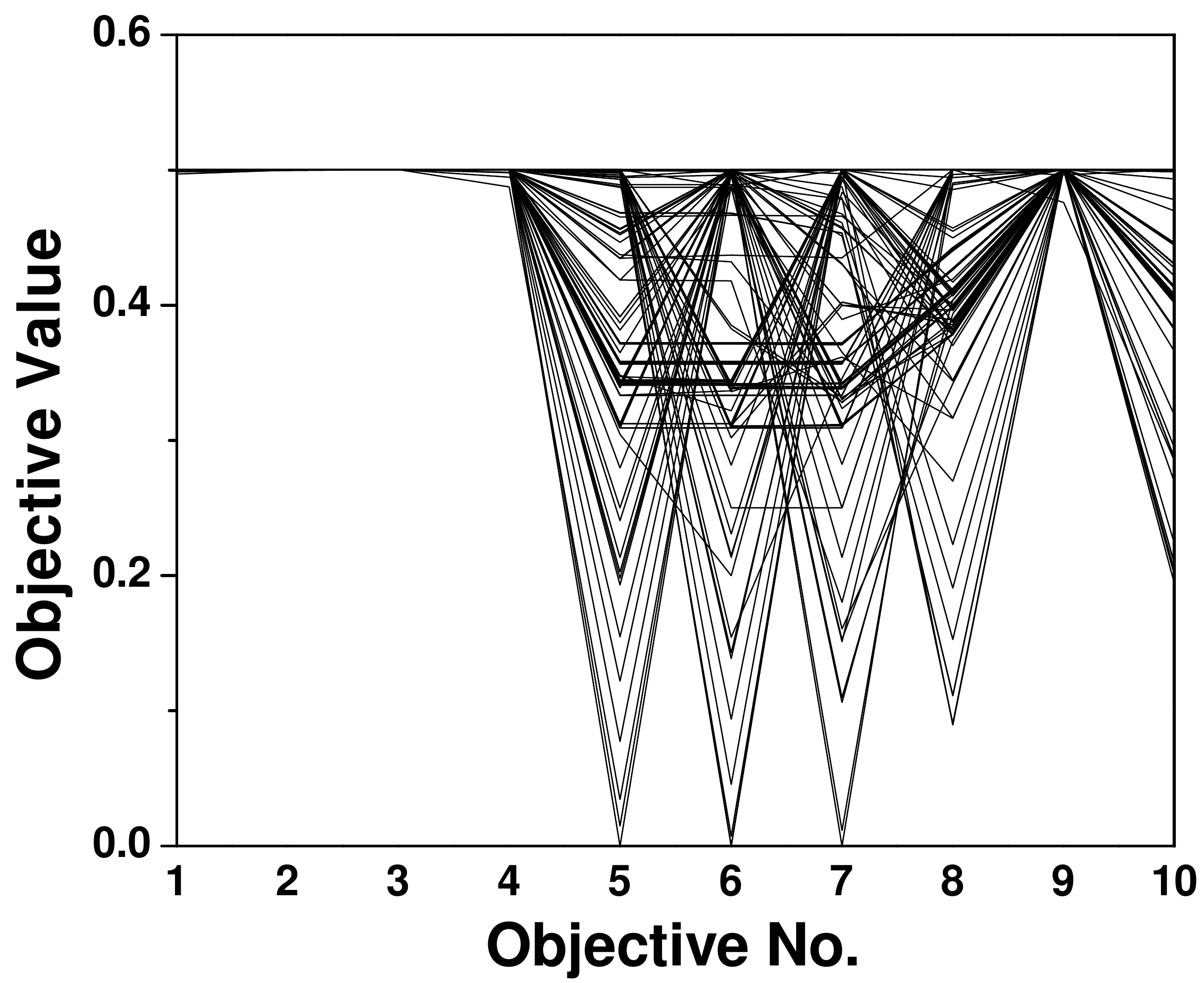}~&
			~\includegraphics[scale=0.15]{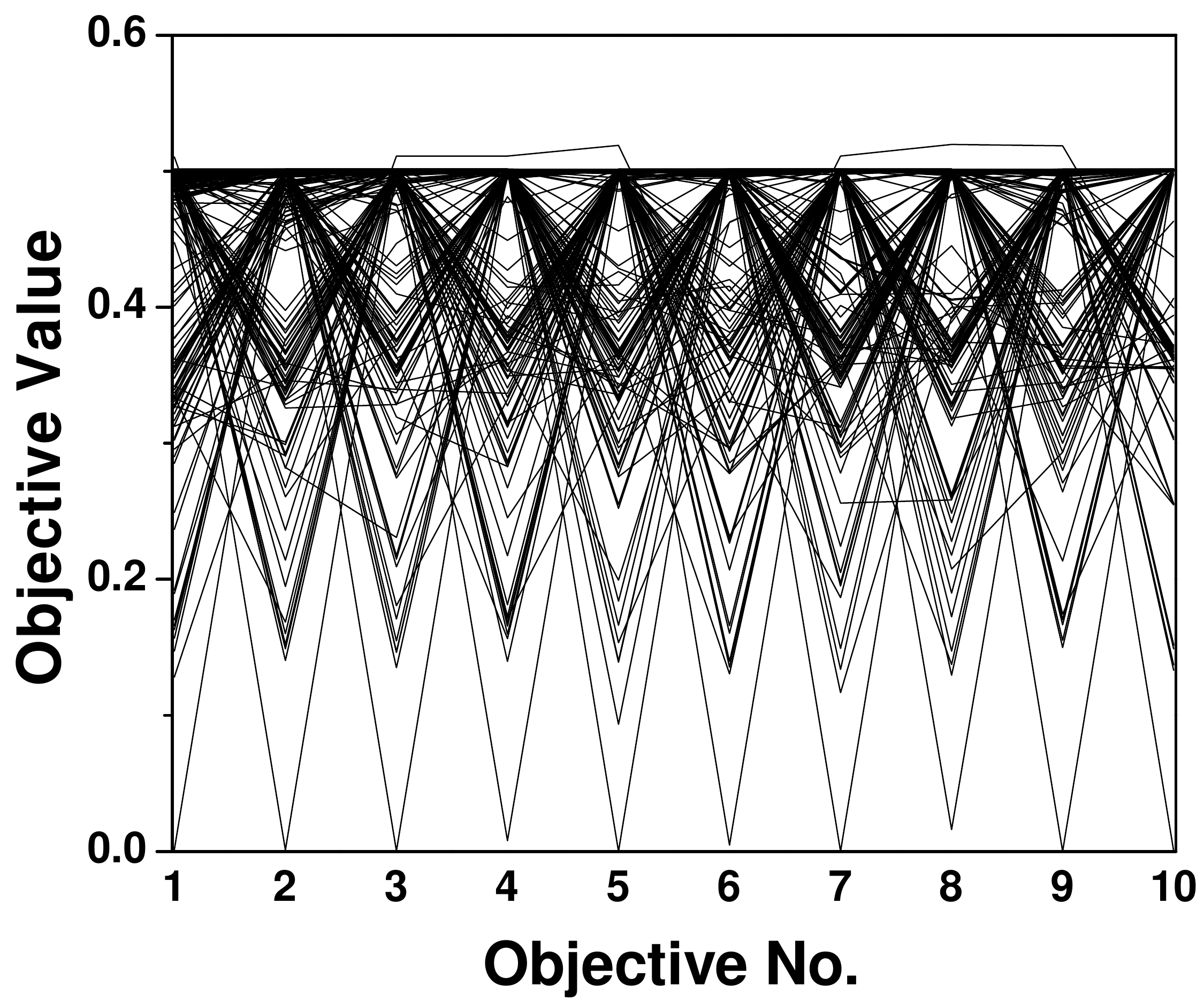}\\
			(a) MOEA/D & (b) A-NSGA-III & (c) RVEA & (d) MOEA/D-AWA & (e) AdaW \\
		\end{tabular}
	\end{center}
	\caption{The final solution set of the five algorithms on the 10-objective inverted DTLZ1.}
	\label{Fig:IDTLZ1-10}
\end{figure*}

\begin{figure*}[!]
	\begin{center}
		\footnotesize
		\begin{tabular}{@{}c@{}c@{}c@{}c@{}c@{}}
			\includegraphics[scale=0.15]{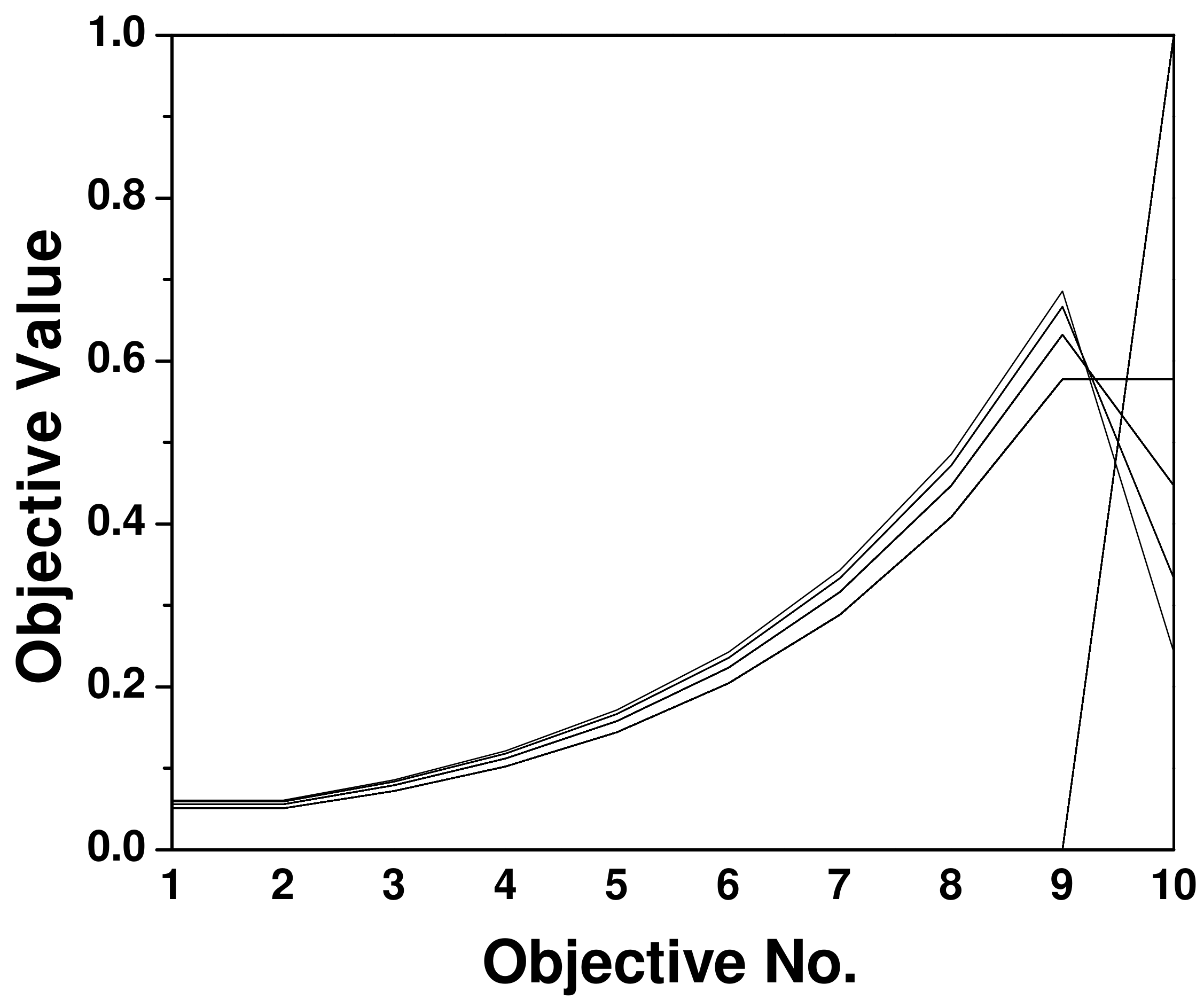}~&
			~\includegraphics[scale=0.15]{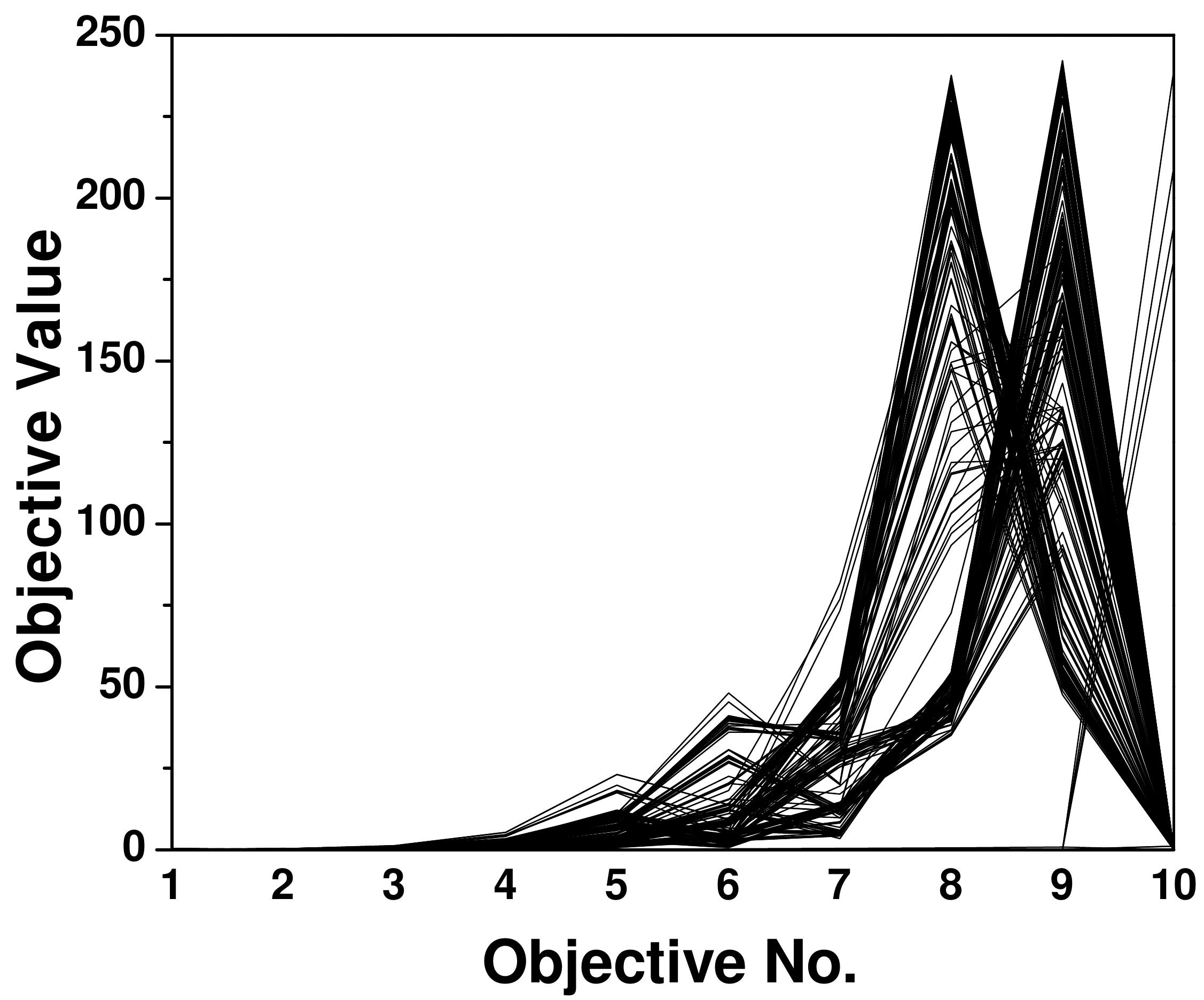}~&
			~\includegraphics[scale=0.15]{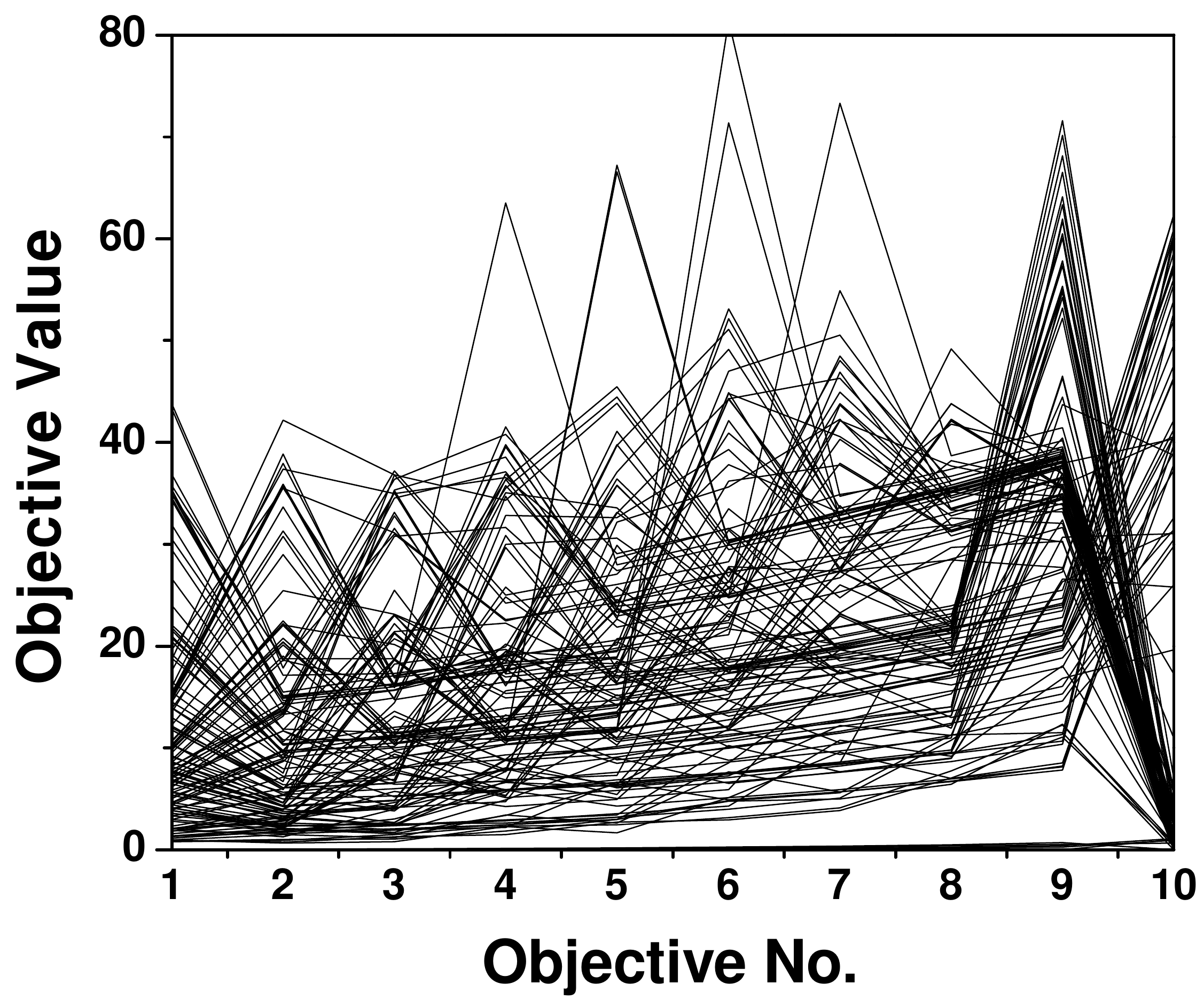}~&
			~\includegraphics[scale=0.15]{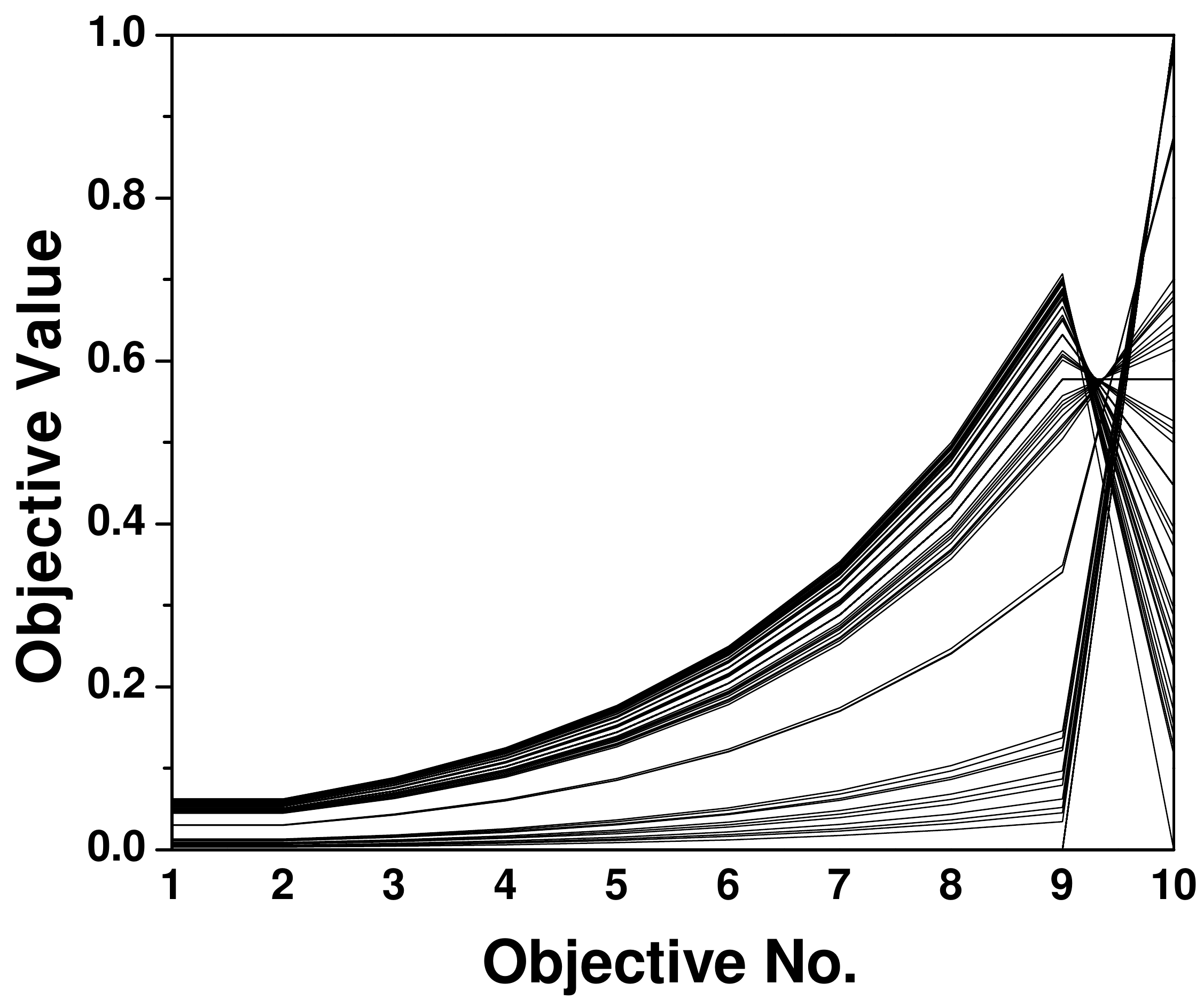}~&
			~\includegraphics[scale=0.15]{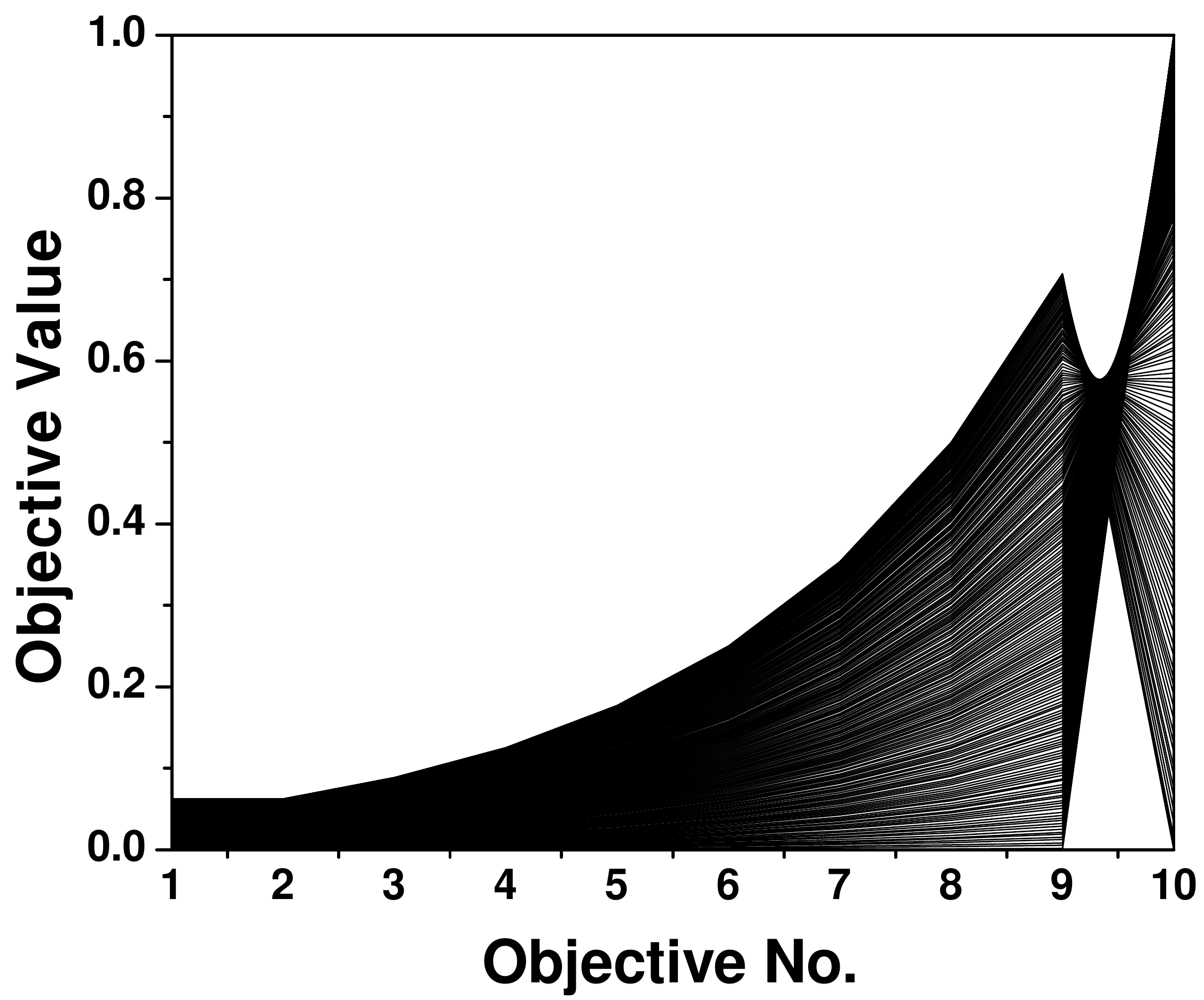}\\
			(a) MOEA/D & (b) A-NSGA-III & (c) RVEA & (d) MOEA/D-AWA & (e) AdaW \\
		\end{tabular}
	\end{center}
	\caption{The final solution set of the five algorithms on the DTLZ5(2,10).}
	\label{Fig:DTLZ5IM}
\end{figure*}

For the many-objective problems whose Pareto front is far from the standard simplex, 
a clear advantage of AdaW over its competitors is shown (\mbox{Figures~\ref{Fig:IDTLZ1-10} and \ref{Fig:DTLZ5IM}}).
The peer algorithms either fail to cover the whole Pareto front 
(i.e., MOEA/D, A-NSGA-III and MOEA/D-AWA on the 10-objective IDTLZ1 and 
MOEA/D and MOEA/D-AWA on DTLZ5(2,10)), 
or struggle to converge into the front (i.e., RVEA on the 10-objective IDTLZ1 and 
A-NSGA-III and RVEA on DTLZ5(2,10)).
In contrast, 
the proposed AdaW has shown its ability in dealing with irregular Pareto fronts in the high-dimensional space, 
by which a spread of solutions over the whole Pareto front is obtained.



\section{Conclusions}

Adaptation of the weight vectors during the optimisation process 
provides a viable approach to enhance existing decomposition-based EMO. 
This paper proposed an adaptation method to periodically update the weight vectors 
by contrasting the current evolutionary population with a well-maintained archive set.
From experimental studies on seven categories of problems with various properties,
the proposed algorithm has shown its high performance over a wide variety of different Pareto fronts.

However, 
it is worth noting that the proposed algorithm needs more computational resources than the basic MOEA/D.
The time complexity of AdaW is bounded by $O(mN^2)$ or $O(TN^2)$ whichever is greater 
(where $m$ is the number of objectives and $T$ is the neighbourhood size), 
in contrast to $O(mTN)$ of MOEA/D.
In addition, 
AdaW also incorporates several parameters, 
such as the maximum capacity of the archive and the time of updating the weight vectors.
Although these parameters were fixed on all test problems in our study, 
customised settings for specific problems may lead to better performance.
For example, 
a longer duration allowing the weight vectors evolving along the constant weight vectors is expected 
to achieve better convergence on problems with many objectives.

\end{document}